\documentclass[10pt,twocolumn,letterpaper]{article}

\usepackage[pagenumbers]{cvpr} %

\usepackage{microtype}

\makeatletter
\def\cvprsection{\@startsection{section}{1}{\z@}{-8pt plus -1pt minus -1pt}{5pt}{\large\bf}}
\def\cvprsubsection{\@startsection{subsection}{2}{\z@}{-6pt plus -1pt minus -1pt}{4pt}{\elvbf}}
\def\cvprsubsubsection{\@startsection{subsubsection}{3}{\z@}{-5pt plus -1pt minus -1pt}{2.5pt}{\tenbf}}
\makeatother

\usepackage{multirow}
\usepackage{tabularx}

\definecolor{cvprblue}{rgb}{0.21,0.49,0.74}
\usepackage[pagebackref,breaklinks,colorlinks,allcolors=cvprblue]{hyperref}

\usepackage{booktabs}
\usepackage{multirow}
\usepackage{pifont}

\newcommand{\cmark}{\ding{51}}  %
\newcommand{\xmark}{\ding{55}}  %

\newcommand{\model}{MERG3R}
\newcommand{\geomodel}{geometric foundation model}

\def\paperID{} %
\def\confName{CVPR}
\def\confYear{2026}

\title{\model{}: A Divide-and-Conquer Approach to Large-Scale Neural Visual Geometry}

\author{
    Leo Kaixuan Cheng\textsuperscript{1,*}\qquad
    Abdus Shaikh\textsuperscript{1,*}\qquad
    Ruofan Liang\textsuperscript{1, 2} \\[.2em]
    Zhijie Wu\textsuperscript{1, 2}\qquad
    Yushi Guan\textsuperscript{1, 2}\qquad
    Nandita Vijaykumar\textsuperscript{1, 2}\\[.5em]
    \textsuperscript{1}University of Toronto\quad
    \textsuperscript{2}Vector Institute %
}

\newcommand{\name}{MERG3R}

\begin{document}
\twocolumn[{
    \maketitle
    \centering
    \vspace{-1em}
    \includegraphics[width=\textwidth]{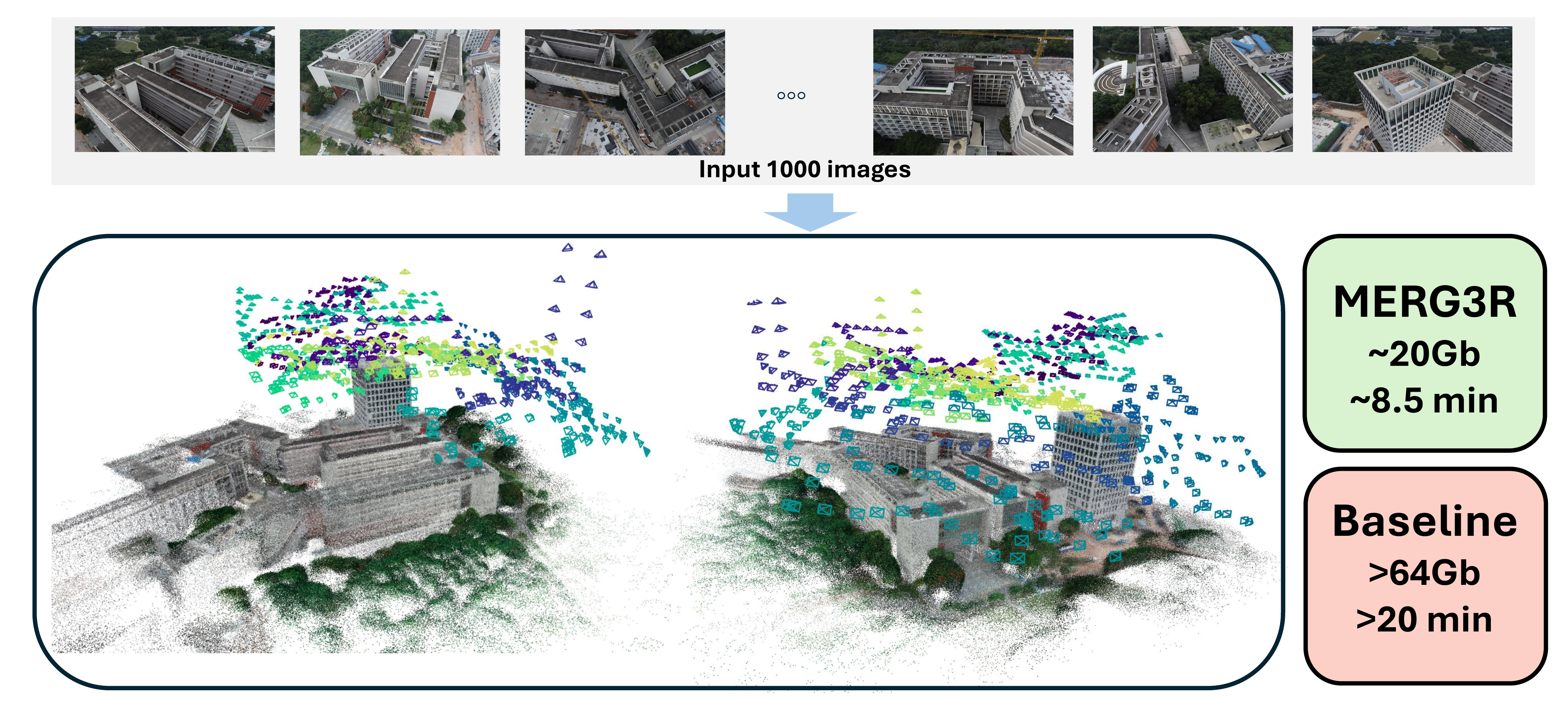}
    \captionof{figure}{Given a large unordered set of 1,000 input images, \name{} reconstructs accurate camera poses and a high-quality point cloud. Despite the long sequence of images that may not fit on device memory and challenging viewpoints, our pipeline enables scalable and  reliable geometry reconstruction. Project page: \url{https://leochengkx.github.io/MERG3R/}}
    \label{fig:teaser}
    \vspace{1.5em}
}]
\iftoggle{cvprfinal}{%
{\let\thefootnote\relax\footnotetext{{* Equal contribution}}}
}{}
\begin{abstract}
Recent advancements in neural visual geometry, including transformer-based models such as VGGT and Pi3, have achieved impressive accuracy on 3D reconstruction tasks. However, their reliance on full attention makes them fundamentally limited by GPU memory capacity, preventing them from scaling to large, unordered image collections. We introduce \name, a training-free divide-and-conquer framework that enables geometric foundation models to operate far beyond their native memory limits. \name{} first reorders and partitions unordered images into overlapping, geometrically diverse subsets that can be reconstructed independently. It then merges the resulting local reconstructions through an efficient global alignment and confidence-weighted bundle adjustment procedure, producing a globally consistent 3D model. Our framework is model-agnostic and can be paired with existing neural geometry models. Across large-scale datasets—including 7-Scenes, NRGBD, Tanks \& Temples, and Cambridge Landmarks—\name{} consistently improves reconstruction accuracy, memory efficiency, and scalability, enabling high-quality reconstruction when the dataset exceeds memory capacity limits.
\end{abstract}
    
\section{Introduction}
Reconstructing 3D scenes from a collection of 2D images is a fundamental problem in computer vision, powering a wide range of applications from autonomous navigation~\cite{mur2015orb} to virtual/mixed reality~\cite{applearkit} and cultural heritage preservation~\cite{agarwal2011building}. 
Traditional pipelines based on Structure-from-Motion (SfM) and Multi-View Stereo (MVS), 
 while robust and widely adopted, may require extensive engineering and struggle in low-texture or repetitive regions, motivating a shift toward learned, feed-forward 3D reconstruction with neural visual geometry methods. 
End-to-end, transformer-based models like Dust3R~\cite{wang2024dust3r}, Mast3R~\cite{leroy2024grounding,duisterhof2025mast3r}, VGGT~\cite{wang2025vggt}, Pi3~\cite{wang2025pi}, and MapAnything~\cite{keetha2025mapanything} have demonstrated remarkable performance, learning to jointly infer camera parameters and dense 3D point clouds with remarkable accuracy directly from images.

Despite the rapid progress of neural visual geometry,  transformer-based reconstruction models share a critical limitation: poor scalability as these models are fundamentally limited by GPU memory capacity. Monolithic models such as VGGT, Pi3, and MapAnything must encode all input images simultaneously, which causes the number of visual tokens to grow linearly with 
the number of images, while the self-attention mechanism~\cite{vaswani2017attention} grows quadratically in both computation and memory.
This scalability bottleneck severely limits their practical utility for real-world applications like city-scale modeling or reconstructing large, complex environments from thousands of images.

Efforts to improve the scalability of neural visual geometry models often come at the cost of reconstruction accuracy. Approaches such as VGGT-Long~\cite{deng2025vggt},  FastVGGT~\cite{shen2025fastvggt}, and Fast3R~\cite{yang2025fast3r} reduce computational burden by chunking inputs or merging tokens, but these approximations weaken long-range geometric reasoning and degrade pose or depth estimation, especially in scenes with wide viewpoint variation. Furthermore, FastVGGT and Fast3R must encode images simultaneously, and thus are still limited by memory capacity. In contrast, more classical neural approaches like CUT3R~\cite{wang2025continuous} and TTT3R~\cite{, chen2025ttt3r} avoid full self-attention by relying on independent per-image depth prediction followed by multi-view fusion or test-time optimization, giving them much better raw scalability. However, these models do not maintain a global geometric representation across all images, leading to rapid degradation in accuracy as the number of input images increases. Overall, current neural approaches must choose between memory scalability and geometric accuracy.

Our goal in this work is to develop a scalable divide-and-conquer pipeline for neural geometry models that enables robust reconstruction from large, unordered image sets without sacrificing global geometric accuracy. We propose \name, a framework built on three key ideas: First, we develop a clustering strategy that partitions unordered images into subsets that (1) provide sufficient multi-view coverage for accurate local reconstruction and (2) maintain overlap with other subsets to enable consistent downstream global alignment. Each cluster is designed to fit within the GPU memory constraints and is reconstructed independently using any geometric foundation model~\cite{wang2025vggt,wang2025pi} to produce a high-quality local reconstruction. 

Second, we introduce an efficient method for constructing global point tracks across clusters using a lightweight feature-matching model, producing reliable multi-view correspondences that link all local reconstructions. Third, to achieve high global consistency and reconstruction accuracy, we introduce an efficient global bundle adjustment step to jointly optimize the camera intrinsics, extrinsics and 3D point positions. This gradient-based optimization is performed over the previously generated confidence-weighted multi-view point tracks to enable both better efficiency and global consistency than prior approaches that optimize over every image pair in the scene graph~\cite{duisterhof2025mast3r}.

The key contributions are:
\begin{itemize}
\item We introduce a training-free pipeline that enables modern geometric foundation models to operate on large, unordered image collections far beyond their native memory limits. Our modular divide-and-conquer approach also enables parallelizing computation across multiple GPUs and significantly faster execution times by partitioning images into clusters.
\item We demonstrate that how images are clustered plays an important role in the success of local reconstruction with neural methods and downstream global alignment.
\item When integrated with state-of-art neural geometry models, our extensive experiments on the 7-Scenes, NRGBD, Tanks \& Temples, and Cambridge Landmarks datasets shows that our method achieves superior accuracy, memory efficiency, and scalability when images do not fit in GPU memory. 
\end{itemize}
\begin{table}[htbp]
\centering
\setlength{\tabcolsep}{2pt}
\resizebox{0.9\linewidth}{!}{
\begin{tabular}{lccccc}
\toprule
\multirow{2}{*}{Method} &
\multicolumn{2}{c}{State-of-art accuracy} &
\multirow{2}{*}{\shortstack{Memory\\scalability}} &
\multirow{2}{*}{\shortstack{Works for\\unordered}}  \\
\cmidrule(lr){2-3}
& Ordered & Unordered & & & \\
\midrule

Mast3R-SFM \cite{duisterhof2025mast3r} & \xmark & \xmark & \xmark & \cmark  \\
CUT3R~\cite{wang2025continuous}        & \xmark & \xmark & \cmark & \cmark  \\
TTT3R~\cite{chen2025ttt3r}        & \xmark & \xmark & \cmark & \cmark  \\
VGGT~\cite{wang2025vggt}        & \xmark & \xmark & \xmark & \cmark  \\
VGGT-Long~\cite{deng2025vggt}    & \xmark & \xmark & \cmark & \xmark  \\
FastVGGT~\cite{shen2025fastvggt}    & \xmark & \xmark & \xmark & \cmark  \\
Pi3~\cite{wang2025pi}         & \cmark & \cmark & \xmark & \cmark \\
Fast3R~\cite{yang2025fast3r}       & \xmark & \xmark & \xmark & \cmark \\
\midrule\midrule

Ours+FastVGGT & \xmark & \xmark & \cmark & \cmark   \\
Ours+VGGT / +Pi3 & \cmark & \cmark & \cmark & \cmark \\

\bottomrule
\end{tabular}
}
\caption{Conceptual comparison of methods across accuracy, memory scalability, and unordered-image compatibility. 
Our method can achieve superior or comparable accuracy over other SOTA baselines.}
\end{table}

\vspace{-0.3cm}

\begin{figure*}[t]
    \centering
    \includegraphics[width=0.9\textwidth,trim=0 0cm 0 0cm,clip]{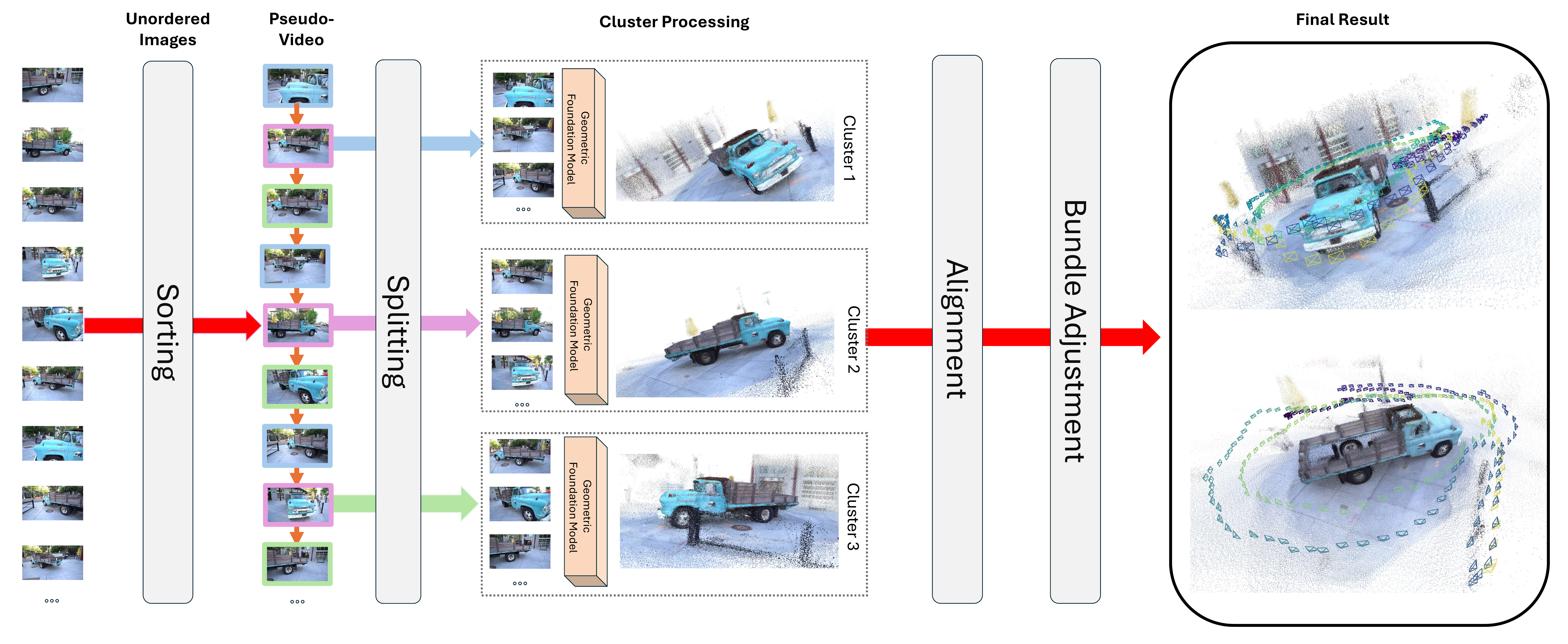}
    \caption{Overview of our large-scale 3D reconstruction pipeline. Given an unordered set of images, we first sort them into a pseudo-video sequence, then split the sequence into multiple interleaved subsets. Each subset is independently processed by a geometric foundation model to produce local pointmaps and poses. The resulting clusters are aligned into a common reference frame and jointly refined via global bundle adjustment, producing a coherent final reconstruction.}
    \label{fig:pipeline}
\end{figure*}

\section{Related Work}

\subsection{Traditional 3D Reconstruction}
3D reconstruction has historically relied on traditional Structure-from-Motion (SfM) and Multi-View Stereo (MVS) pipelines, which recover camera parameters and scene geometry through a sequence of geometric operations~\citep{ozyecsil2017survey}.  
SfM methods~\citep{agarwal2011building,schoenberger2016sfm,snavely2006photo} detect and match local features across image pairs, estimate pairwise epipolar geometry, incrementally or globally register cameras, and refine all parameters via bundle adjustment (BA) to jointly optimize intrinsics, extrinsics, and sparse 3D points. 
Systems such as COLMAP~\citep{schoenberger2016sfm}  are widely used today due to their robustness and generality, powering applications from photo-realistic 3D reconstruction to indoor SLAM~\citep{mildenhall2020nerf,kerbl20233d}. 
To obtain dense geometry, SfM is typically combined with MVS techniques~\citep{furukawa2015multi,schoenberger2016mvs,yao2018mvsnet,gu2020cascade} which compute per-pixel depth or dense point clouds by enforcing photometric consistency across multiple calibrated views. Despite their maturity, these pipelines remain computationally expensive, and reconstruction quality can significantly degrade in low-texture, repetitive, or strongly varying illumination conditions where correspondences are unstable. Their multi-stage nature also requires extensive heuristics and careful engineering. These limitations have motivated the development of neural visual geometry methods.

\subsection{Feed-Forward Neural 3D Reconstruction}
The recent paradigm shift towards end-to-end learning has introduced a new class of feed-forward neural 3D reconstruction models that directly infer camera poses and scene geometry without relying on the traditional SfM pipeline~\citep{wang20243d,yang2025fast3r,wang2025continuous,zhang2025flare,keetha2025mapanything}.
Early work such as DUSt3R~\cite{wang2024dust3r} , demonstrated that transformer-based architectures can predict dense “pointmaps” from image pairs, enabling recovery of camera poses and coarse 3D structure. MASt3R~\citep{leroy2024grounding} extends this idea by additionally predicting pixelwise correspondences, producing more reliable multi-view constraints and enabling downstream systems like MASt3R-SfM~\citep{duisterhof2025mast3r} to align multiple pairwise reconstructions.
Building on these foundations, VGGT~\citep{wang2025vggt} generalizes feed-forward reconstruction to unordered multi-view inputs, jointly predicting camera intrinsics, extrinsics, depth maps, features for correspondence, and per-pixel confidence scores within a single network pass. %
Although these models achieve state-of-art results, they rely on full transformer attention, whose memory and computational cost grows quadratically with the number of images, thereby restricting their applicability to relatively small image sets. $\pi^3$~\citep{wang2025pi} further improves VGGT by introducing a permutation-equivariant architecture that improves generalization and efficiency over VGGT, but like other monolithic transformer-based approaches, it remains fundamentally limited by attention scaling. %

\subsection{Large-Scale 3D Neural Reconstruction}
While monolithic transformer-based models like VGGT have excelled at reconstructing scenes from a few hundred images, their reliance on a full attention mechanism creates a significant scalability bottleneck.
Classical SfM pipelines have long addressed large-scale settings using divide-and-conquer strategies—partitioning images based on visual similarity graphs and merging partial reconstructions through global alignment techniques~\citep{zhu2018very,bhowmick2014divide,chen2020graph}.
Inspired by these ideas, recent neural approaches propose scalable variants of transformer-based reconstruction:
VGGT-Long~\citep{deng2025vggt} processes long sequences by chunking video frames into manageable segments and aligning overlapping chunks to form a global trajectory. This approach requires ordered images or videos, and as we demonstrate, chunking can compromise reconstruction quality. Other efforts aim to reduce the computational burden of attention itself; for example, 
FastVGGT~\citep{shen2025fastvggt} merges redundant tokens to efficiently handle large inputs while preserving geometric fidelity. Fast3R~\cite{yang2025fast3r} restructures the transformer to handle high token counts by combining efficient attention, token reduction, and a hierarchical feature fusion strategy, allowing it to scale to image collections that exceed the limits of previous architectures. However, this approach is still fundamentally bounded by memory constraints, due to the need to process the entire set simultaneously. 

In comparison to these approaches, we propose a novel divide-and-conquer framework that operates directly on large, \emph{unordered} image collections, combining a new principled partitioning algorithm with alignment and global bundle adjustment to achieve significantly higher geometric consistency while keeping memory usage bounded; moreover, our system is model-agnostic and can be paired with any pretrained geometric foundation model to further extend its scalability and accuracy.

\section{Method}
We outline our large-scale 3D reconstruction pipeline (Fig.~\ref{fig:pipeline}), covering (i) preliminaries on \geomodel{}s (\ref{subsec:preliminaries}), (ii) image set partitioning (\ref{subsec:ordering_and_partitioning}), (iii) local reconstruction (\ref{subsec:local_reconstruction}), and (iv) track merging (\ref{subsec:track_merging}).

\subsection{Preliminaries}\label{subsec:preliminaries}
We use VGGT~\cite{wang2025vggt} to provide representative brief 
preliminaries of the \geomodel{}s.
Given a set of $N$ images $\mathcal{I} = (I_i)_{i=1}^N$, VGGT tokenizes images using DINO~\cite{oquab2023dinov2} and processes the resulting tokens with multiple blocks of Alternating Attention, mixing global attention across all images with per-frame attention. The resulting latents are fed to multiple heads that predict:
\begin{itemize}
    \item Camera parameters for each frame $I_i$: intrinsics $\mathbf{K}_i$ and pose $(\mathbf{R}_i,\,\mathbf{t}_i)$. 
    We refer to the camera parameters of frame $I_i$ as $\mathbf{G}_i = (\mathbf{K}_i, \mathbf{R}_i, \mathbf{t}_i)$.
    \item Dense depth maps $\mathbf{D}_i$ from DPT~\cite{ranftl2021vision} for each image.
    \item Dense features for tracking and correspondence~\cite{wang2024vggsfm}.
    \item Confidence scores $\mathbf{C}_i$ that quantify per-pixel uncertainty in the depth and correspondence estimates.
\end{itemize}
A \geomodel{} $F_g$ can thus be summarized as:
\begin{equation}
    F_g(\mathcal{I}) = (\mathcal{G}, \mathcal{D}, \mathcal{C})
\end{equation}
where $\mathcal{G}, \mathcal{D}, \mathcal{C}$ are the sets of camera parameters, depth/point maps, and confidence scores for the entire image set $\mathcal{I}$.
The dense point cloud of the captured scene can then be obtained by inverse projecting pixels using the predicted depths and camera matrices.

\begin{figure}[t]
    \centering
    \includegraphics[width=\columnwidth,trim=0 22.6cm 0 0cm, clip]{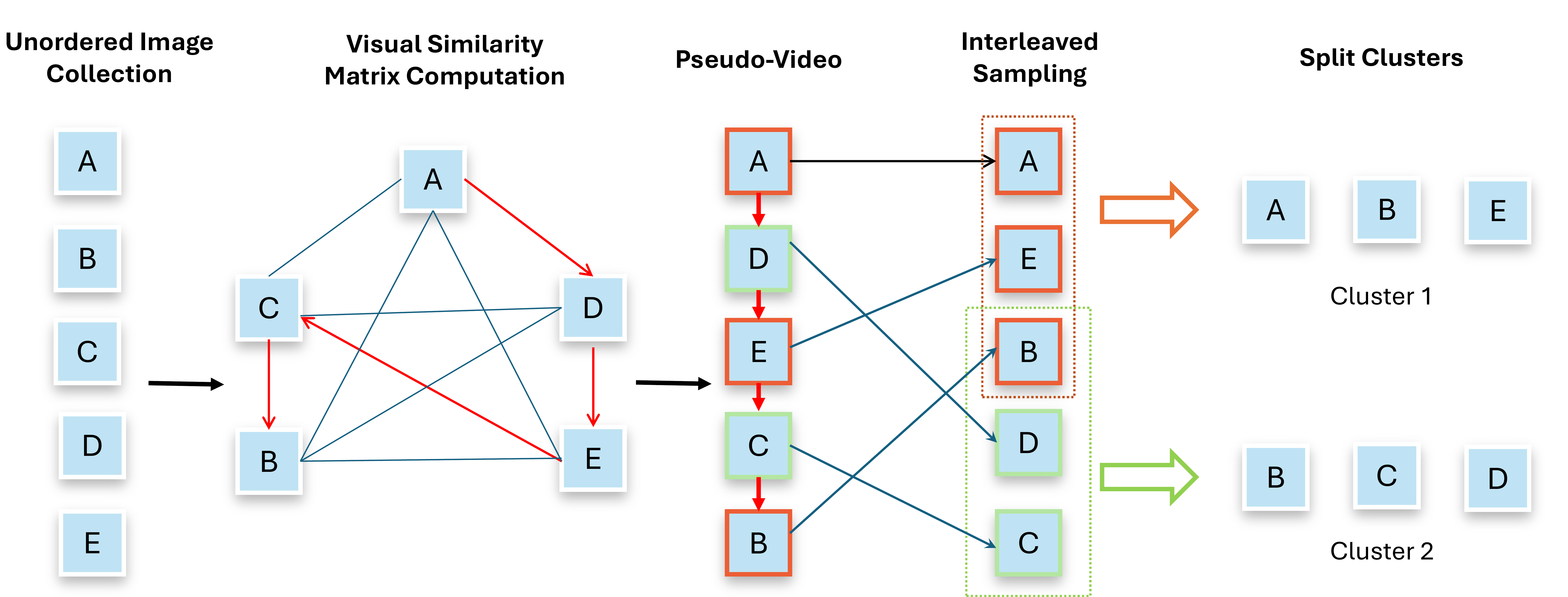}
    \caption{Illustration of the image partitioning process.
    Given an unordered set of $n$ images, we first compute the visual-similarity matrix $\mathbf{M} \in \mathbb{R}^{n \times n}$ and use it to search for a Hamiltonian path (shown in red) to produce the pseudo-video sequence.
    We then reorder the images using interleaved sampling and divide them into overlapping clusters.
    }
    \label{fig:image_partition}
\end{figure}

\subsection{Image Set Ordering and Partitioning}\label{subsec:ordering_and_partitioning}

Given a large, unordered image set $\mathcal{I} = (I_i)_{i=1}^N$ with potentially thousands of images, the first step of our pipeline is to partition this set into smaller, more manageable subsets to make \geomodel{} inference feasible in terms of memory and computation. Effective partitioning should ensure that each subset contains adequate variation in viewpoints for robust multi-view stereo, while keeping viewpoint changes moderate enough to preserve reliable correspondences. Furthermore, maintaining sufficient overlap between adjacent subsets is crucial so the resulting submaps can be accurately aligned and merged into a coherent global reconstruction.

To meet these requirements, we propose a two-step image set partitioning strategy. 
First, we impose a pseudo-temporal ordering on the unordered image set by constructing a sequence that maximizes visual continuity. Specifically, we compute a dense visual-similarity matrix $\mathbf{M} \in \mathbb{R}^{N \times N}$, where $\mathbf{M}_{i,j}$ is the DINO-based visual similarity between image $I_i$ and $I_j$. Interpreting this matrix as a weighted complete graph (with nodes corresponding to images and edge weights $\mathbf{M}_{i,j}$ denoting pairwise visual similarity), we approximate a Hamiltonian path $P = (p_1, \ldots, p_N)$ through all images that maximizes the sum of similarities between consecutive frames:
\begin{equation}
    P^* = \arg\max_{P} \sum_{k=1}^{N-1} \mathbf{M}_{p_k,\,p_{k+1}}
\end{equation}

Second, we partition this ordered sequence into overlapping subsets that satisfy our reconstruction and alignment requirements (i.e., to ensure sufficient overlap and geometric diversity across subsets.)
We first interleave the ordering across 
$K$ target subsequences, by permuting $P^*$ into a sequence $\tilde{P}$ that draws frames cyclically from across the full ordering. Formally, the $i$-th element of $\tilde{P}$ is $\tilde{P}_i = P^*\{(i \bmod K) \cdot K + \lfloor i / K \rfloor\}$. This prevents any cluster from receiving only temporally adjacent—and thus overly similar—views, ensuring richer viewpoint diversity. Without this interleaving, subsets can contain images with nearly identical viewpoints and thus produce unreliable local reconstructions. For particularly long sequences, we add a DINO similarity constraint on choosing the next frame.

We then slide a fixed-length window of $T$ elements across $\tilde{P}$ with stride $T - O$, where $O$ is the desired overlap. Each window defines a subset $\mathcal{S}_k = \{\tilde{P}{i} \mid i \in [k(T - O),\, k(T - O) + T)]\}$, and we retain only windows that contain at least $O$ frames beyond their starting point. 
This construction yields a sequence of locally manageable, overlapping subsets whose shared indices $\mathcal{S}_k \cap \mathcal{S}_{k+1}$ provide the necessary constraints for consistent global alignment.

\subsection{Local Reconstruction}\label{subsec:local_reconstruction}
With the partitioned subsets in hand, we independently reconstruct each subset using a pretrained geometric foundation model. For each subset $\mathcal{S}_k$, the model $F_g$ predicts camera parameters, depth maps, and confidence scores: $F_g(\mathcal{S}_k) = (\mathcal{G}_k, \mathcal{D}_k, \mathcal{C}_k)$. 
When a monolithic transformer processes the full image set, the cost of self-attention grows as $\mathcal{O}(N^2)$ attention complexity. By dividing the image set into $K$ subsets each of size $T$, this is reduced to $\mathcal{O}(KT^2)$, or equivalently $\mathcal{O}(\frac{N^2}{K})$ when subsets are processed sequentially. This not only reduces peak GPU memory requirements but also allows the reconstruction of different subsets to be parallelized across multiple GPUs, offering additional practical speedups.

\subsection{Cluster Alignment}\label{subsec:alignment}
Once each subset is reconstructed independently, we must align them into a single global model. Because subsets contain many unique views, even overlapping regions may yield inconsistent point maps, making naive alignment unreliable. To solve this problem, we adapt the weighted iterative similarity-transform estimator from VGGT-Long \cite{deng2025vggt}, which proved to be effective and robust.
For each pair of overlapping subset $\mathcal{S}_k$ and $\mathcal{S}_{k+1}$, we first identify a set of corresponding 3D points $\{(p_k^i, p^i_{k+1})\}$ and their confidence scores $\{(c^i_k, c^i_{k+1})\}$. We then filter out points whose confidence is below a percentile threshold $\tau_\text{conf}$. Finally, we align each adjacent pair of subsets using by solving for the similarity transform  $\mathbf{T} \in \mathrm{Sim}(3)$ that minimizes a robust Huber-based objective:
\begin{equation}
\mathbf{T}^{\ast}_{k,k+1}
= 
\arg\min_{\mathbf{T} \in \mathrm{Sim}(3)}
\sum_i \rho\!\left( ||p_i^{k} - \mathbf{T} p_i^{k+1}||_2 \right)
\end{equation}
where $\rho(\cdot)$ is the Huber loss function. We solve this optimization problem using Iteratively Reweighted Least Squares (IRLS). At iteration $t$ we solve
\begin{equation}
\mathbf{T}^{(t+1)}
=
\arg\min_{\mathbf{T} \in \mathrm{Sim}(3)}
\sum_i
w_i^{(t)} \,
\| p_i^{k} - \mathbf{T} p_i^{k+1} \|_2^2
\end{equation}
with weights
\begin{equation}
w_i^{(t)} = c_i \, \frac{\rho'\!\left(r_i^{(t)}\right)}{r_i^{(t)}}, 
\qquad
r_i^{(t)} = \| p_i^{k} - \mathbf{T}^{(t)} p_i^{k+1} \|_2
\end{equation}

\subsection{Tracking}\label{subsec:track_merging}

Accurate pixel correspondences are critical for the subsequent global bundle adjustment, but naive pairwise matching is of quadratic complexity. To achieve scalable tracking, for each subset $\mathcal{S}_k$, we build a sparse $k$-NN graph over the frames using the similarity matrix $\mathbf{M}$. For each retained edge $(i,j)$ between images $I_i$ and $I_j$, we extract SuperPoint \cite{detone2018superpoint} features and match them with LightGlue \cite{lindenberger2023lightglue}. To avoid redundant matching, if $(i,j)$ has already been matched, we skip $(j,i)$ and select the next nearest neighbor instead. 

Feature matching models like LightGlue are prone to false correspondences. We mitigate this by lifting the raw matches into 3D and filtering them via geometric consistency checks. Specifically, we unproject the raw correspondences $\{(x^i_{m,n}, x^j_{u,v})\}$ into 3D via the per-pixel depth map $\mathbf{D}_i$, then reprojecting them into the paired view ($x_{m,n}^i\to I_j$ and $x_{u,v}^j\to I_i$) via the known intrinsics $\mathbf{K}_i$, $\mathbf{K}_j$ and poses $(\mathbf{R}_i, \mathbf{t}_i)$, $(\mathbf{R}_j, \mathbf{t}_j)$. Matches with bidirectional reprojection error exceeding $\tau_{\text{reproj}}$ are discarded. 

The remaining correspondences $\{(x_{m,n}^i, x_{u,v}^j)\}$ are merged using a disjoint-set union to form multi-view tracks $T_l=(x^{l,i_1}_{m_1,n_1}, x^{l,i_2}_{m_2,n_2}, \dots)$. Each track $T_l$ 's 3D location and confidence is obtained by 
\begin{equation}
    \mathbf{x}_l=\frac{\sum_{k=1}^{K_l}\mathbf{C}_{i_k}[m_k,n_k]\mathbf{x}^{l,i_k}_{m_k,n_k}}{\sum_{k=1}^{K_l}\mathbf{C}_{i_k}[m_k,n_k]},
\end{equation}

\begin{equation}
    C_l=\frac{\sum_{k=1}^{K_l}\mathbf{C}_{i_k}[m_k,n_k]}{K_l},
\end{equation}

where $\mathbf{x}^{l,i_k}_{m_k,n_k}$ is the corresponding 3D point for each pixel. This final set of tracks $\mathcal{T}=\{(T_l, \mathbf{x}_l,C_l)\}$ is used in global bundle adjustment. The number of all LightGlue matchings scales linearly as $\mathcal{O}(kN)$ with the number of images, enabling efficient scaling to large-scale scenes. With fixed number of keypoints, this ensures an overall linearly scalability for tracking. 
\subsection{Global Bundle Adjustment}\label{subsec:global_ba}

To maintain global consistency after alignment and further enhance 3D reconstruction quality, we introduce an efficient global bundle adjustment step to jointly optimize the camera intrinsics, extrinsics and 3D point positions. We operate on the combined multi-view tracks $\mathcal{T}$ (Sec.~\ref{subsec:track_merging}), leveraging reliable 3D priors and correspondences. The optimization minimizes the confidence-weighted 2D reprojection error of all 3D points across cameras via gradient descent for $\nu$ iterations: 

\begin{equation}
    \mathbf{R}^*,\mathbf{t}^*,\mathbf{K}^*,\mathbf{P}^*=\arg\min_{\mathbf{R},\mathbf{t},\mathbf{K},\mathbf{P}}\mathcal{L_\text{BA}},
\end{equation}

\begin{equation}
    \mathcal{L_\text{BA}}=\sum_{(T_l,\mathbf{x}_l, C_l)\in\mathcal{T}}C_l\sum_{y^{l,i}\in T_l}||y^{l,i}-\pi_i(\mathbf{x}_l)||_2^\lambda,
\end{equation}

where $\pi_i: \mathbb{R}^3\to\mathbb{R}^2$ is the projection of 3D points onto the image plane of $I_i$ and $\lambda=0.5$. 

MASt3R-SfM \cite{duisterhof2025mast3r}, which also leverages gradient-based refinement, optimizes over image pairs. However, this limits global consistency, accuracy, and scalability when applied to a large number of views. 

\section{Experiments}

\subsection{Experimental Setup}

We evaluate our method on several datasets, including 7-Scenes~\cite{shotton2013scene}, Tanks and Temples (T\&T)~\cite{knapitsch2017tanks}, Cambridge Landmarks~\cite{kendall2015posenet}, and NRGBD~\cite{azinovic2022neural}.
Unlike prior work, we use all images in each scene without subsampling unless otherwise specified.
Our approach achieves performance that is comparable to or better than existing baselines in both accuracy and scalability.
Following standard practice, we report camera pose estimation results on 7-Scenes, T\&T, and Cambridge Landmarks, and evaluate point cloud quality on 7-Scenes and NRGBD.
 We compare against strong baselines including VGGT*~\cite{wang2025vggt}, Pi-3~\cite{wang2025pi}, FastVGGT~\cite{shen2025fastvggt}, VGGT-Long~\cite{deng2025vggt}, MaST3R-SfM~\cite{leroy2024grounding}, CUT3R~\cite{wang2024dust3r}, and TTT3R~\cite{chen2025ttt3r}. Note that VGGT* refers to the VRAM-efficient version of VGGT introduced in \cite{shen2025fastvggt}. 
All experiments are conducted on a single AMD Instinct MI210 GPU (64 GB) to measure runtime and memory consumption.
To make a fair comparison, we use the images of our pseduo-video as input to VGGT.

\subsection{Camera Pose Estimation}

We evaluate the predicted camera poses against the provided ground-truth camera poses. As done in DuST3R~\cite{wang2024dust3r}, we compute pairwise angular errors of the relative motions and translation on the 7-Scenes dataset, yielding the Relative Rotation Accuracy (RRA) and the Relative Translation Accuracy (RTA) at a given threshold $\tau$ (e.g. RRA@$\tau$). 
To demonstrate our scalability on a large number of input views, we evaluate the models on 500 images (stride = 2) and 1,000 images (no subsampling). As shown in Table \ref{tab:seven_scenes_pose}, our method maintains or surpasses the accuracy of base models such as VGGT*, FastVGGT and $\pi^3$ while outperforming other baselines such as MAStR-SfM, CUT3R, TTT3R and VGGT-Long. On 1,000 image sequences, we achieve the best accuracy while requiring a substantially reduced memory overhead compared to the base models as shown in Figure~\ref{fig:memory_comparison}. 
These results demonstrate the robustness, scalability, and general applicability of our approach to large-scale image collections. 

Similar to prior work~\cite{duisterhof2025mast3r}, we also report distance-based pose metrics: Absolute Trajectory Error (ATE), Relative Pose Error w.r.t rotation (RRE), and Relative Pose Error w.r.t. translation (RTE) on Tanks \& Temples and Cambridge Landmarks datasets. 
Predicted camera trajectories are aligned with the ground truth using Umeyama algorithm~\cite{umeyama2002least} before error computation. As shown in Table \ref{tab:tnt_lm_pose} and Figure~\ref{fig:camera_pose_vis}, our method achieves the best overall performance among all baselines and exhibits strong robustness on challenging outdoor scenes where many existing models struggle. Additionally, we compare our method with two recent, efficiency focused, COLMAP \cite{schoenberger2016sfm} based traditional SfM techniques, GLOMAP \cite{pan2024glomap} and InstantSfM \cite{zhong2025instantsfm}, with the results presented in Table~\ref{tab:camera_pose_colmap}  below on 7-Scenes. 

\begin{table}[h]
    \centering

\setlength{\tabcolsep}{2pt}
\caption{\small Camera pose estimation results on 7-Scenes.}

\label{tab:seven_scenes_pose}

\resizebox{0.9\linewidth}{!}{
\begin{tabular}{l|ccc|ccc}
\toprule
\multirow{2}{*}{Method} & \multicolumn{3}{c|}{7-Scenes, 500 Images} & \multicolumn{3}{c}{7-Scenes, 1000 Images} \\
\cline{2-7}
& RRA@30$\uparrow$ & RTA@30$\uparrow$ & AUC@30$\uparrow$ & RRA@30$\uparrow$ & RTA@30$\uparrow$ & AUC@30$\uparrow$ \\
\hline
MAStR-SfM\cite{duisterhof2025mast3r} & 100 & 97.30 & 79.47 & OOM & OOM & OOM \\
CUT3R \cite{wang2025continuous}& \underline{75.89} & 40.16 & 38.82 & 60.47 & 30.50 & 14.11 \\
TTT3R \cite{chen2025ttt3r}& 100 & 86.55 & 57.44 & \underline{97.19} & 53.69 & 30.95 \\
VGGT* \cite{wang2025vggt}& 100 & 96.87 & 81.13& OOM & OOM & OOM \\
FastVGGT \cite{shen2025fastvggt}& \textbf{100} & 96.75 & 80.59& OOM & OOM & OOM \\
VGGT-Long \cite{deng2025vggt}& 100 & 97.24 & 79.51& 100 & 95.54 & 75.11\\
$\pi^3$ \cite{wang2025pi} & \textbf{100} & \textbf{97.74} & \textbf{83.89}& OOM & OOM & OOM \\
\hline
\textbf{Ours + VGGT*} & 100 & 97.65 & 82.41& 100 & 97.42 & 82.20\\
\textbf{Ours + FastVGGT} & 100 & \underline{97.71} & 81.76& 100 & \underline{97.56} & 81.45\\
\textbf{Ours} + $\pi^3$ & \textbf{100} & \textbf{97.74} & 82.97& \textbf{100} & \textbf{97.69} & \textbf{83.63}\\
\bottomrule
\end{tabular}
}

\end{table}

\begin{table}[h]
    \centering

\caption{\small Camera pose estimation on the T\&T and Cambridge Landmarks datasets.}

\label{tab:tnt_lm_pose}

\resizebox{0.8\linewidth}{!}{
\begin{tabular}{l|ccc|ccc}
\toprule
\multirow{2}{*}{Method} & \multicolumn{3}{c|}{T\&T} & \multicolumn{3}{c}{Cambridge Landmarks} \\
\cline{2-7}
& ATE$\downarrow$ & RRE$\downarrow$ & RTE$\downarrow$ & ATE$\downarrow$ & RRE$\downarrow$ & RTE$\downarrow$ \\
\hline
MAStR-SfM \cite{duisterhof2025mast3r} & 0.202 & 0.521 & 0.024 & 7.695 & 4.426 & \underline{0.987} \\
CUT3R \cite{wang2025continuous}& 1.575 & 1.217 & 0.087 & 16.645 & 4.436 & 1.212 \\
TTT3R \cite{chen2025ttt3r} & 0.951 & 2.025 & 0.090 & 7.162 & 4.856 & 1.419 \\
VGGT* \cite{wang2025vggt} & 0.535 & 1.498 & 0.071 & 0.793 & 5.728 & 1.085 \\
FastVGGT \cite{shen2025fastvggt} & 0.549 & 1.400 & 0.072 & 0.812 & \underline{4.198} & 1.095 \\
VGGT-Long\cite{deng2025vggt} & 0.585 & 0.768 & 0.057 & 0.970 & \textbf{4.176} & \textbf{0.971} \\
$\pi^3$ \cite{wang2025pi} & \underline{0.090} & \underline{0.229} & \underline{0.025} & 1.630 & 4.416 & 1.109 \\
\hline
\textbf{Ours + VGGT*} & 0.522 & 1.563 & 0.026 & \textbf{0.661} & 4.482 & 1.215 \\
\textbf{Ours + FastVGGT} & 0.527 & 1.598 & 0.061 & \underline{0.780} & 4.433 & 1.261 \\
\textbf{Ours} + $\pi^3$ & \textbf{0.077} & \textbf{0.178} & \textbf{0.013} & 1.022 & 4.795 & 1.552 \\

\bottomrule
\end{tabular}
}

\end{table}

\label{tab:camera_pose_colmap}

\begin{table}[h]
    \vspace{-5pt}
    \caption{Camera pose estimation results with classical baselines on 7-Scenes (500 images). }
    \label{tab:camera_pose_colmap}

    \centering
    \setlength{\tabcolsep}{2pt}
    \resizebox{0.7\linewidth}{!}{
    \begin{tabular}{l|cccc}
    \hline
    Method & RRA@30$\uparrow$ & RTA@30$\uparrow$ & AUC@30$\uparrow$ & Time \\
    \hline
    GLOMAP \cite{pan2024glomap}  & 100 & 96.67 & 81.38 & 10 min 34 s \\
    InstantSfM \cite{zhong2025instantsfm} & 100 & 95.55 & 76.32 & 8 min 56 s \\
    Ours + $\pi^3$ & {100} & \textbf{97.61} & \textbf{83.31} & \textbf{4 min 48 s} \\
    \hline
    \end{tabular}
    }
\end{table}

\begin{figure}[htbp]
    \centering
    \setlength{\tabcolsep}{2pt}
    \renewcommand{\arraystretch}{1.0}
    \begin{tabular}{ccc}
        \includegraphics[width=0.30\columnwidth]{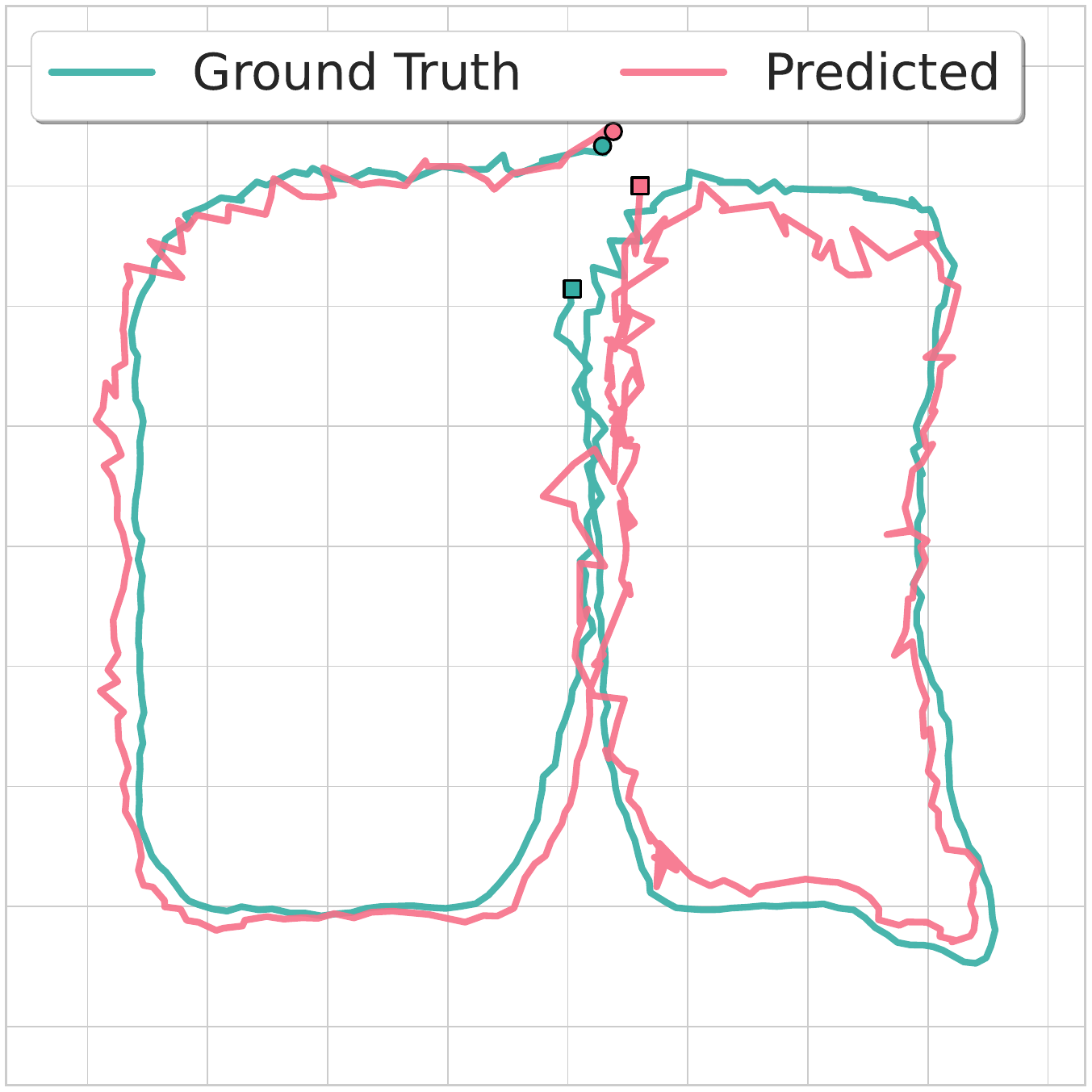} &
        \includegraphics[width=0.30\columnwidth]{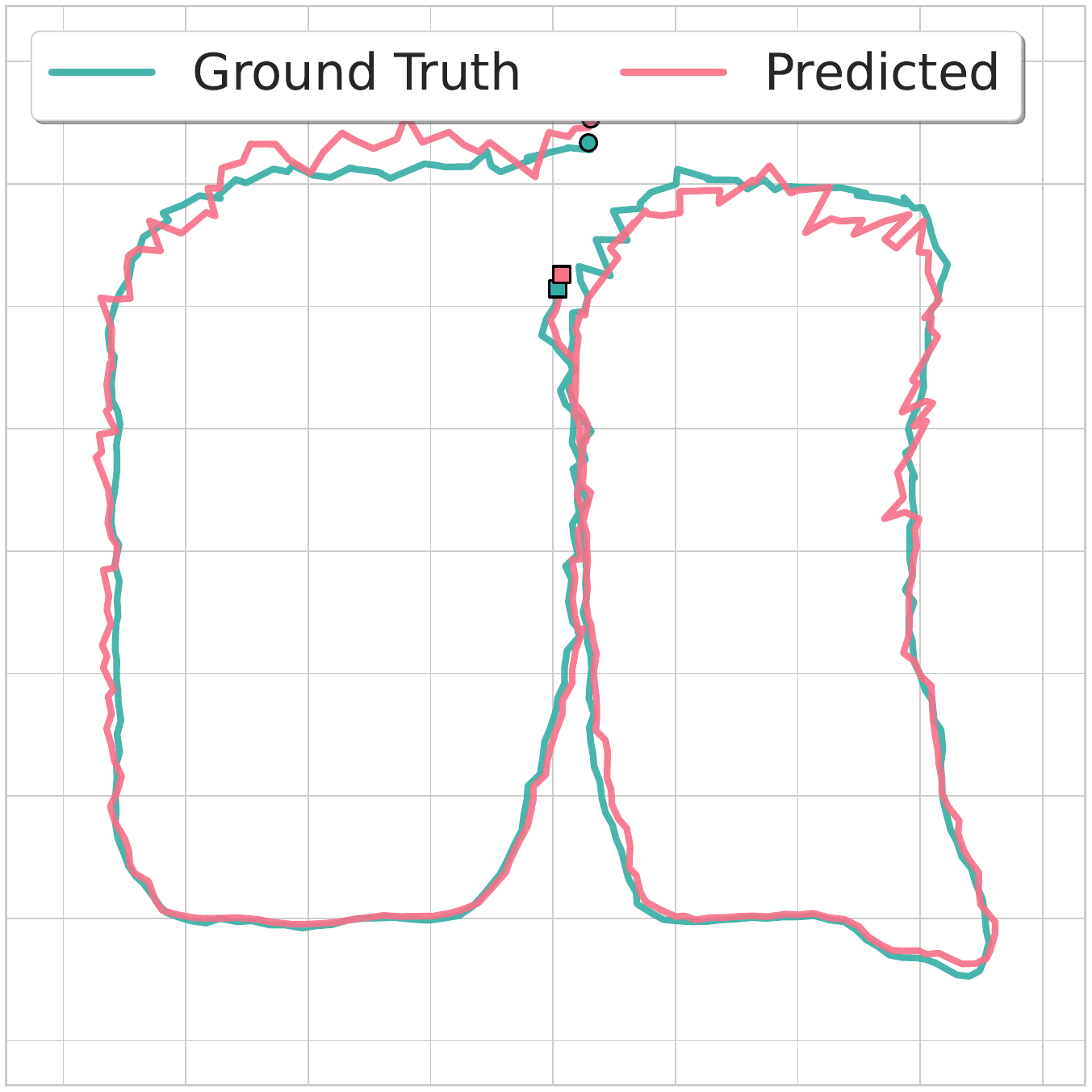} & 
        \includegraphics[width=0.30\columnwidth]{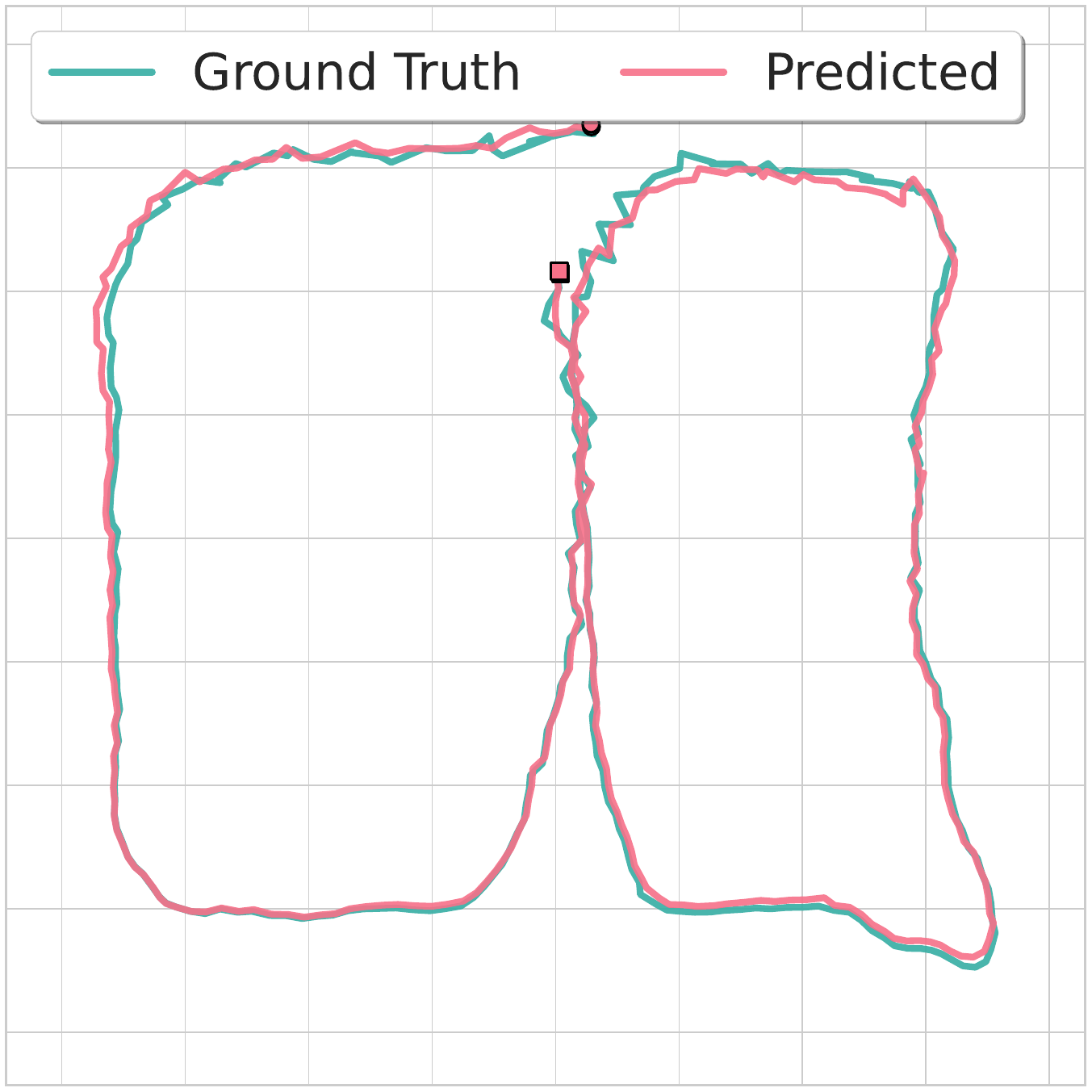}\\
        \small (a) TTT3R & \small (b) $\pi^3$ & \small  (c) Ours + $\pi^3$ \\[8pt]
    \end{tabular}
    \vspace{-0.5cm}
    \caption{\small Qualitative comparison of a predicted camera trajectory on a 300-image sequence from the Cambridge Landmarks dataset. Each subplot shows the estimated camera poses (red) overlaid with ground truth trajectories (green). Our method produces accurate and consistent trajectories, effectively handling long sequences with hundreds of frames.}
    \label{fig:camera_pose_vis}
\end{figure}

\subsection{Point Cloud Estimation}

\begin{figure}[htbp]
    \centering
    \includegraphics[width=0.9\linewidth]{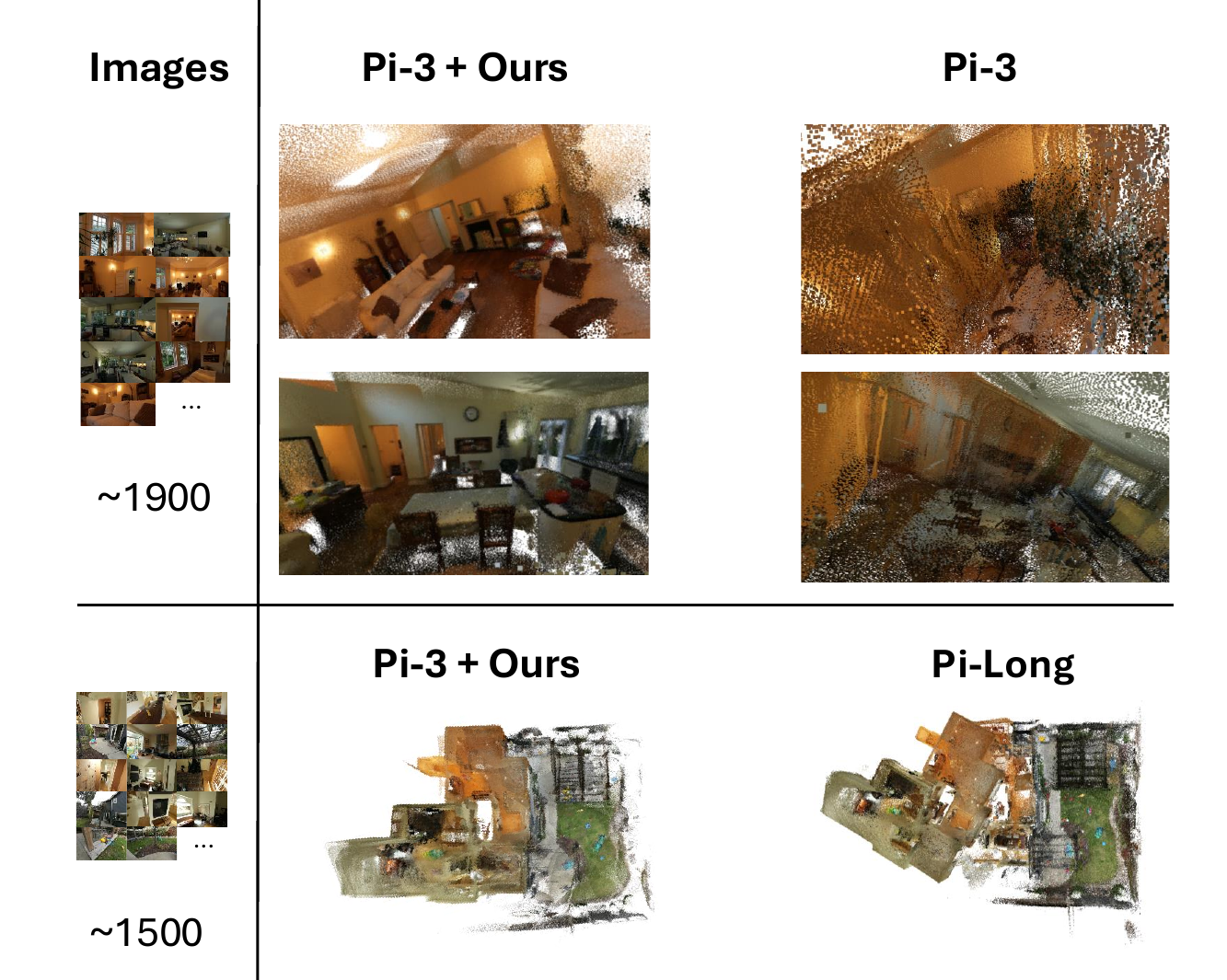}
    \caption{\small 
    \label{fig:zip_nerf_long_pointcloud} Qualitative comparison of 3D reconstructions on Zip-NERF \cite{barron2023zip}. For a fair comparison, the input to $\pi^3$ is subsampled to the same subset size with our method and the input ordering of sequence is the original one. Our method uses the same subset size as Pi-Long. }
\end{figure}

As in prior work~\cite{wang2025continuous,chen2025ttt3r,wang2025pi}, we evaluate the reconstructed point clouds for the 7-Scenes and NRGBD datasets. Keyframes are sampled with stride of 2 and 3 for 7-Scenes, and 3 and 5 for NRGBD. We report Accuracy (Acc.), Completion (Comp.) and Normal Consistency (N.C.) in Table \ref{tab:seven_scenes_nrgbd_points}. Unlike previous work, we are able to evaluate on a substantially larger number of input views. The performance of CUT3R and TTT3R degrades rapidly as the number of input images increases, while our method maintains consistently high accuracy and completeness across all scenes. Additionally, we compare our method on inputs exceeding 1{,}000 images against $\pi^3$. The input sequence for $\pi^3$ is subsampled to match the number of images used by each of the subset in our method. As shown in Fig.~\ref{fig:zip_nerf_long_pointcloud}, our approach preserves fine-grained details in both outdoor and indoor scenes, whereas $\pi^3$ shows noticeable degradation under subsampled inputs. Fig.~\ref{fig:zip_nerf_long_pointcloud} also demonstrates the qualitative comparison of our method with Pi-Long, which is a variant of VGGT-Long \cite{deng2025vggt}, using the same base model and subset size.

\begin{table}[htbp]
    \centering

\caption{\small Point cloud estimation on 7-Scenes and NRGBD.}

\label{tab:seven_scenes_nrgbd_points}

\resizebox{1.01\linewidth}{!}{
\scriptsize
\setlength{\tabcolsep}{2pt}
\begin{tabular}{l|c|cccccc|cccccc}

\toprule
\multirow{3}{*}{Method} & \multirow{3}{*}{Stride} & \multicolumn{6}{c|}{7-Scenes} & \multicolumn{6}{c}{NRGBD} \\
\cline{3-14}
& & \multicolumn{2}{c}{Acc$\downarrow$} & \multicolumn{2}{c}{Comp$\downarrow$} & \multicolumn{2}{c|}{NC$\uparrow$} & \multicolumn{2}{c}{Acc$\downarrow$} & \multicolumn{2}{c}{Comp$\downarrow$} & \multicolumn{2}{c}{NC$\uparrow$} \\
\cline{3-14}
& & Mean & Med. & Mean & Med. & Mean & Med. & Mean & Med. & Mean & Med. & Mean & Med. \\
\hline
CUT3R \cite{wang2025continuous} & 3/5 & 0.086 & 0.054 & 0.055 & 0.017 & 0.564 & 0.596 & 0.266 & 0.178 & 0.117 & 0.047 & 0.600 & 0.659 \\
TTT3R \cite{chen2025ttt3r}& 3/5 & 0.137 & 0.090 & 0.072 & 0.027 & 0.554 & 0.581 & 0.769 & 0.504 & 0.089 & 0.046 & 0.702 & 0.796 \\
VGGT$^*$ \cite{wang2025vggt}& 3/5 & 0.019 & 0.008 & 0.028 & 0.011 & \underline{0.607}& \underline{0.660}& 0.028 & 0.018 & 0.018 & 0.007 & 0.710 & 0.816 \\
FastVGGT \cite{shen2025fastvggt} & 3/5 & \textbf{0.014} & \underline{0.006}& 0.028 & 0.010 & \textbf{0.631} & \textbf{0.705} & 0.024 & 0.013 & 0.018 & 0.010 & 0.665 & 0.791 \\
$\pi^3$ \cite{wang2025pi} & 3/5 & 0.019 & \textbf{0.004} & 0.036 & 0.018 & 0.539 & 0.556 & \textbf{0.018}& \textbf{0.008}& \textbf{0.013}& \textbf{0.004}& 0.652 & 0.743 \\
VGGT-Long \cite{deng2025vggt} & 3/5 & 0.016 & 0.007 & 0.028 & 0.011 & 0.617 & 0.676 & 0.021 & 0.012 & 0.016 & 0.007 & \textbf{0.780} & \textbf{0.887} \\ 
\textbf{Ours+VGGT$^*$} & 3/5 & 0.018 & 0.007 & \underline{0.021}& 0.009 & 0.592 & 0.640 & 0.025 & 0.015 & \underline{0.015}& \underline{0.006} & 0.703 & 0.814 \\
\textbf{Ours+F.VGGT} & 3/5 & 0.018 & 0.008 & 0.023 &\textbf{0.007} & 0.576 & 0.613 & 0.032 & 0.02 & 0.016 & 0.007 & 0.676 & 0.776\\
\textbf{Ours+$\pi^3$} & 3/5 & \underline{0.017}& 0.007 & \textbf{0.019} & 0.009 & 0.592 & 0.640 & \underline{0.021}& \underline{0.011}& 0.016 & \underline{0.006} & \underline{0.736}& \underline{0.845}\\
\hline
CUT3R \cite{wang2025continuous} & 2/3 & 0.165 & 0.111 & 0.086 & 0.025 & 0.537 & 0.553 & 0.329 & 0.250 & 0.144 & 0.044 & 0.568 & 0.603 \\
TTT3R \cite{chen2025ttt3r} & 2/3 & 0.159 & 0.117 & 0.099 & 0.037 & 0.533 & 0.548 & 0.258 & 0.170 & 0.111 & 0.032 & 0.592 & 0.648 \\
VGGT$^*$ \cite{wang2025vggt} & 2/3 & 0.021 & 0.008 & 0.027 & 0.011 & \underline{0.604}& \underline{0.656}& 0.027 & 0.019 & 0.024 & 0.008 & {0.726} & {0.830} \\
FastVGGT \cite{shen2025fastvggt} & 2/3 & \textbf{0.014}& \textbf{0.007}& 0.028 & 0.010 & \textbf{0.630}& \textbf{0.702}& 0.032 & 0.020 & 0.020 & 0.010 & 0.662 & 0.788 \\
$\pi^3$ \cite{wang2025pi} & 2/3 & 0.030 & 0.010 & 0.055 & 0.055 & 0.525 & 0.535 & 0.064 & 0.042 & 0.038 & 0.018 & 0.600 & 0.656 \\
VGGT-Long \cite{deng2025vggt} & 3/5 & 0.016& \textbf{0.007} & 0.028 & 0.011 & 0.614 & 0.672 & \textbf{0.018} & 0.01 & 0.015 & \textbf{0.006} & \textbf{0.780} & \textbf{0.887} \\ 
\textbf{Ours+VGGT$^*$} & 2/3 & 0.020 & 0.008 & 0.024 & \underline{0.008}& 0.580 & 0.621 & \underline{0.020}& \underline{0.013}& \textbf{0.013}& \textbf{0.006}& 0.700 & 0.805 \\
\textbf{Ours+F.VGGT} & 2/3 &0.018 & 0.008 & \underline{0.022}& \textbf{0.007}& 0.573 & 0.61 & 0.026& 0.017 & \underline{0.014}& \underline{0.007} & 0.67 & 0.764 \\
\textbf{Ours+$\pi^3$} & 2/3 & \underline{0.016}& \textbf{0.007}& \textbf{0.017}& \underline{0.008}& 0.599 & 0.649 & \underline{0.020}& \textbf{0.011}& 0.015 & \textbf{0.006}& \underline{0.735}& \underline{0.843}\\
\bottomrule

\end{tabular}

}

\end{table}

\begin{figure}[htbp]
    \centering
    \includegraphics[width=\linewidth]{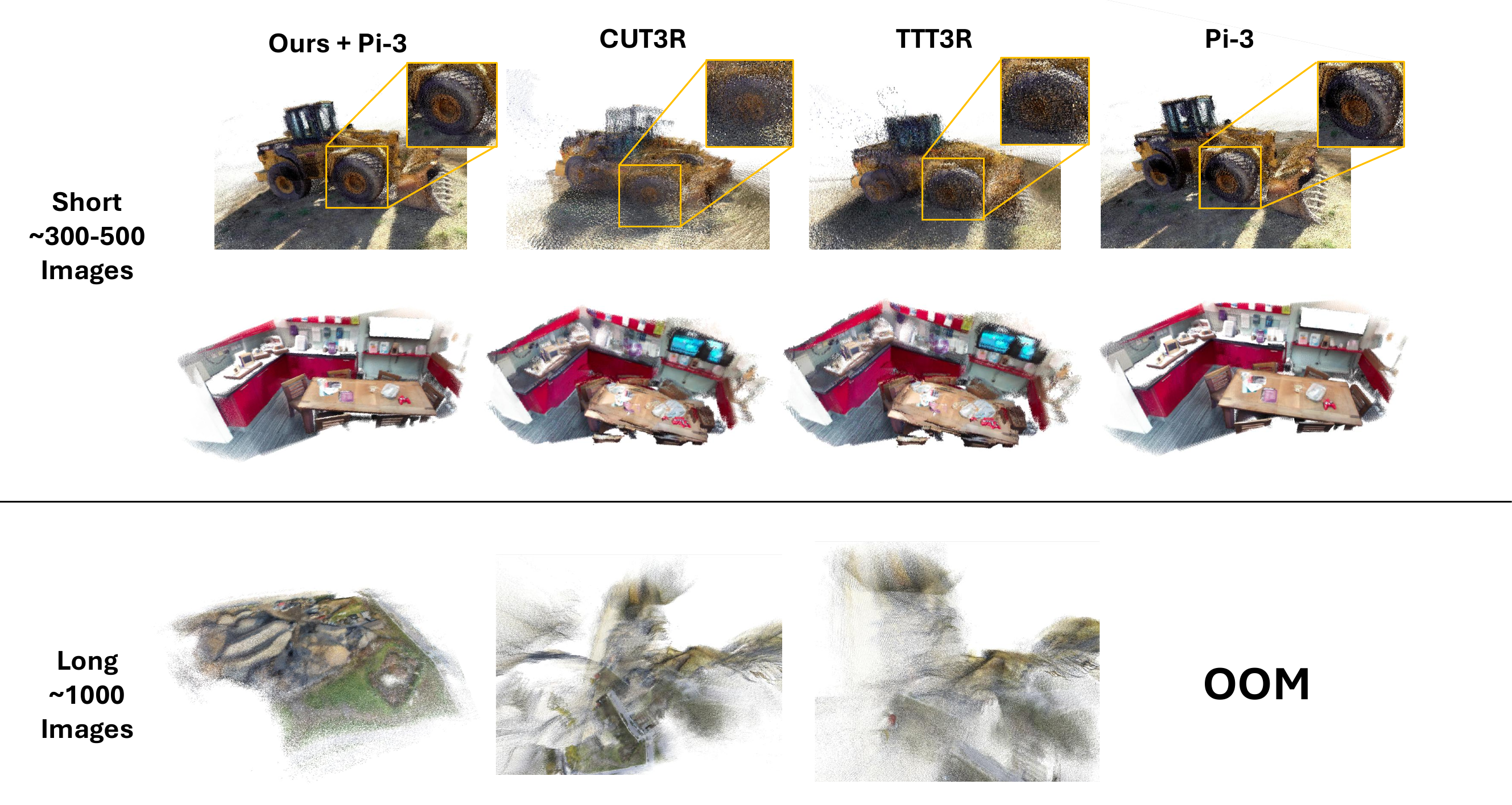}
    \caption{\small 
    \label{fig:point_cloud_vis} \textbf{Qualitative comparison of 3D reconstructions on short (appox. 300–500 images) and long (approx. 1,000 images) sequences.} Our method (Ours + Pi-3) produces sharper and more complete point clouds than CUT3R, TTT3R, and $\pi^3$. Competing methods fail or run out of memory (OOM) on long sequences, while ours remains stable. }
\end{figure}

\subsection{Computation Time and Memory Usage}
\begin{figure*}[htbp]
    \centering
    \begin{minipage}{0.42\textwidth}
        \centering
        \includegraphics[width=1.0\linewidth]{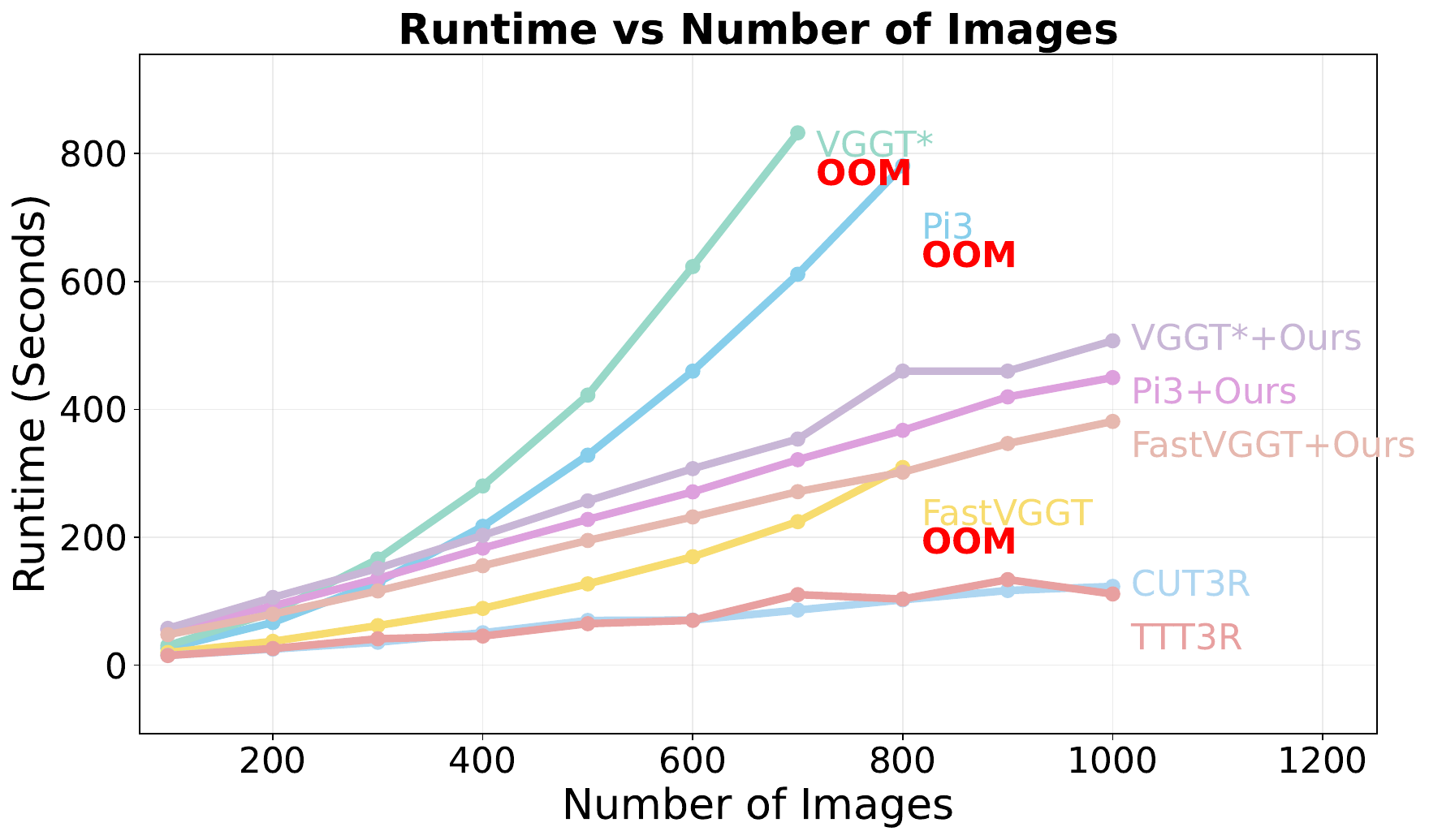}
        \caption*{\small \textbf{(a)} Runtime}
    \end{minipage}
    \hfill
    \begin{minipage}{0.42\textwidth}
        \centering
        \includegraphics[width=1.0\linewidth]{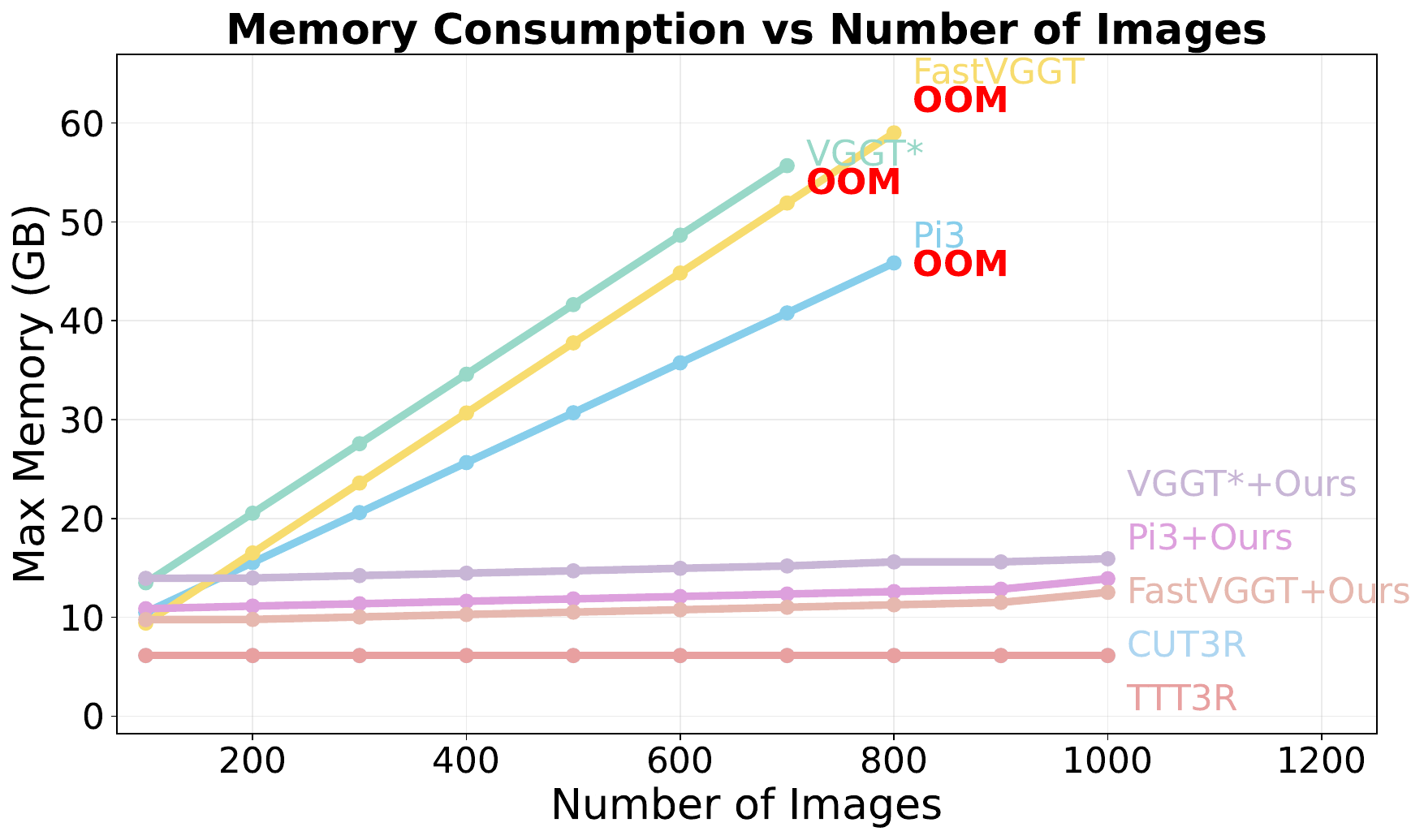}
        \caption*{\small \textbf{(b)} Peak GPU Memory}
    \end{minipage}
    \caption{\small Our method substantially reduces both runtime and memory consumption for base models processing long inputs.}
    \label{fig:memory_comparison}
\end{figure*}

To quantitatively evaluate the scalability of our model, we collect the computation time and memory usage for all models on one scene of the 7-Scenes dataset. As shown in Figure~\ref{fig:memory_comparison}, we plot the memory consumption in GB and computation latency in seconds for varying amounts of input images (100 to 1,000). 
We demonstrate that our method can achieve significantly improved memory and compute efficiency over the base models (e.g. VGGT and $\pi^3$). We note that memory consumption with our approach stays stable for any size of the input data set. 

\subsection{Ablation Study}
All the following experiments are conducted on all scenes of the Tanks \& Temples dataset. 

\noindent\textbf{Sequential input vs. unordered input.}
Table \ref{tab:ablation_seq} compares model performance when using the ground-truth video ordering versus our pseudo-video ordering derived from unordered images. Across all methods, ATE remains nearly identical (differences $\leq$ 0.001), indicating that our ordering algorithm preserves the global scene trajectory. 
Relative pose metrics show only minor variations, and for $\pi^3$, the pseudo ordering even produces slightly better results. The discrepancies between true and pseudo orderings are negligible, showing that our algorithm reconstructs a sequence as informative as the original video. Even with arbitrarily shuffled images, it recovers geometry and motion with comparable accuracy.

\begin{table}[h]
    \centering

\caption{\small Effect of the sequential input vs. unordered input.}

\label{tab:ablation_seq}

\resizebox{0.9\linewidth}{!}{
\begin{tabular}{l|ccc|ccc}
\toprule
\multirow{2}{*}{Method} & \multicolumn{3}{c|}{w/ real video seq.} & \multicolumn{3}{c}{w/ pseudo seq.} \\
\cline{2-7}
& ATE$\downarrow$ & RRE$\downarrow$ & RTE$\downarrow$ & ATE$\downarrow$ & RRE$\downarrow$ & RTE$\downarrow$ \\
\hline
VGGT~\cite{wang2025vggt} & 0.521 & 1.317 & 0.544 & 0.522 & 1.563 & 0.061 \\
FastVGGT~\cite{shen2025fastvggt} & 0.525 & 1.364 & 0.056 & 0.526 & 1.598 & 0.079\\
$\pi^3$~\cite{wang2025pi} & \textbf{0.077} & 0.196 & 0.014 & \textbf{0.077} &\textbf{ 0.178} & \textbf{0.013}\\

\bottomrule
\end{tabular}
}

\end{table}

\noindent\textbf{The image set splitting strategy.}
Table \ref{tab:ablation_splitting} evaluates three strategies for partitioning the ordered image sequence into subsets for local reconstruction. 
The graph-based clustering approach which groups images using similarity scores derived from DINO features, performs worst across all metrics, suggesting that geometric reconstruction models require more than feature-level similarity and benefits from diverse multi-view observations. 
The sliding-window strategy performs better, particularly in ATE and RTE, but it inherently restricts each subset to a narrow portion of the trajectory. 
Consequently, some subsets may see only a single façade of the scene (e.g., the front of a building but not the sides or back), limiting their ability to estimate reliable relative poses and contributing to poorer RRE performance. In contrast, our interleaving strategy achieves the strongest results by a clear margin across all metrics. 
By distributing views from the entire trajectory into every subset, it ensures broad and varied perspective coverage for each local reconstruction. 
These findings underscore the effectiveness of our splitting method and its importance for robust multi-view reconstruction.
\begin{table}[ht]
    \centering

\caption{\small Effect of the image set splitting strategy for camera pose accuracy.}

\label{tab:ablation_splitting}

\resizebox{0.7\linewidth}{!}{
\begin{tabular}{l|ccc}
\toprule
Splitting Method & ATE$\downarrow$ & RRE$\downarrow$ & RTE$\downarrow$ \\
\hline
Graph clustering & 0.319 & 1.280 & 0.060 \\
Sliding window & 0.106 & 0.492 & 0.024 \\
Ours & \textbf{0.077} & \textbf{0.178} & \textbf{0.013} \\
\bottomrule
\end{tabular}
}

\end{table}

\noindent\textbf{The hyperparameters.} We investigate the effect of different split sizes and overlaps in the partitioning step (Sec. \ref{subsec:ordering_and_partitioning}). Theoretically, larger subsets increase both inference time and memory usage due to the higher computational cost of the geometric foundation model. 
Results in Table \ref{tab:hyperparams_analysis} empirically confirm this trend. 
As expected, larger subsets generally yield better accuracy because they provide more diverse multi-view observations. 
We note that a split size of 100 images already achieves peak performance, indicating that our framework is robust even on GPUs with limited memory. We therefore use a split size of 100 as a practical balance between reconstruction quality and efficiency. As shown in Table \ref{tab:hyperparams_analysis}, camera pose estimation quality shows little sensitivity to the overlap size; accordingly, we set the overlap size to 5.

\begin{table}[ht]
\centering
\caption{\small Effect of subset size (left) and overlap size (right).}
\label{tab:hyperparams_analysis}

\resizebox{0.95\linewidth}{!}{
\begin{tabular}{c|ccccc||c|ccc}
\toprule
\multicolumn{6}{c||}{Subset Size} &
\multicolumn{4}{c}{Overlap Size} \\

\cmidrule(r){1-6} \cmidrule(l){7-10}

Split Size & ATE$\downarrow$ & RRE$\downarrow$ & RTE$\downarrow$ & Time (s) & Mem. (GB) &
Overlap & ATE$\downarrow$ & RRE$\downarrow$ & RTE$\downarrow$ \\

\hline

25  & 0.084 & 0.674 & 0.026 & 127.41 & 10.95 & 3  & 0.075 & 0.198 & 0.017 \\
50  & 0.078 & 0.415 & 0.016 & 127.20 & 10.63 & 5  & 0.076 & 0.175 & 0.013 \\
75  & 0.076 & 0.180 & 0.014 & 132.43 & 10.98 & 7  & 0.076 & 0.171 & 0.012 \\
100 & 0.076 & 0.177 & 0.013 & 136.51 & 11.63 & 10 & 0.076 & 0.167 & 0.012 \\
125 & 0.075 & 0.192 & 0.016 & 139.63 & 12.50 & 15 & 0.076 & 0.162 & 0.011 \\
150 & 0.076 & 0.166 & 0.013 & 144.61 & 12.70 &    &       &       &  \\
200 & 0.077 & 0.170 & 0.012 & 155.39 & 14.31 &    &       &       &  \\

\bottomrule
\end{tabular}
}
\end{table}

\noindent\textbf{Effect of different tracking and bundle adjustment methods.} 
We analyze the impact of different tracking and bundle adjustment (BA) configurations. ``w/o BA'' refers to the pipeline without a global bundle adjustment step. 
``BA w/ VGGT* Tracking uses VGGT’s tracking module  which directly merges tracks across subsets via nearest-neighbor point matching. 
``BA w/ LightGlue (pseudo video)'' applies pairwise LightGlue matching on adjacent frames of the pseudo-video sequence with skip connections. 
``BA w/ LightGlue (graph)'' corresponds to our proposed graph-based method described in Sec. \ref{subsec:track_merging}.

These experiments use VGGT as the geometric foundation model with a subset size of 50 to avoid the OOM issue. 
Together with its already high memory footprint, this shows that VGGT’s iterative transformer-based tracking is not well suited for large-scale BA and reconstruction. 
The results further demonstrate that our graph-based matching captures a broader range of inter-image correspondences, leading to improved geometric consistency across views.

\vspace{-1em}
\begin{table}[ht]
    \centering

\caption{\small Effect of the bundle adjustment method.}

\label{tab:ablation_ba}

\resizebox{0.9\linewidth}{!}{
\begin{tabular}{l|ccccc}
\toprule
Variant & ATE$\downarrow$ & RRE$\downarrow$ & RTE$\downarrow$ & Time & Mem. \\
\hline
w/o BA & 0.551 & 2.284 & 0.119 & 62.86 & 12.58 \\
BA w/ VGGT Tracking~\cite{wang2025vggt} & 0.602 & 2.659 & 0.155 & 137.68 & 33.22 \\
BA w/ LightGlue (pseudo video)~\cite{lindenberger2023lightglue}  & 0.526 & 1.572 & 0.061 & 140.53 & 12.58 \\
BA w/ LightGlue (graph)~\cite{lindenberger2023lightglue}  & \textbf{0.522} & \textbf{1.563} & \textbf{0.060} & 141.08 & 12.68 \\
\bottomrule
\end{tabular}
}

\end{table}

\vspace{-2em}

\section{Conclusion}
We introduced \name{}, a scalable framework that enables geometric foundation models to reconstruct large, unordered image collections far beyond their native memory limits. By combining principled clustering, efficient global tracking, and confidence-weighted bundle adjustment, \name{} achieves high global accuracy while keeping memory usage bounded. Our divide-and-conquer mechanism also enables faster execution times and distribution across multiple GPUs. Our approach demonstrates the strength of merging traditional geometric optimization with modern neural geometry models. This scalability opens the door to developing more complex neural geometry models without being constrained by GPU capacity and reduces reliance on powerful hardware, making 3D reconstruction more accessible, robust, and broadly deployable. 

\section{Acknowledgments}
We gratefully acknowledge the Sony Research Award program and NSERC for supporting this project. We also thank the members of the embARC and DGP research group at the University of Toronto for all the feedback and discussions.

{
    \small
    \bibliographystyle{ieeenat_fullname}
    \bibliography{main}

\begin{thebibliography}{43}
\providecommand{\natexlab}[1]{#1}
\providecommand{\url}[1]{\texttt{#1}}
\expandafter\ifx\csname urlstyle\endcsname\relax
  \providecommand{\doi}[1]{doi: #1}\else
  \providecommand{\doi}{doi: \begingroup \urlstyle{rm}\Url}\fi

\bibitem[Agarwal et~al.(2011)Agarwal, Furukawa, Snavely, Simon, Curless, Seitz,
  and Szeliski]{agarwal2011building}
Sameer Agarwal, Yasutaka Furukawa, Noah Snavely, Ian Simon, Brian Curless,
  Steven~M Seitz, and Richard Szeliski.
\newblock Building rome in a day.
\newblock \emph{Communications of the ACM}, 54\penalty0 (10):\penalty0
  105--112, 2011.

\bibitem[{Apple Inc.}(2017)]{applearkit}
{Apple Inc.}
\newblock Arkit.
\newblock https://developer.apple.com/augmented-reality/arkit/, 2017.
\newblock Accessed: 2024-06-30.

\bibitem[Azinovi{\'c} et~al.(2022)Azinovi{\'c}, Martin-Brualla, Goldman,
  Nie{\ss}ner, and Thies]{azinovic2022neural}
Dejan Azinovi{\'c}, Ricardo Martin-Brualla, Dan~B Goldman, Matthias
  Nie{\ss}ner, and Justus Thies.
\newblock Neural rgb-d surface reconstruction.
\newblock In \emph{Proceedings of the IEEE/CVF Conference on Computer Vision
  and Pattern Recognition}, 2022.

\bibitem[Barron et~al.(2023)Barron, Mildenhall, Verbin, Srinivasan, and
  Hedman]{barron2023zip}
Jonathan~T Barron, Ben Mildenhall, Dor Verbin, Pratul~P Srinivasan, and Peter
  Hedman.
\newblock Zip-nerf: Anti-aliased grid-based neural radiance fields.
\newblock In \emph{Proceedings of the IEEE/CVF International Conference on
  Computer Vision}, pages 19697--19705, 2023.

\bibitem[Bhowmick et~al.(2014)Bhowmick, Patra, Chatterjee, Govindu, and
  Banerjee]{bhowmick2014divide}
Brojeshwar Bhowmick, Suvam Patra, Avishek Chatterjee, Venu~Madhav Govindu, and
  Subhashis Banerjee.
\newblock Divide and conquer: Efficient large-scale structure from motion using
  graph partitioning.
\newblock In \emph{Asian Conference on Computer Vision}, pages 273--287.
  Springer, 2014.

\bibitem[Chen et~al.(2025)Chen, Chen, Xiu, Geiger, and Chen]{chen2025ttt3r}
Xingyu Chen, Yue Chen, Yuliang Xiu, Andreas Geiger, and Anpei Chen.
\newblock Ttt3r: 3d reconstruction as test-time training.
\newblock \emph{arXiv preprint arXiv:2509.26645}, 2025.

\bibitem[Chen et~al.(2020)Chen, Shen, Chen, and Wang]{chen2020graph}
Yu Chen, Shuhan Shen, Yisong Chen, and Guoping Wang.
\newblock Graph-based parallel large scale structure from motion.
\newblock \emph{Pattern Recognition}, 107:\penalty0 107537, 2020.

\bibitem[Deng et~al.(2025)Deng, Ti, Xu, Yang, and Xie]{deng2025vggt}
Kai Deng, Zexin Ti, Jiawei Xu, Jian Yang, and Jin Xie.
\newblock Vggt-long: Chunk it, loop it, align it--pushing vggt's limits on
  kilometer-scale long rgb sequences.
\newblock \emph{arXiv preprint arXiv:2507.16443}, 2025.

\bibitem[DeTone et~al.(2018)DeTone, Malisiewicz, and
  Rabinovich]{detone2018superpoint}
Daniel DeTone, Tomasz Malisiewicz, and Andrew Rabinovich.
\newblock Superpoint: Self-supervised interest point detection and description,
  2018.

\bibitem[Duisterhof et~al.(2025)Duisterhof, Zust, Weinzaepfel, Leroy, Cabon,
  and Revaud]{duisterhof2025mast3r}
Bardienus~Pieter Duisterhof, Lojze Zust, Philippe Weinzaepfel, Vincent Leroy,
  Yohann Cabon, and Jerome Revaud.
\newblock Mast3r-sfm: a fully-integrated solution for unconstrained
  structure-from-motion.
\newblock In \emph{2025 International Conference on 3D Vision (3DV)}, pages
  1--10. IEEE, 2025.

\bibitem[Furukawa et~al.(2015)Furukawa, Hern{\'a}ndez,
  et~al.]{furukawa2015multi}
Yasutaka Furukawa, Carlos Hern{\'a}ndez, et~al.
\newblock Multi-view stereo: A tutorial.
\newblock \emph{Foundations and trends{\textregistered} in Computer Graphics
  and Vision}, 9\penalty0 (1-2):\penalty0 1--148, 2015.

\bibitem[Gu et~al.(2020)Gu, Fan, Zhu, Dai, Tan, and Tan]{gu2020cascade}
Xiaodong Gu, Zhiwen Fan, Siyu Zhu, Zuozhuo Dai, Feitong Tan, and Ping Tan.
\newblock Cascade cost volume for high-resolution multi-view stereo and stereo
  matching.
\newblock In \emph{Proceedings of the IEEE/CVF conference on computer vision
  and pattern recognition}, pages 2495--2504, 2020.

\bibitem[Keetha et~al.(2025)Keetha, M\"uller, Sch\"onberger, Porzi, Zhang,
  Fischer, Knapitsch, Zauss, Weber, Antunes, Luiten, Lopez-Antequera, Bul\`o,
  Richardt, Ramanan, Scherer, and Kontschieder]{keetha2025mapanything}
Nikhil Keetha, Norman M\"uller, Johannes Sch\"onberger, Lorenzo Porzi, Yuchen
  Zhang, Tobias Fischer, Arno Knapitsch, Duncan Zauss, Ethan Weber, Nelson
  Antunes, Jonathon Luiten, Manuel Lopez-Antequera, Samuel~Rota Bul\`o,
  Christian Richardt, Deva Ramanan, Sebastian Scherer, and Peter Kontschieder.
\newblock {MapAnything}: Universal feed-forward metric {3D} reconstruction.
\newblock In \emph{arXiv:2509.13414}, 2025.

\bibitem[Kendall et~al.(2015)Kendall, Grimes, and Cipolla]{kendall2015posenet}
Alex Kendall, Matthew Grimes, and Roberto Cipolla.
\newblock Posenet: A convolutional network for real-time 6-dof camera
  relocalization.
\newblock In \emph{Proceedings of the IEEE international conference on computer
  vision}, pages 2938--2946, 2015.

\bibitem[Kerbl et~al.(2023)Kerbl, Kopanas, Leimk{\"u}hler, and
  Drettakis]{kerbl20233d}
Bernhard Kerbl, Georgios Kopanas, Thomas Leimk{\"u}hler, and George Drettakis.
\newblock 3d gaussian splatting for real-time radiance field rendering.
\newblock \emph{ACM Transactions on Graphics (ToG)}, 42\penalty0 (4):\penalty0
  1--14, 2023.

\bibitem[Knapitsch et~al.(2017)Knapitsch, Park, Zhou, and
  Koltun]{knapitsch2017tanks}
Arno Knapitsch, Jaesik Park, Qian-Yi Zhou, and Vladlen Koltun.
\newblock Tanks and temples: Benchmarking large-scale scene reconstruction.
\newblock \emph{ACM Transactions on Graphics (ToG)}, 36\penalty0 (4):\penalty0
  1--13, 2017.

\bibitem[Leroy et~al.(2024)Leroy, Cabon, and Revaud]{leroy2024grounding}
Vincent Leroy, Yohann Cabon, and J{\'e}r{\^o}me Revaud.
\newblock Grounding image matching in 3d with mast3r.
\newblock In \emph{European Conference on Computer Vision}, pages 71--91.
  Springer, 2024.

\bibitem[Lin et~al.(2022)Lin, Liu, Hu, Yan, Xie, and Huang]{UrbanScene3D}
Liqiang Lin, Yilin Liu, Yue Hu, Xingguang Yan, Ke Xie, and Hui Huang.
\newblock Capturing, reconstructing, and simulating: the urbanscene3d dataset.
\newblock In \emph{ECCV}, pages 93--109, 2022.

\bibitem[Lindenberger et~al.(2023)Lindenberger, Sarlin, and
  Pollefeys]{lindenberger2023lightglue}
Philipp Lindenberger, Paul-Edouard Sarlin, and Marc Pollefeys.
\newblock Lightglue: Local feature matching at light speed.
\newblock In \emph{Proceedings of the IEEE/CVF international conference on
  computer vision}, pages 17627--17638, 2023.

\bibitem[Mildenhall et~al.(2020)Mildenhall, Srinivasan, Tancik, Barron,
  Ramamoorthi, and Ng]{mildenhall2020nerf}
Ben Mildenhall, Pratul~P Srinivasan, Matthew Tancik, Jonathan~T Barron, Ravi
  Ramamoorthi, and Ren Ng.
\newblock Nerf: Representing scenes as neural radiance fields for view
  synthesis.
\newblock In \emph{European Conference on Computer Vision}, pages 405--421.
  Springer, 2020.

\bibitem[Mur-Artal et~al.(2015)Mur-Artal, Montiel, and Tardos]{mur2015orb}
Raul Mur-Artal, Jose Maria~Martinez Montiel, and Juan~D Tardos.
\newblock Orb-slam: A versatile and accurate monocular slam system.
\newblock \emph{IEEE transactions on robotics}, 31\penalty0 (5):\penalty0
  1147--1163, 2015.

\bibitem[Oquab et~al.(2023)Oquab, Darcet, Moutakanni, Vo, Szafraniec, Khalidov,
  Fernandez, Haziza, Massa, El-Nouby, et~al.]{oquab2023dinov2}
Maxime Oquab, Timoth{\'e}e Darcet, Th{\'e}o Moutakanni, Huy Vo, Marc
  Szafraniec, Vasil Khalidov, Pierre Fernandez, Daniel Haziza, Francisco Massa,
  Alaaeldin El-Nouby, et~al.
\newblock Dinov2: Learning robust visual features without supervision.
\newblock \emph{arXiv preprint arXiv:2304.07193}, 2023.

\bibitem[{\"O}zye{\c{s}}il et~al.(2017){\"O}zye{\c{s}}il, Voroninski, Basri,
  and Singer]{ozyecsil2017survey}
Onur {\"O}zye{\c{s}}il, Vladislav Voroninski, Ronen Basri, and Amit Singer.
\newblock A survey of structure from motion*.
\newblock \emph{Acta Numerica}, 26:\penalty0 305--364, 2017.

\bibitem[Pan et~al.(2024)Pan, Baráth, Pollefeys, and
  Sch\"{o}nberger]{pan2024glomap}
Linfei Pan, Dániel Baráth, Marc Pollefeys, and Johannes~Lutz Sch\"{o}nberger.
\newblock Global structure-from-motion revisited.
\newblock In \emph{European Conference on Computer Vision (ECCV)}, 2024.

\bibitem[Ranftl et~al.(2021)Ranftl, Bochkovskiy, and Koltun]{ranftl2021vision}
Ren{\'e} Ranftl, Alexey Bochkovskiy, and Vladlen Koltun.
\newblock Vision transformers for dense prediction.
\newblock In \emph{Proceedings of the IEEE/CVF international conference on
  computer vision}, pages 12179--12188, 2021.

\bibitem[Sch\"{o}nberger and Frahm(2016)]{schoenberger2016sfm}
Johannes~Lutz Sch\"{o}nberger and Jan-Michael Frahm.
\newblock Structure-from-motion revisited.
\newblock In \emph{Conference on Computer Vision and Pattern Recognition
  (CVPR)}, 2016.

\bibitem[Sch\"{o}nberger et~al.(2016)Sch\"{o}nberger, Zheng, Pollefeys, and
  Frahm]{schoenberger2016mvs}
Johannes~Lutz Sch\"{o}nberger, Enliang Zheng, Marc Pollefeys, and Jan-Michael
  Frahm.
\newblock Pixelwise view selection for unstructured multi-view stereo.
\newblock In \emph{European Conference on Computer Vision (ECCV)}, 2016.

\bibitem[Shen et~al.(2025)Shen, Zhang, Qu, and Cao]{shen2025fastvggt}
You Shen, Zhipeng Zhang, Yansong Qu, and Liujuan Cao.
\newblock Fastvggt: Training-free acceleration of visual geometry transformer.
\newblock \emph{arXiv preprint arXiv:2509.02560}, 2025.

\bibitem[Shotton et~al.(2013)Shotton, Glocker, Zach, Izadi, Criminisi, and
  Fitzgibbon]{shotton2013scene}
Jamie Shotton, Ben Glocker, Christopher Zach, Shahram Izadi, Antonio Criminisi,
  and Andrew Fitzgibbon.
\newblock Scene coordinate regression forests for camera relocalization in
  rgb-d images.
\newblock In \emph{Proceedings of the IEEE conference on computer vision and
  pattern recognition}, 2013.

\bibitem[Snavely et~al.(2006)Snavely, Seitz, and Szeliski]{snavely2006photo}
Noah Snavely, Steven~M Seitz, and Richard Szeliski.
\newblock Photo tourism: exploring photo collections in 3d.
\newblock In \emph{ACM siggraph 2006 papers}, pages 835--846. 2006.

\bibitem[Umeyama(2002)]{umeyama2002least}
Shinji Umeyama.
\newblock Least-squares estimation of transformation parameters between two
  point patterns.
\newblock \emph{IEEE Transactions on pattern analysis and machine
  intelligence}, 2002.

\bibitem[Vaswani et~al.(2017)Vaswani, Shazeer, Parmar, Uszkoreit, Jones, Gomez,
  Kaiser, and Polosukhin]{vaswani2017attention}
Ashish Vaswani, Noam Shazeer, Niki Parmar, Jakob Uszkoreit, Llion Jones,
  Aidan~N Gomez, {\L}ukasz Kaiser, and Illia Polosukhin.
\newblock Attention is all you need.
\newblock \emph{Advances in neural information processing systems}, 30, 2017.

\bibitem[Wang and Agapito(2024)]{wang20243d}
Hengyi Wang and Lourdes Agapito.
\newblock 3d reconstruction with spatial memory.
\newblock \emph{arXiv preprint arXiv:2408.16061}, 2024.

\bibitem[Wang et~al.(2024{\natexlab{a}})Wang, Karaev, Rupprecht, and
  Novotny]{wang2024vggsfm}
Jianyuan Wang, Nikita Karaev, Christian Rupprecht, and David Novotny.
\newblock Vggsfm: Visual geometry grounded deep structure from motion.
\newblock In \emph{Proceedings of the IEEE/CVF conference on computer vision
  and pattern recognition}, pages 21686--21697, 2024{\natexlab{a}}.

\bibitem[Wang et~al.(2025{\natexlab{a}})Wang, Chen, Karaev, Vedaldi, Rupprecht,
  and Novotny]{wang2025vggt}
Jianyuan Wang, Minghao Chen, Nikita Karaev, Andrea Vedaldi, Christian
  Rupprecht, and David Novotny.
\newblock Vggt: Visual geometry grounded transformer.
\newblock In \emph{Proceedings of the Computer Vision and Pattern Recognition
  Conference}, pages 5294--5306, 2025{\natexlab{a}}.

\bibitem[Wang et~al.(2025{\natexlab{b}})Wang, Zhang, Holynski, Efros, and
  Kanazawa]{wang2025continuous}
Qianqian Wang, Yifei Zhang, Aleksander Holynski, Alexei~A Efros, and Angjoo
  Kanazawa.
\newblock Continuous 3d perception model with persistent state.
\newblock In \emph{Proceedings of the Computer Vision and Pattern Recognition
  Conference}, pages 10510--10522, 2025{\natexlab{b}}.

\bibitem[Wang et~al.(2024{\natexlab{b}})Wang, Leroy, Cabon, Chidlovskii, and
  Revaud]{wang2024dust3r}
Shuzhe Wang, Vincent Leroy, Yohann Cabon, Boris Chidlovskii, and Jerome Revaud.
\newblock Dust3r: Geometric 3d vision made easy.
\newblock In \emph{Proceedings of the IEEE/CVF Conference on Computer Vision
  and Pattern Recognition}, pages 20697--20709, 2024{\natexlab{b}}.

\bibitem[Wang et~al.(2025{\natexlab{c}})Wang, Zhou, Zhu, Chang, Zhou, Li, Chen,
  Pang, Shen, and He]{wang2025pi}
Yifan Wang, Jianjun Zhou, Haoyi Zhu, Wenzheng Chang, Yang Zhou, Zizun Li, Junyi
  Chen, Jiangmiao Pang, Chunhua Shen, and Tong He.
\newblock $\\pi^3$: Scalable permutation-equivariant visual geometry learning.
\newblock \emph{arXiv preprint arXiv:2507.13347}, 2025{\natexlab{c}}.

\bibitem[Yang et~al.(2025)Yang, Sax, Liang, Henaff, Tang, Cao, Chai, Meier, and
  Feiszli]{yang2025fast3r}
Jianing Yang, Alexander Sax, Kevin~J Liang, Mikael Henaff, Hao Tang, Ang Cao,
  Joyce Chai, Franziska Meier, and Matt Feiszli.
\newblock Fast3r: Towards 3d reconstruction of 1000+ images in one forward
  pass.
\newblock In \emph{Proceedings of the Computer Vision and Pattern Recognition
  Conference}, pages 21924--21935, 2025.

\bibitem[Yao et~al.(2018)Yao, Luo, Li, Fang, and Quan]{yao2018mvsnet}
Yao Yao, Zixin Luo, Shiwei Li, Tian Fang, and Long Quan.
\newblock Mvsnet: Depth inference for unstructured multi-view stereo.
\newblock In \emph{Proceedings of the European conference on computer vision
  (ECCV)}, pages 767--783, 2018.

\bibitem[Zhang et~al.(2025)Zhang, Wang, Xu, Xue, Rupprecht, Zhou, Shen, and
  Wetzstein]{zhang2025flare}
Shangzhan Zhang, Jianyuan Wang, Yinghao Xu, Nan Xue, Christian Rupprecht,
  Xiaowei Zhou, Yujun Shen, and Gordon Wetzstein.
\newblock Flare: Feed-forward geometry, appearance and camera estimation from
  uncalibrated sparse views.
\newblock In \emph{Proceedings of the Computer Vision and Pattern Recognition
  Conference}, pages 21936--21947, 2025.

\bibitem[Zhong et~al.(2025)Zhong, Zhan, Gao, Chen, Lou, Mao, Neumann, and
  Wang]{zhong2025instantsfm}
Jiankun Zhong, Zitong Zhan, Quankai Gao, Ziyu Chen, Haozhe Lou, Jiageng Mao,
  Ulrich Neumann, and Yue Wang.
\newblock Instantsfm: Fully sparse and parallel structure-from-motion.
\newblock \emph{arXiv preprint arXiv:2510.13310}, 2025.

\bibitem[Zhu et~al.(2018)Zhu, Zhang, Zhou, Shen, Fang, Tan, and
  Quan]{zhu2018very}
Siyu Zhu, Runze Zhang, Lei Zhou, Tianwei Shen, Tian Fang, Ping Tan, and Long
  Quan.
\newblock Very large-scale global sfm by distributed motion averaging.
\newblock In \emph{Proceedings of the IEEE conference on computer vision and
  pattern recognition}, pages 4568--4577, 2018.

\end{thebibliography}
}

\clearpage
\maketitlesupplementary

In this supplementary material, we first present additional qualitative reconstruction results in Section~\ref{sec:qual_results_supp}.
Section~\ref{sec:large_dataset_supp} showcases example reconstructions from large-scale scenes, highlighting the superior scalability of our approach.
In Section~\ref{sec:additonal_results_supp}, we provide per-scene metrics and trajectory visualizations for the Tanks \& Temples dataset.
Section~\ref{sec:subset_recon_supp} compares point-cloud reconstructions from individual subsets with our full reconstruction, illustrating the necessity of partitioning the input images.
Finally, Section~\ref{sec:imp_supp} describes implementation details of our method.

\section{Reconstruction Gallery}
\label{sec:qual_results_supp}
\begin{figure*}[htbp]
    \centering
    \includegraphics[width=\linewidth]{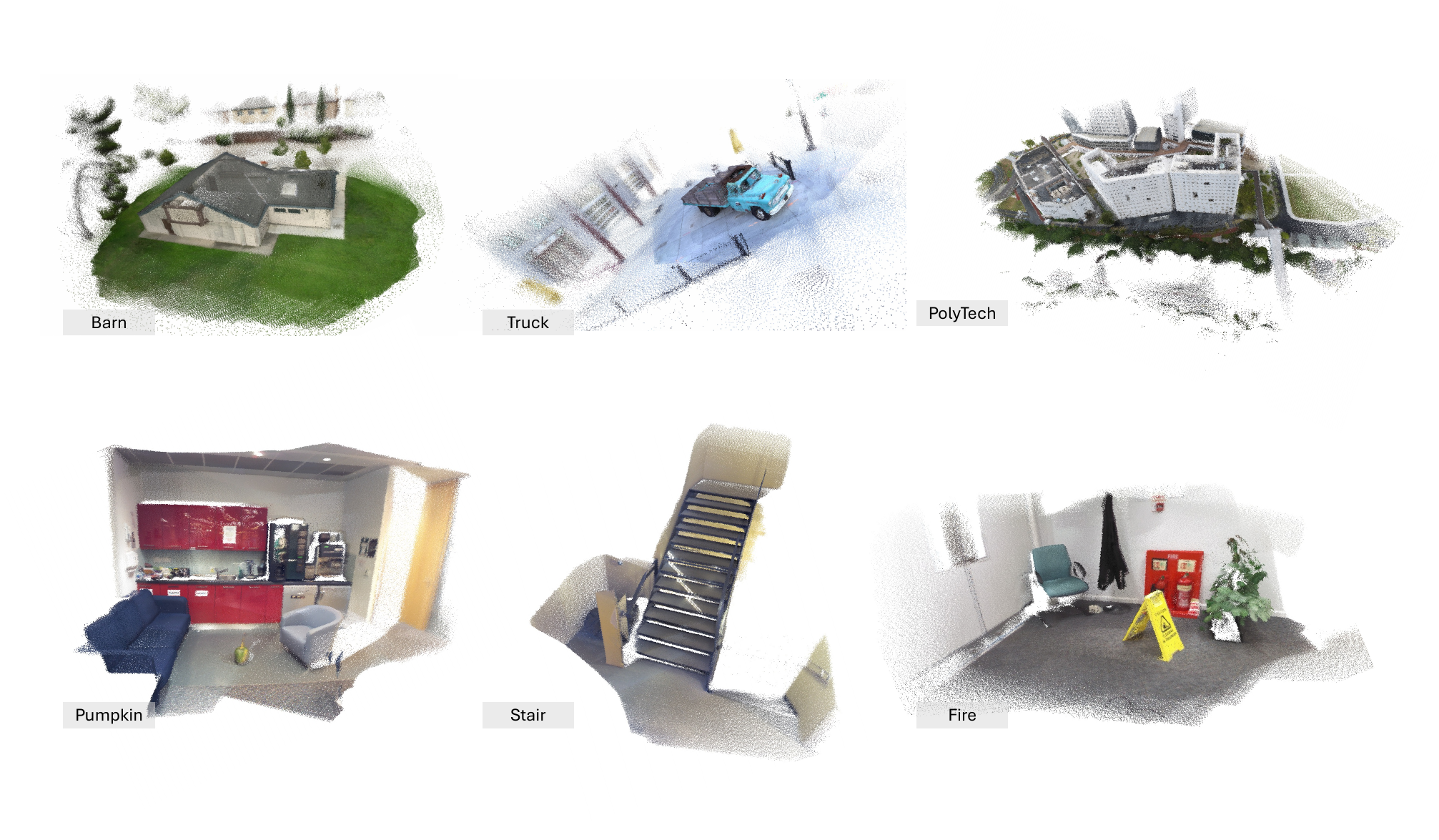}
    \caption{\small 
    \textbf{Qualitative examples of 3D reconstructions on various indoor and outdoor scenes from Tanks \& Temples \cite{knapitsch2017tanks}, 7-Scenes \cite{shotton2013scene} and UrbanScene3D \cite{UrbanScene3D}. } \name{} produces high-quality, detailed reconstructions that preserve fine geometric structure and maintain global consistency. }
    \label{fig:more_qualitative_results} 
\end{figure*}

\begin{figure*}[htbp]
    \centering
    \begin{minipage}{0.9\textwidth}
        \centering
        \includegraphics[width=0.95\linewidth]{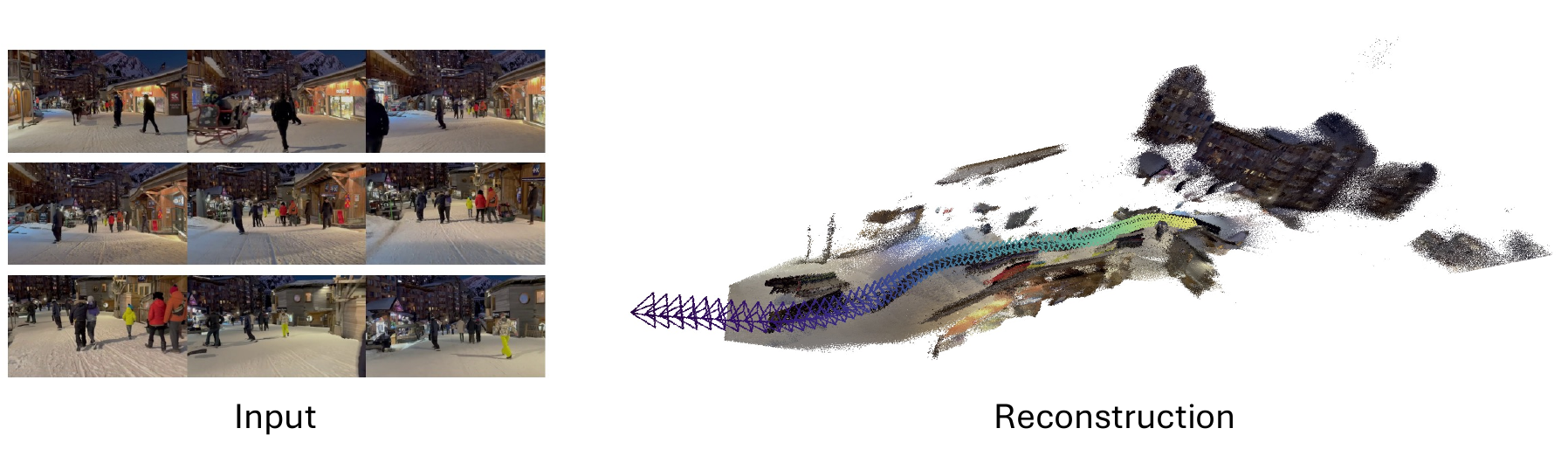}
    \end{minipage}
    \hfill
    \begin{minipage}{0.923\textwidth}
        \centering
        \includegraphics[width=1.0\linewidth]{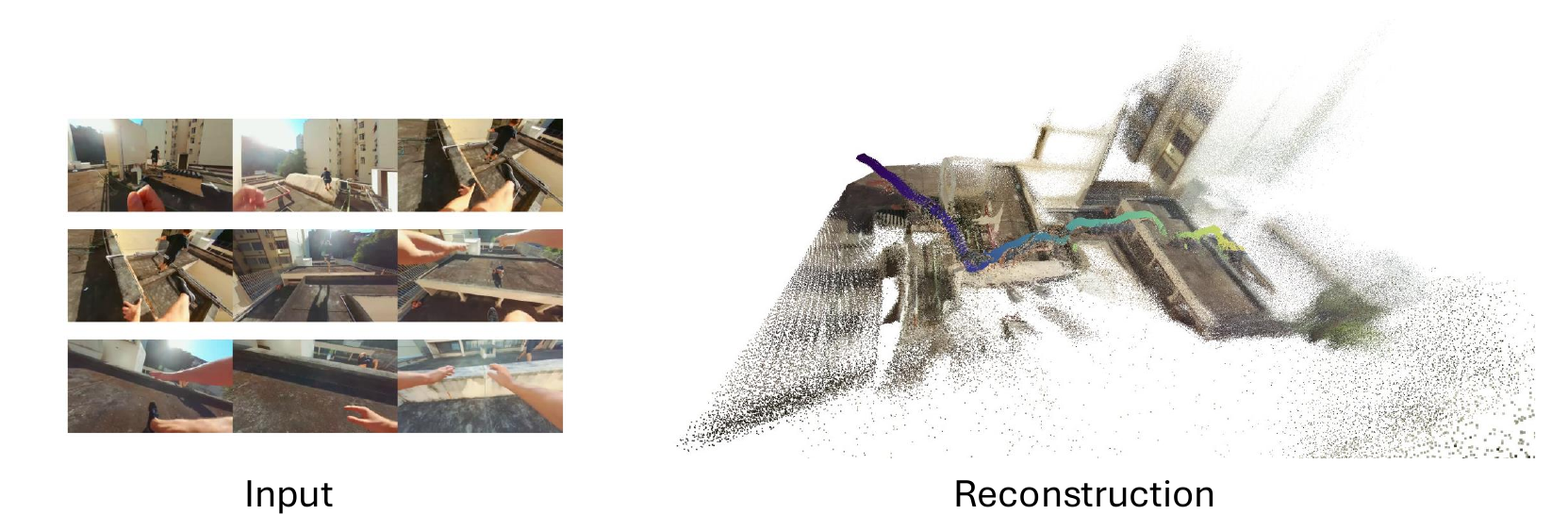}
    \end{minipage}
    \caption{\small Qualitative examples of 3D reconstructions on dynamical scene from internet videos. }
    \label{fig:dynamics_scene}
\end{figure*}

We present additional qualitative reconstruction results on various scenes from Tanks \& Temples \cite{knapitsch2017tanks}, 7-Scenes \cite{shotton2013scene}, and UrbanScene3D \cite{UrbanScene3D} in Fig.~\ref{fig:more_qualitative_results}. \name{} consistently produces robust and detailed reconstructions across both large indoor and outdoor environments. For visualization clarity, we downsample the point clouds and filter points by confidence. We also demonstrate the performance of \name{} on dynamical scenes from internet videos in Fig.~\ref{fig:dynamics_scene}.

\section{Results on Large Scale Dataset}
\label{sec:large_dataset_supp}
\begin{figure*}[htbp]
    \centering
    \includegraphics[width=.84\linewidth, trim={0 0em 0 2em},clip]{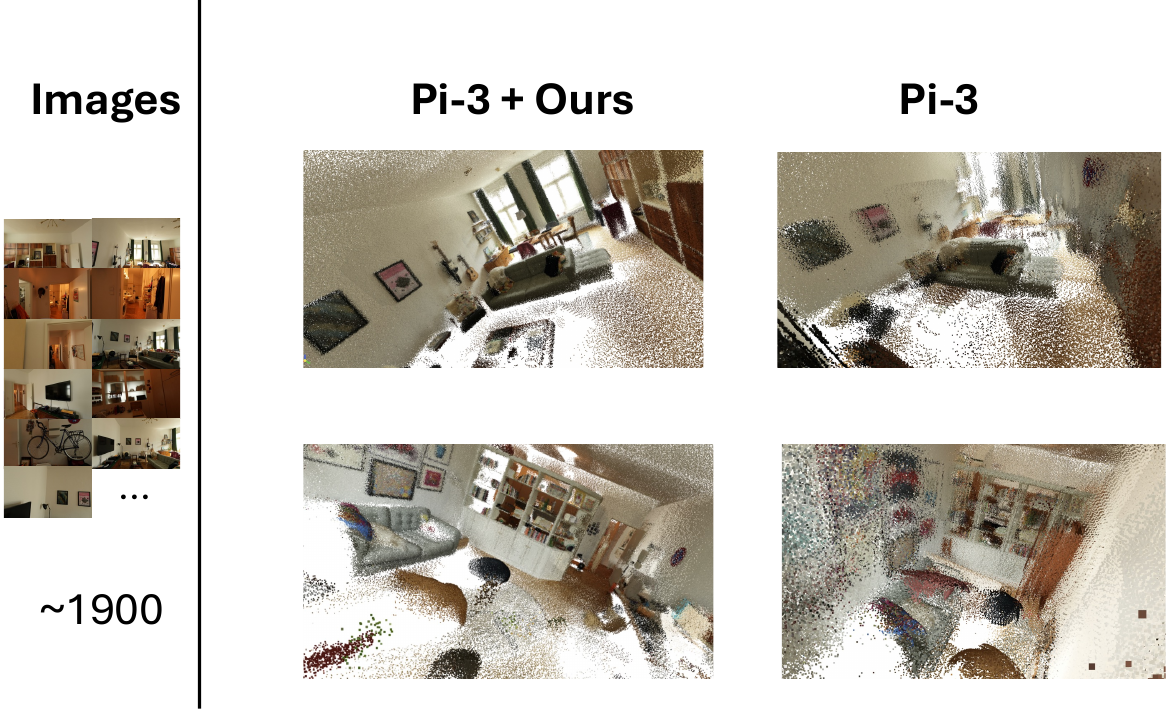}
    \caption{\small 
    \label{fig:large_scale_point_cloud} \textbf{Qualitative comparison of 3D reconstructions on large scale dataset. } \name{} achieves consistently better performance on the Zip-NeRF~\cite{barron2023zip} scenes (Berlin). The input sequence is the original ordering before splitting.}
\end{figure*}

To demonstrate our method’s scalability beyond 1,000 images, we evaluate it on two long sequences from Zip-NeRF \cite{barron2023zip} with approximately 1500 images and 1900 images respectively, as shown in Fig.~\ref{fig:large_scale_point_cloud}. We use the raw Zip-NeRF images directly, which depict complex real-world environments spanning both indoor and outdoor scenes. Compared to the datasets used in the main text, adjacent views in these sequences exhibit substantially less overlap, and the indoor subsets in particular feature challenging spatial layouts with significant geometric complexity. Because $\pi^3$ is unable to process such large numbers of images, we uniformly subsample each dataset to 500 images for a fair comparison, and set our method’s subset size to 500 images accordingly. As illustrated, when presented with large and complex scenes, $\pi^3$ fails to reconstruct the full environment, losing significant structural details and geometric consistency. In the second dataset in particular, $\pi^3$ incorrectly merges two distinct rooms, resulting in severe overlapping artifacts. In addition, we tested $\pi^3$ with our proposed bundle-adjustment step; however, because the initial 3D prior is already severely degraded, the refinement yields minimal visible improvement.

For large-scale datasets with more than 1,000 images, the simple interleaving scheme may select images that are too distant in DINO feature space, resulting in subsets that contain disjoint views and degrade local reconstruction. To address this, after forming the pseudo-video, we refine the interleaving step by searching forward along the sequence for the next image whose (precomputed) DINO similarity to the previously selected image falls within the range $0.5m$ to $0.95m$, where $m$ is the median similarity score with respect to the last chosen image. Once such an image is found, it becomes the new reference point, and the process repeats. This produces the refined sequence $\tilde{P}$ described in Sec.~\ref{subsec:ordering_and_partitioning}. We then apply the sliding-window grouping to form subsets. This procedure avoids selecting images that are overly dissimilar or spatially far apart, ensuring that each subset remains locally coherent.

\section{Additional Results}
\label{sec:additonal_results_supp}
We provide the detailed per-scene comparison on Tanks \& Temples for \name{} and the baselines. Shown in Table~\ref{table:absolute_tanks_metric} and Table~\ref{table:rotation_tanks_metric}, we outperformed the base model and the other baselines on almost every scene. 

In addition, we present qualitative trajectory visualizations for all scenes in the Tanks and Temples dataset (Fig.~\ref{fig:all_tnt_poses}). Each row corresponds to one scene, and each column corresponds to a model. Across scenes, our method produces trajectories that closely follow the ground truth and consistently match or surpass the accuracy of all baselines. These results show that our improvements hold not only in aggregate metrics but also in individual scenes

\section{Subset Reconstruction Results}
\label{sec:subset_recon_supp}
\begin{figure*}[htbp]
    \centering
    \includegraphics[width=.95\linewidth]{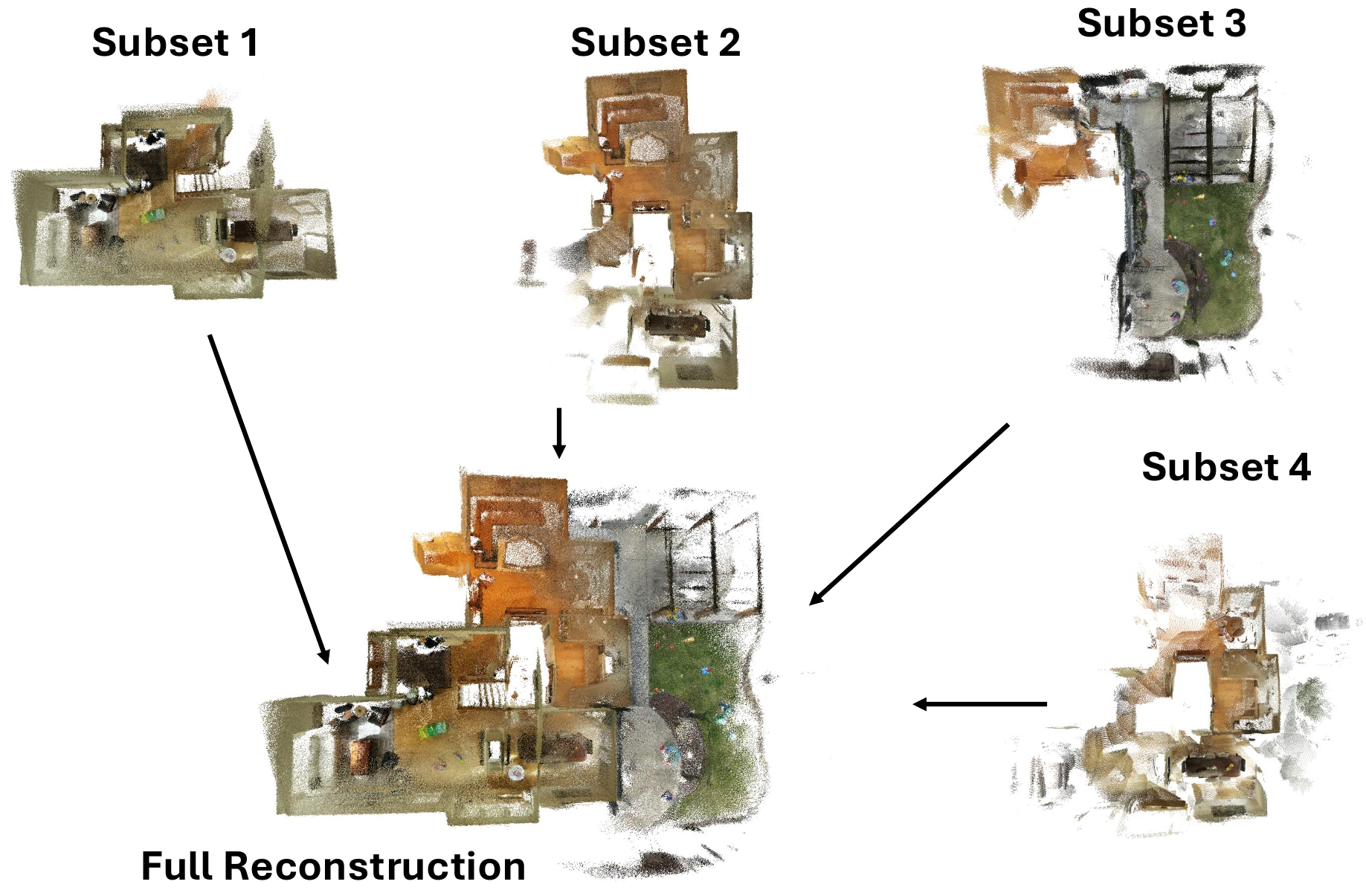}
    
    \caption{\small \label{fig:subset_recon} 
    \textbf{Visualization results for each subset.} Our full reconstruction faithfully recovers the entire house, while each individual subset captures only partial and fragmented portions of the scene. }
\end{figure*}

In Fig.~\ref{fig:subset_recon}, we visualize the 3D reconstruction results for each subset, along with the final merged reconstruction. Each subset captures only a partial portion of the scene, highlighting the need to divide truly large-scale environments into manageable subsets and subsequently merge them to obtain a globally consistent reconstruction.

\section{Splitting Robustness Analysis}
\begin{figure*}[htbp]
    \centering
    \includegraphics[width=1.0\linewidth]{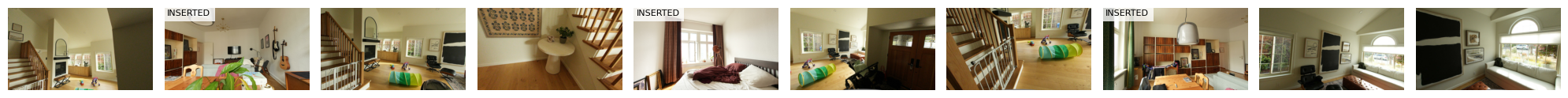}
    
    \includegraphics[width=1.0\linewidth]{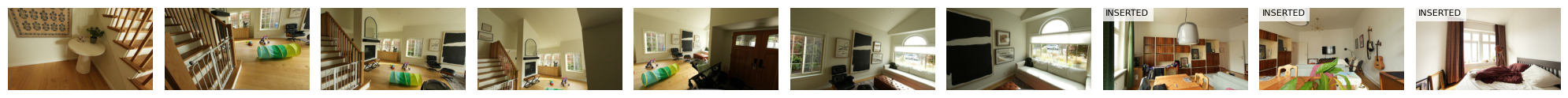}
    
    \caption{\small \label{fig:perturbation_seq} A sample sequence from Zip-NeRF dataset with randomly inserted frames from other scenes (above). The reordered sequence (bottom) places all inserted frames at the end.
     }
\end{figure*}

\begin{figure*}[htbp]
    \centering
    \includegraphics[width=.95\linewidth]{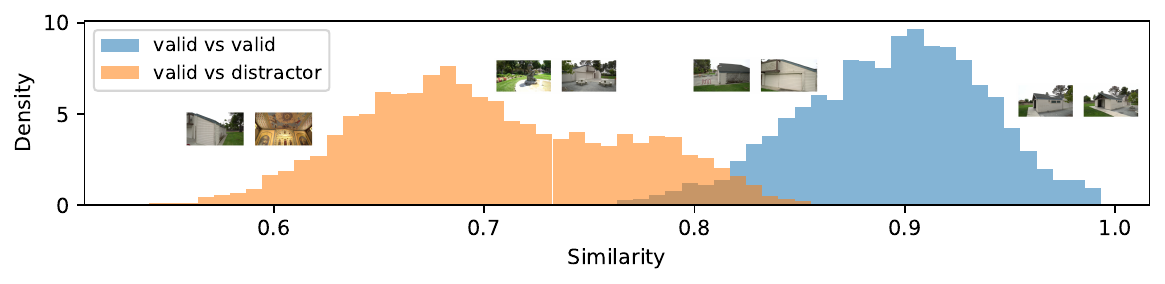}
    
    \caption{\small \label{fig:dino_similarity_perturbation} Histograms of two pair types (valid–valid vs. valid–distractor). The clear separation between the distributions demonstrates the robustness of DINO similarity for detecting outlier frames.
     }
\end{figure*}

We further investigate the robustness of the DINO feature similarity used for sequence construction. We conduct a perturbation experiment by randomly inserting outlier frames from other scenes into a sequence and quantitatively analyzing the resulting DINO similarities. Specifically, we plot histograms of similarity scores for valid–valid pairs and valid–distractor pairs. As shown in Figure \ref{fig:dino_similarity_perturbation}, the two distributions are clearly separated, indicating that DINO similarity reliably distinguishes outlier frames and supports robust pseudo-video rearrangement. Figure \ref{fig:perturbation_seq} illustrates an example sequence with randomly inserted frames alongside its rearranged counterpart, where the outlier frames are effectively pushed to the end of the sequence.

\section{Implementation Details}
\label{sec:imp_supp}
We report all hyperparameters used in our experiments. For bundle adjustment, we set the initial learning rate to $3 \times 10^{-3}$ and optimize for 300 iterations using a cosine-annealing schedule. For subset alignment, we retain only points whose confidence exceeds the 70th percentile. For tracking, we perform direct matching across at most five frames, extract up to 4,096 keypoints per image, and apply a reprojection error threshold of 8 pixels during geometric verification.

\section{Limitations}
Although \model{} demonstrates strong scalability and improved accuracy on existing benchmarks, several limitations remain. First, the method may degrade when viewpoint changes between input images are extremely drastic. In such cases, feature correspondences become sparse or unreliable, which can lead to fragmented reconstruction or unstable pose estimation. This issue is particularly pronounced in large-scale indoor scenes with wide baselines or severe occlusions. \model{} typically performs better on outdoor scenes. Second, the DINO similarity–based splitting heuristic assumes that feature similarity provides a reliable proxy for geometric consistency. When scenes contain large textureless regions or overly similar images from different places, DINO embeddings may become less discriminative. This can result in suboptimal clustering, where geometrically related images are split across different groups or unrelated images are merged together. Consequently, local reconstructions may lack sufficient overlap, negatively affecting downstream global alignment.

\begin{table*}[t]
\centering
\scriptsize
\setlength{\tabcolsep}{3pt}

\begin{tabular}{
l|
c c c c c c c|
c c c c c c c|
c c c c c c c
}
\hline

\multicolumn{1}{c|}{} &
\multicolumn{7}{c|}{\textbf{ATE} $\downarrow$} &
\multicolumn{7}{c|}{\textbf{RRE}$\downarrow$} &
\multicolumn{7}{c}{\textbf{RTE}$\downarrow$}
\\ \hline

Scene &
\rotatebox{90}{CUT3R \cite{wang2025continuous}} &
\rotatebox{90}{TT3R \cite{chen2025ttt3r}} &
\rotatebox{90}{MAST3R-SfM \cite{duisterhof2025mast3r}} &
\rotatebox{90}{VGGT* \cite{wang2025vggt}} &
\rotatebox{90}{Pi3 \cite{wang2025pi}} &
\rotatebox{90}{VGGT* + Ours} &
\rotatebox{90}{Pi3 + Ours} &
\rotatebox{90}{CUT3R \cite{wang2025continuous}} &
\rotatebox{90}{TT3R \cite{chen2025ttt3r}} &
\rotatebox{90}{MAST3R-SfM \cite{duisterhof2025mast3r}} &
\rotatebox{90}{VGGT* \cite{wang2025vggt}} &
\rotatebox{90}{Pi3 \cite{wang2025pi}} &
\rotatebox{90}{VGGT* + Ours} &
\rotatebox{90}{Pi3 + Ours} &
\rotatebox{90}{CUT3R \cite{wang2025continuous}} &
\rotatebox{90}{TT3R \cite{chen2025ttt3r}} &
\rotatebox{90}{MAST3R-SfM \cite{duisterhof2025mast3r}} &
\rotatebox{90}{VGGT* \cite{wang2025vggt}} &
\rotatebox{90}{Pi3 \cite{wang2025pi}} &
\rotatebox{90}{VGGT* + Ours} &
\rotatebox{90}{Pi3 + Ours} 
\\
\hline

Barn &
1.246 & 0.827 & 0.093 & 0.074 & 0.048 & \underline{0.035} & \textbf{0.030} &
0.793 & 0.773 & 0.268 & 0.524 & \underline{0.234} & 0.440 & \textbf{0.179} &
0.058 & 0.064 & \underline{0.009} & 0.038 & 0.015 & 0.026 & \textbf{0.006}
\\ 

Caterpillar &
0.974 & 0.277 & 0.036 & 0.051 & 0.035 & \textbf{0.021} & \underline{0.023} &
0.987 & 0.931 & 0.658 & 0.299 & 0.195 & \underline{0.104} & \textbf{0.097} &
0.069 & 0.058 & 0.012 & 0.028 & 0.018 & \underline{0.007} & \textbf{0.005}
\\ 

Church &
2.288 & 1.176 & 0.983 & 2.934 & \underline{0.358} & 3.002 & \textbf{0.330} &
1.684 & 5.337 & 1.638 & 7.378 & \underline{0.396} & 8.326 & \textbf{0.380} &
0.124 & 0.172 & 0.101 & 0.274 & \underline{0.057} & 0.298 & \textbf{0.046}
\\ 

Ignatius &
0.663 & 0.188 & 0.024 & 0.043 & 0.035 & \textbf{0.020} & \underline{0.021} &
1.043 & 0.785 & 0.197 & 0.210 & 0.156 & \underline{0.144} & \textbf{0.117} &
0.091 & 0.069 & \textbf{0.008} & 0.025 & 0.020 & \underline{0.010} & \textbf{0.008}
\\ 

Meeting Room &
1.819 & 0.646 & 0.046 & 0.071 & \textbf{0.035} & 0.039 & \underline{0.037} &
0.987 & 0.884 & \textbf{0.163} & 0.340 & 0.201 & 0.230 & \underline{0.166} &
0.102 & 0.121 & \textbf{0.008} & 0.036 & 0.021 & 0.016 & \underline{0.009}
\\ 

Truck &
0.637 & 0.149 & 0.032 & 0.035 & 0.031 & \textbf{0.016} & \underline{0.018} &
0.940 & 0.940 & 0.204 & 0.234 & 0.194 & \underline{0.135} & \textbf{0.128} &
0.076 & 0.059 & \textbf{0.006} & 0.023 & \underline{0.019} & \textbf{0.006} & \textbf{0.006}
\\ \hline

\end{tabular}

\caption{Per-scene camera pose evaluation results (ATE, RRE, RTE) across methods for the Tanks \& Temples dataset. }
\label{table:absolute_tanks_metric}
\end{table*}

\begin{table*}[t]
\centering
\scriptsize
\setlength{\tabcolsep}{3pt}

\begin{tabular}{
l|
c c c c c c c|
c c c c c c c|
c c c c c c c
}
\hline

\multicolumn{1}{c|}{} &
\multicolumn{7}{c|}{\textbf{RRA@30} $\uparrow$} &
\multicolumn{7}{c|}{\textbf{RTA@30}$\uparrow$} &
\multicolumn{7}{c}{\textbf{AUC@30}$\uparrow$}
\\ \hline

Scene &
\rotatebox{90}{CUT3R \cite{wang2025continuous}} &
\rotatebox{90}{TT3R \cite{chen2025ttt3r}} &
\rotatebox{90}{MAST3R-SfM \cite{duisterhof2025mast3r}} &
\rotatebox{90}{VGGT* \cite{wang2025vggt}} &
\rotatebox{90}{Pi3 \cite{wang2025pi}} &
\rotatebox{90}{VGGT* + Ours} &
\rotatebox{90}{Pi3 + Ours} &
\rotatebox{90}{CUT3R \cite{wang2025continuous}} &
\rotatebox{90}{TT3R \cite{chen2025ttt3r}} &
\rotatebox{90}{MAST3R-SfM \cite{duisterhof2025mast3r}} &
\rotatebox{90}{VGGT* \cite{wang2025vggt}} &
\rotatebox{90}{Pi3 \cite{wang2025pi}} &
\rotatebox{90}{VGGT* + Ours} &
\rotatebox{90}{Pi3 + Ours} &
\rotatebox{90}{CUT3R \cite{wang2025continuous}} &
\rotatebox{90}{TT3R \cite{chen2025ttt3r}} &
\rotatebox{90}{MAST3R-SfM \cite{duisterhof2025mast3r}} &
\rotatebox{90}{VGGT* \cite{wang2025vggt}} &
\rotatebox{90}{Pi3 \cite{wang2025pi}} &
\rotatebox{90}{VGGT* + Ours} &
\rotatebox{90}{Pi3 + Ours} 
\\
\hline

Barn &
\underline{77.64} & \textbf{100} & \textbf{100} & \textbf{100} & \textbf{100} & \textbf{100} & \textbf{100} &
88.85 & 89.71 & \textbf{99.99} & 99.88 & \underline{99.96} & 99.93 & \textbf{99.99} &
40.33 & 63.90 & 89.23 & 93.86 & \textbf{95.82} &  \underline{95.44} &  \textbf{95.82}
\\

Caterpillar &
87.97 & \textbf{100} & \underline{99.48} & \textbf{100} & \textbf{100} & \textbf{100} & \textbf{100} &
89.51 & 98.66 & 99.70 & 99.91 & 99.77 & \underline{99.99} & \textbf{100} &
48.15 & 90.89 & 92.75 & 95.36 & 95.85 & \underline{96.48} &  \textbf{96.5}
\\

Church &
44.33 & \underline{80.38} & 79.12 & 56.82 & \textbf{100} & 58.21 & \textbf{100} &
52.13 & 69.56 & 78.01 & 54.60 & \underline{95.03} & 61.49 & \textbf{95.46} &
21.65 & 46.04 & 78.01 & 34.40 & \underline{86.18} &  51.79 & \textbf{87.33}
\\

Ignatius &
\underline{99.44} & \textbf{100} & \textbf{100} & \textbf{100} & \textbf{100} & \textbf{100} & \textbf{100} &
1.78 & 99.38 & \underline{99.99} & 99.97 & 99.97 & \underline{99.99} & \textbf{100} &
61.09 & 88.08 & \textbf{99.99} & 95.57 & 95.98 & \textbf{96.47} & 96.41
\\

Meeting Room &
\underline{95.09} & \textbf{100} & \textbf{100} & \textbf{100} & \textbf{100} & \textbf{100} & \textbf{100} &
69.00 & 92.98 & \underline{99.95} & 99.83 & \textbf{99.96} & 99.94 & \underline{99.95} &
39.76 & 68.96 & 94.73 & 93.51 & \textbf{95.72} & \underline{95.32} & 95.27
\\

Truck &
\underline{99.77} & \textbf{100} & \textbf{100} & \textbf{100} & \textbf{100} & \textbf{100} & \textbf{100} &
93.76 & 99.35 & \textbf{100} & \underline{99.99} & 99.98 & \textbf{100} & \textbf{100} &
60.91 & 88.25 & \textbf{100} & 96.05 & 95.97 &  \underline{96.48} & 92.28
\\ \hline

\end{tabular}

\caption{Per-scene camera pose evaluation results (RRA, RTA, AUC) across methods for the Tanks \& Temples dataset. }
\label{table:rotation_tanks_metric}
\end{table*}

\begin{figure*}[htbp]
    \centering
    \setlength{\tabcolsep}{2pt}
    \renewcommand{\arraystretch}{1.0}
    \hspace{-1.3cm}  %
    \begin{tabular}{lcccccc}
        \raisebox{4.5\height}{Barn} & 
        \includegraphics[width=0.15\textwidth]{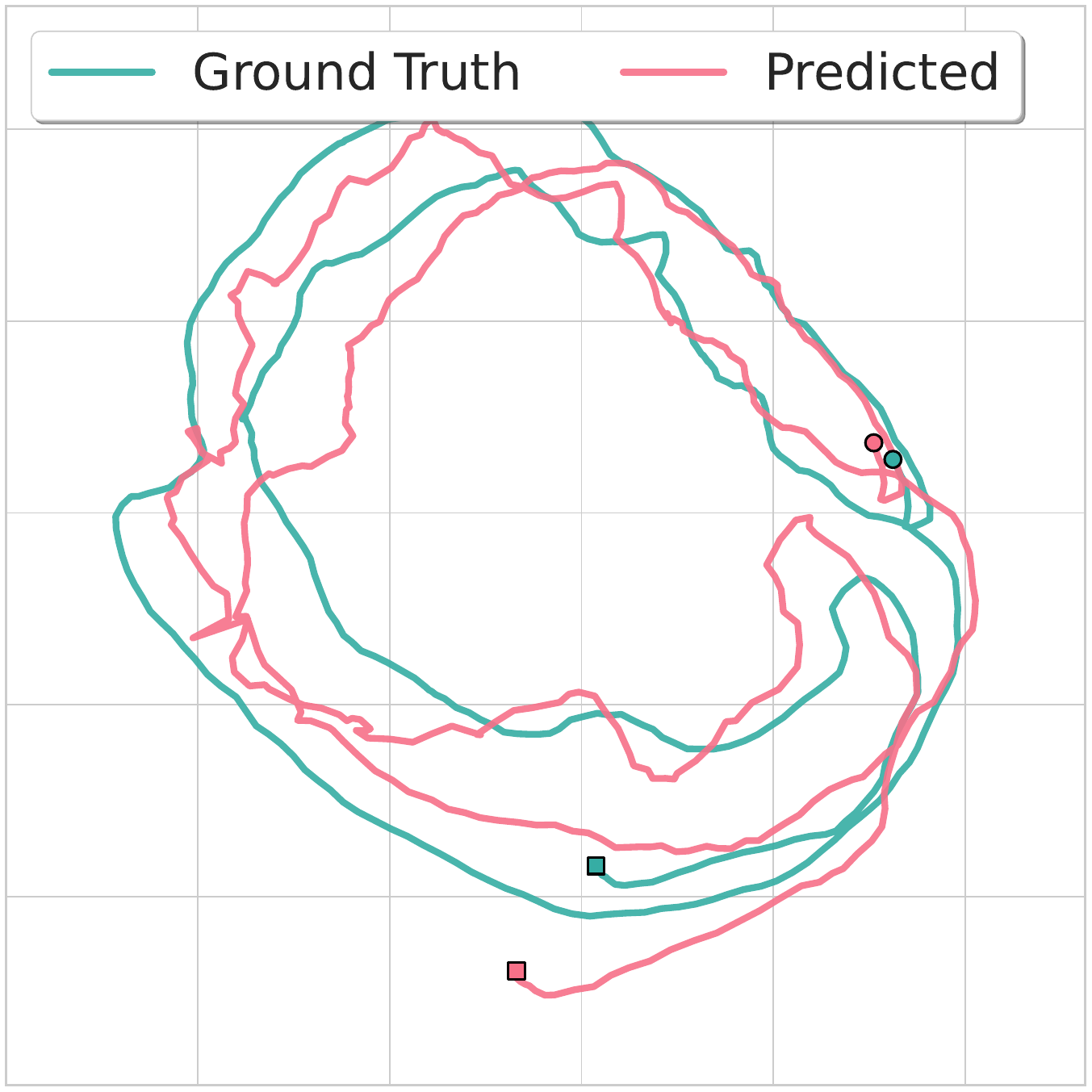} &
        \includegraphics[width=0.15\textwidth]{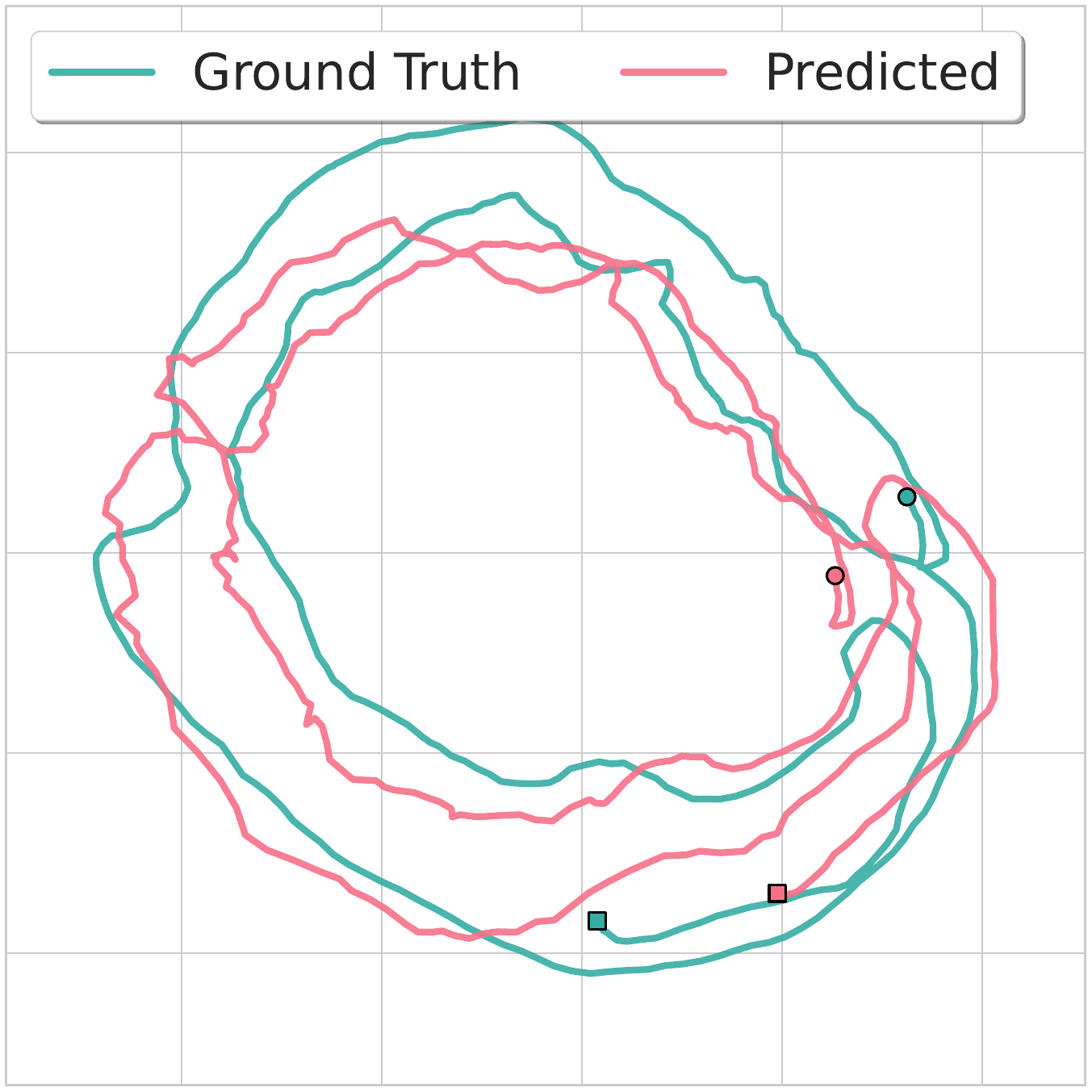} & 
        \includegraphics[width=0.15\textwidth]{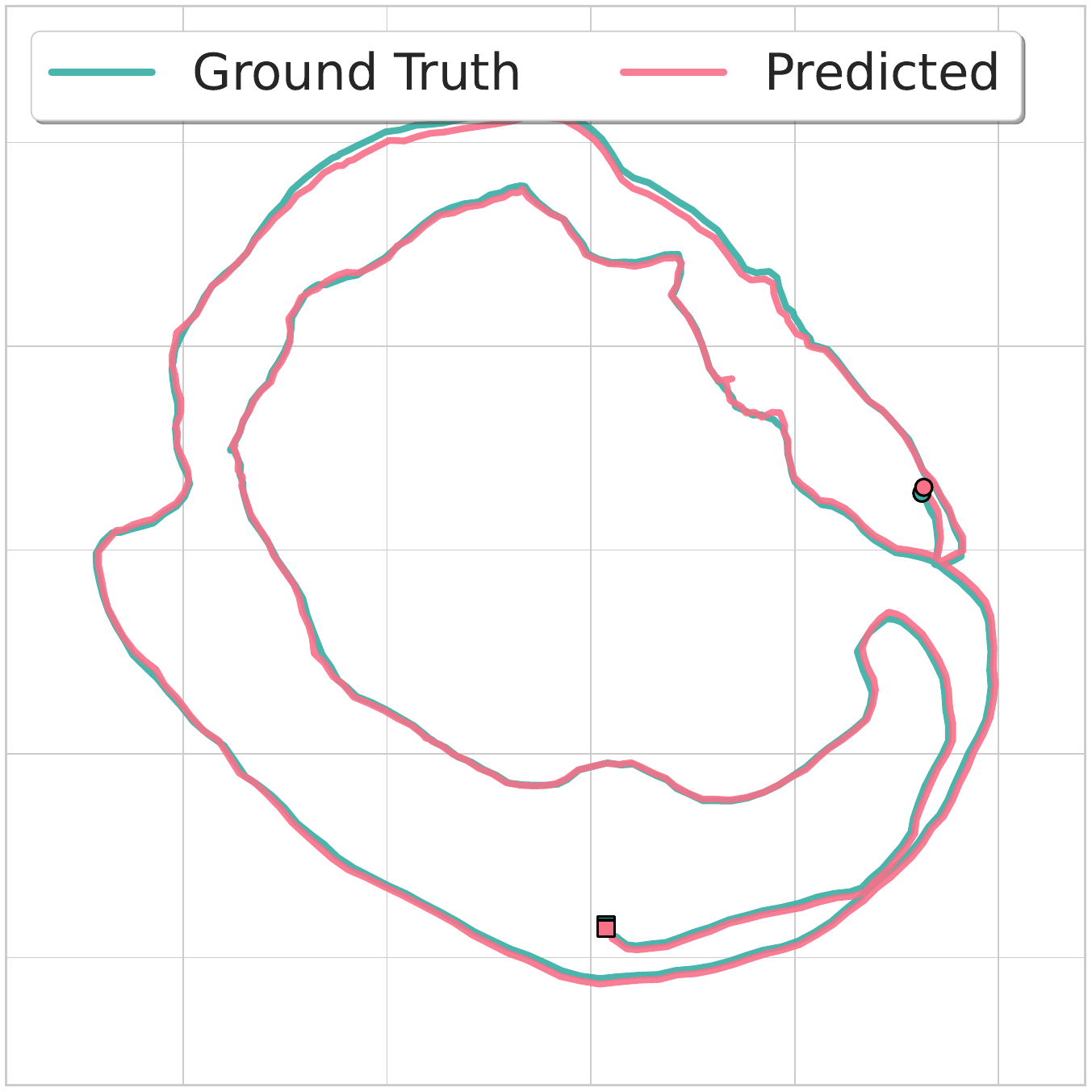} & 
        \includegraphics[width=0.15\textwidth]{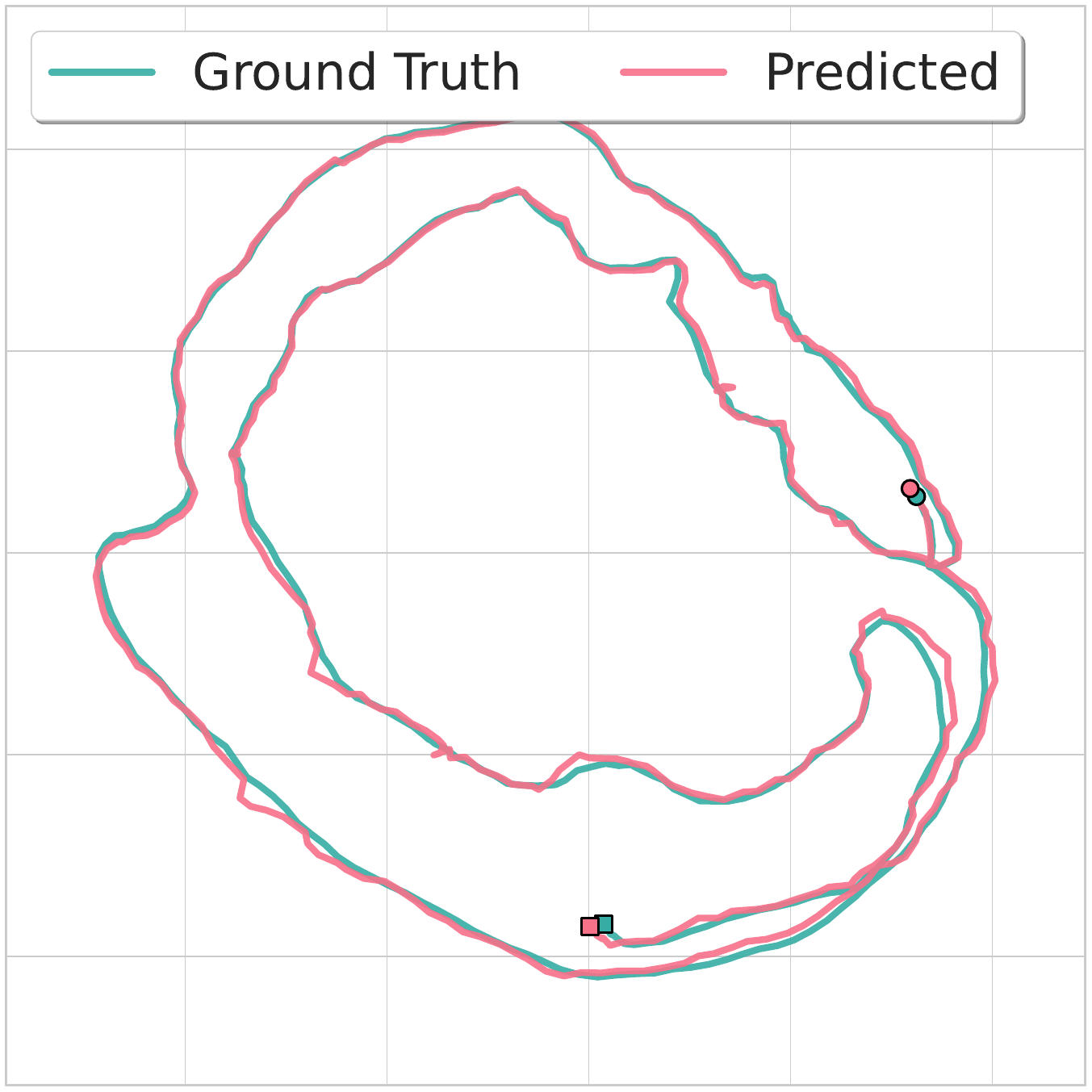} &
        \includegraphics[width=0.15\textwidth]{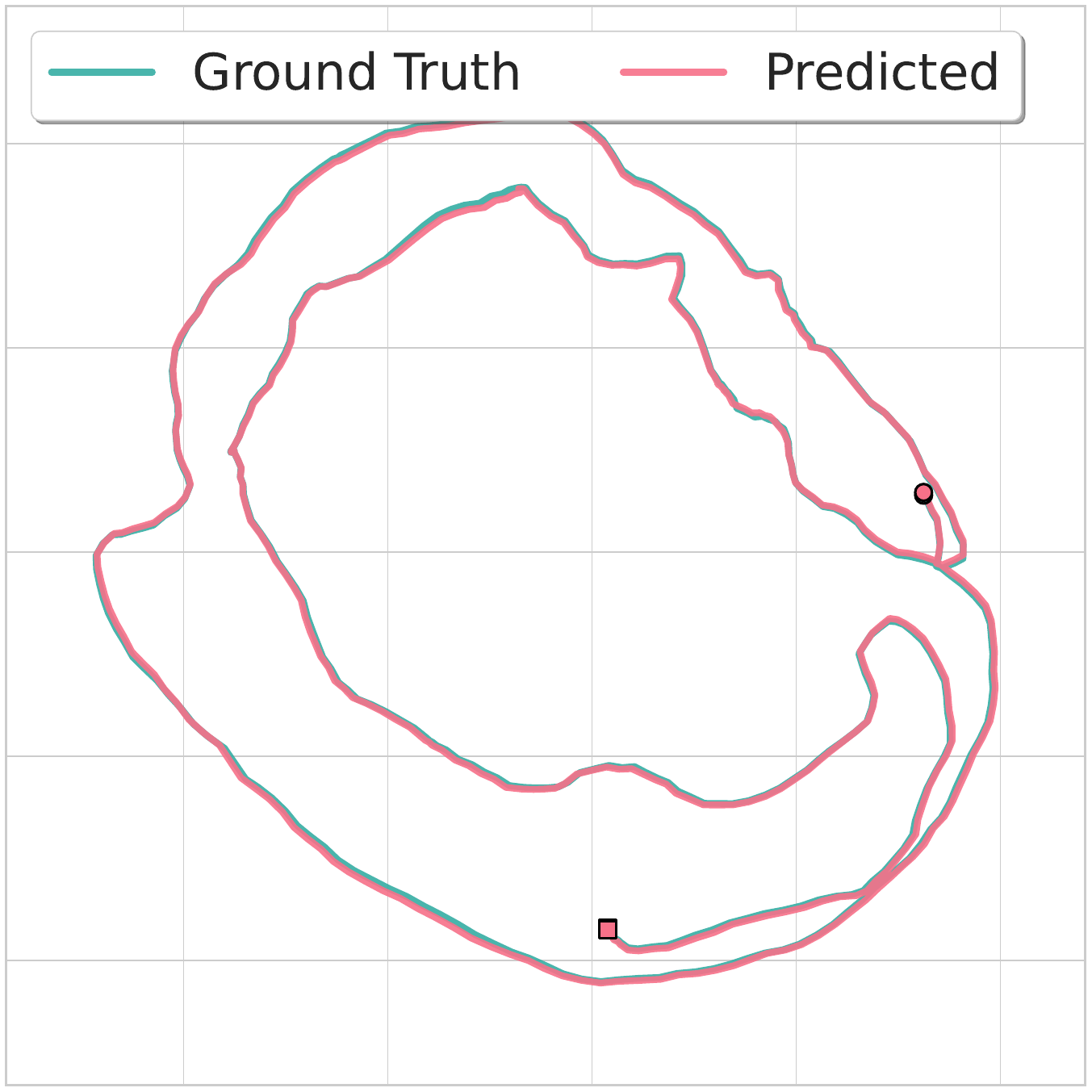} &
        \includegraphics[width=0.15\textwidth]{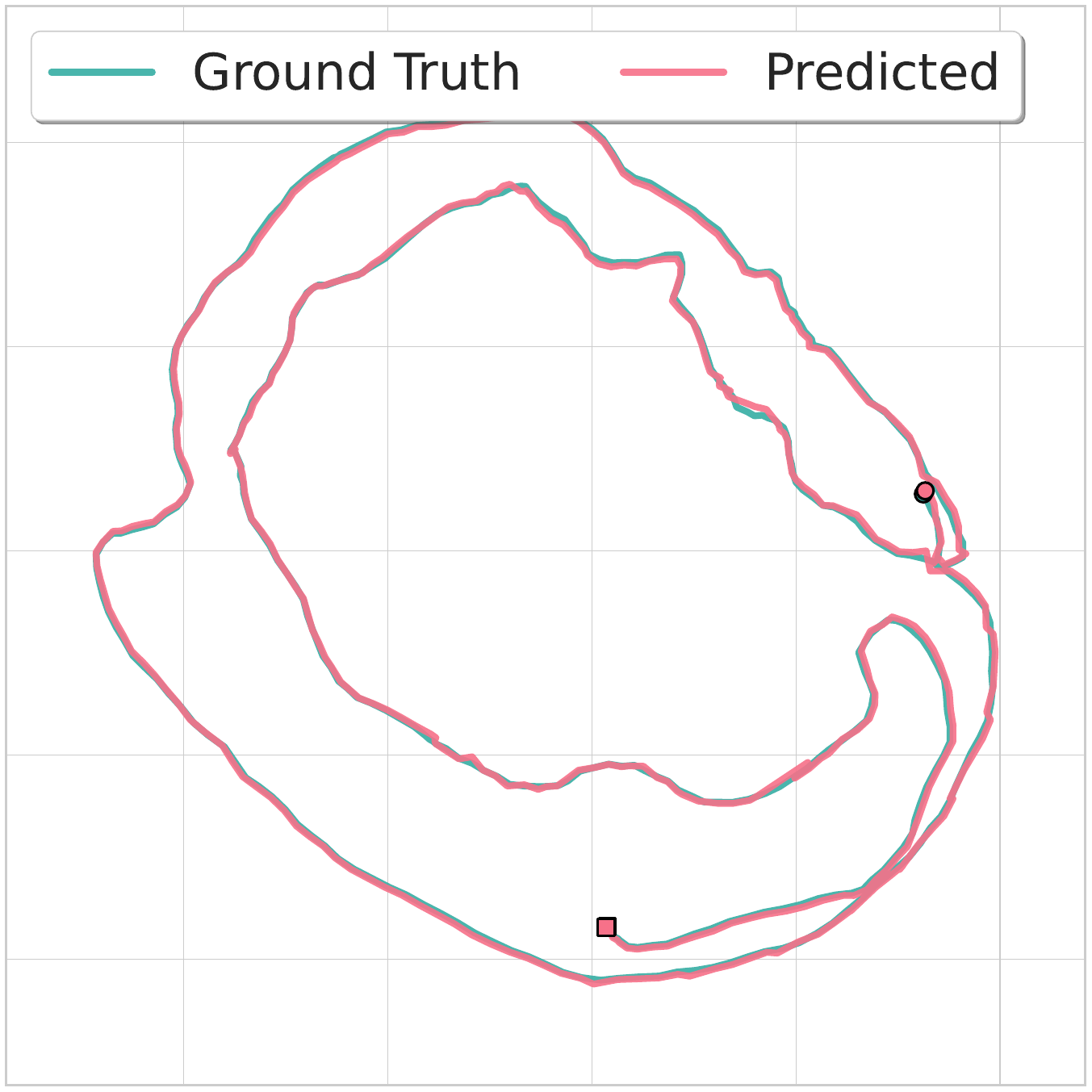} \\

        \raisebox{4.5\height}{Caterpillar} & 
        \includegraphics[width=0.15\textwidth]{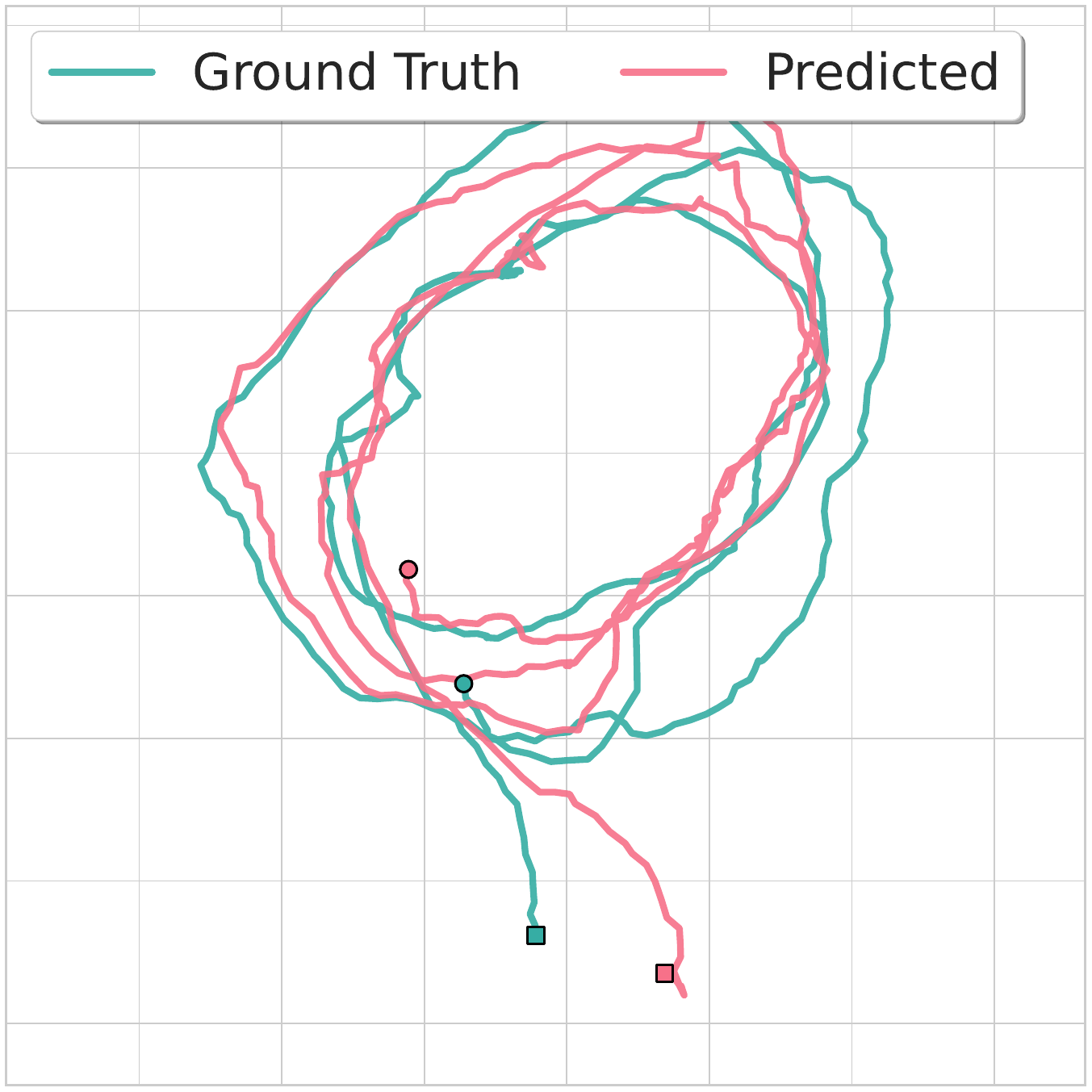} &
        \includegraphics[width=0.15\textwidth]{assets/figures/traj_vis_Caterpillar_cut3r.pdf} & 
        \includegraphics[width=0.15\textwidth]{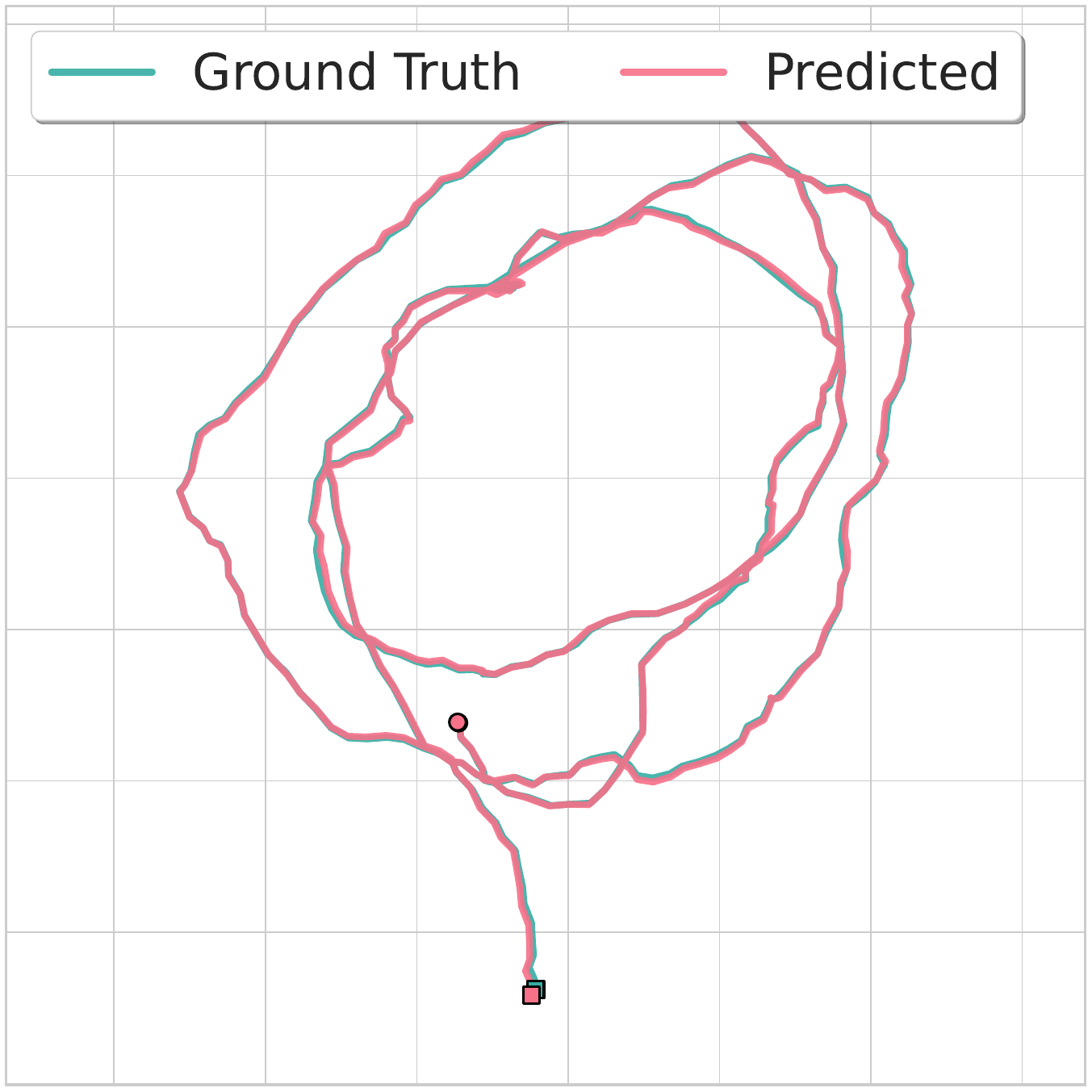} & 
        \includegraphics[width=0.15\textwidth]{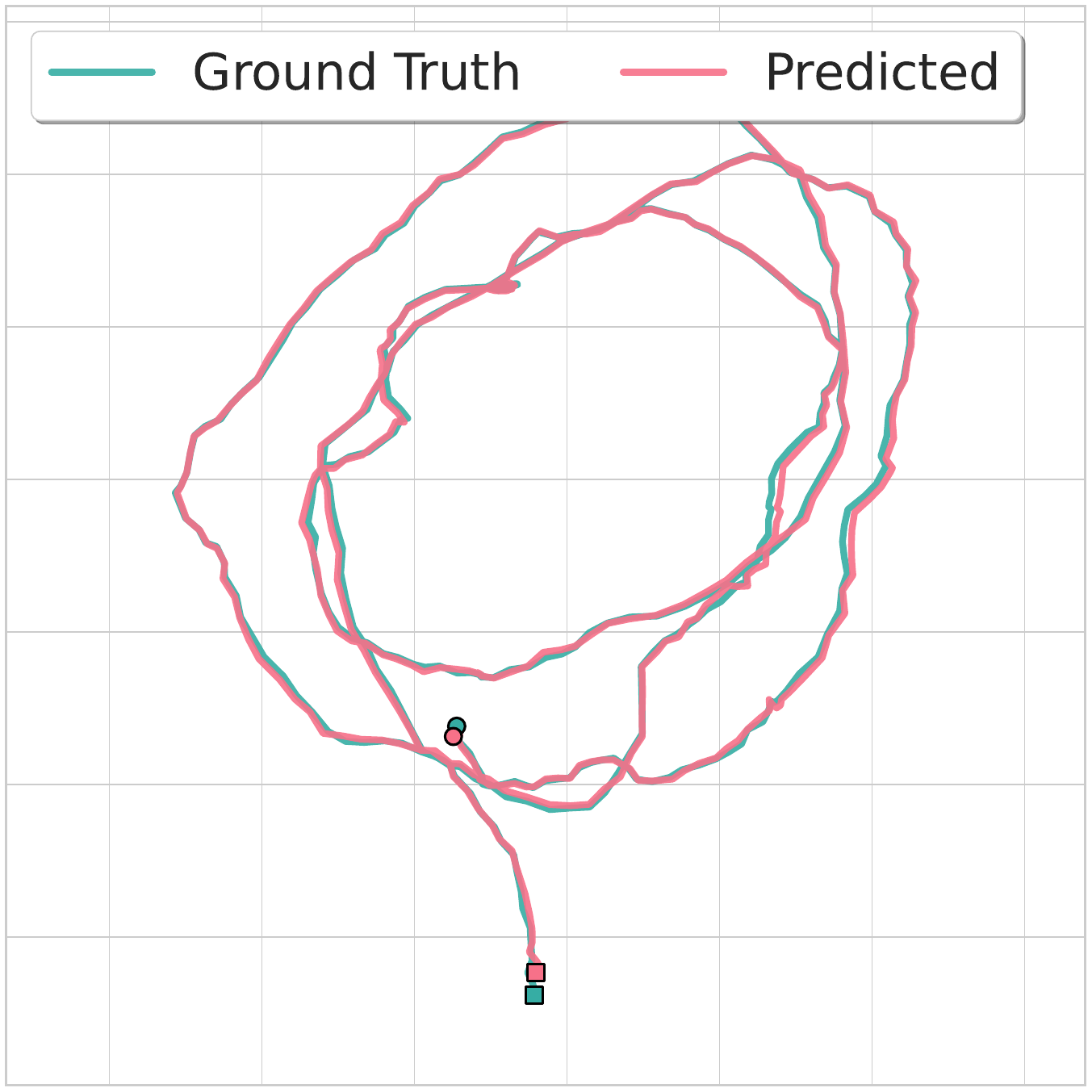} &
        \includegraphics[width=0.15\textwidth]{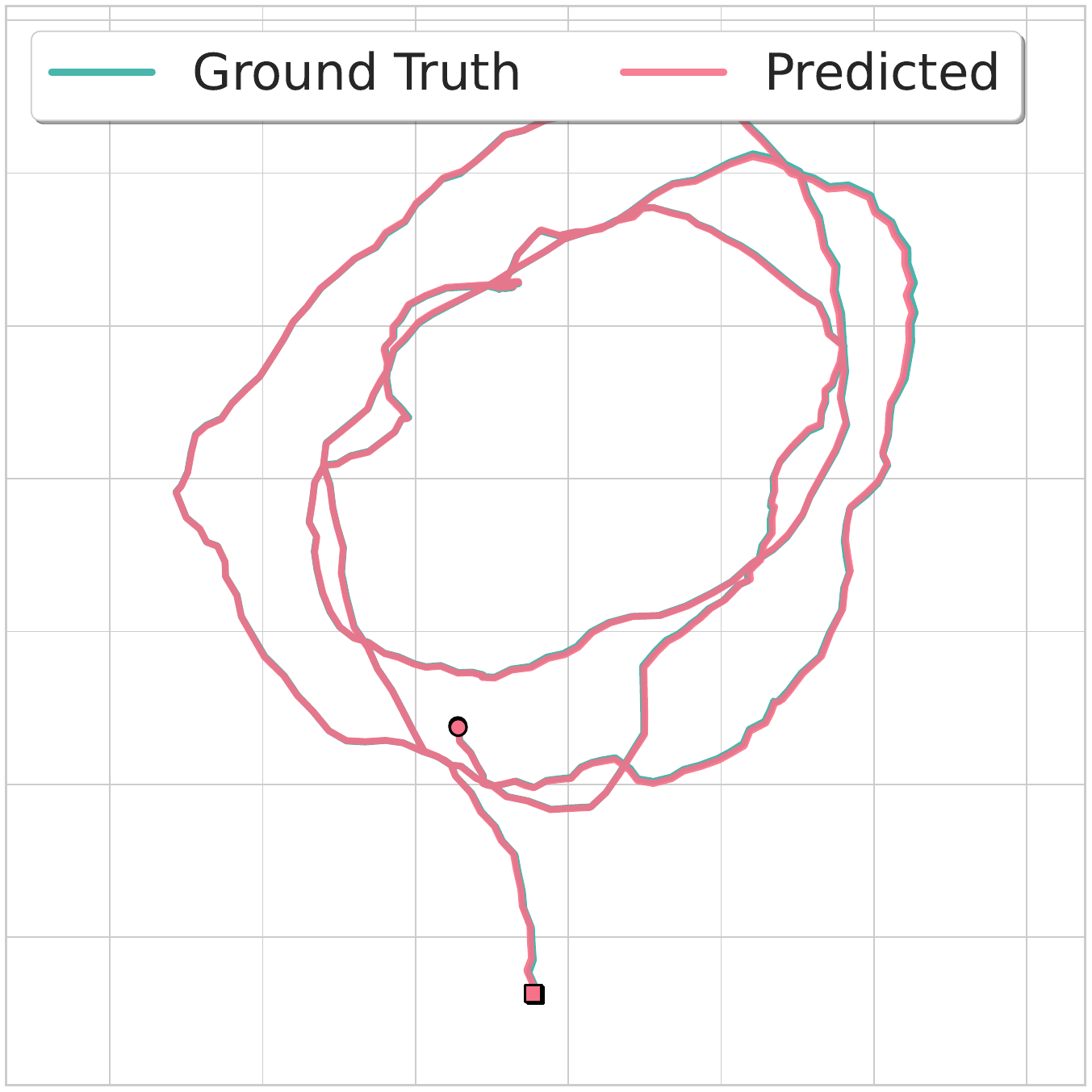} &
        \includegraphics[width=0.15\textwidth]{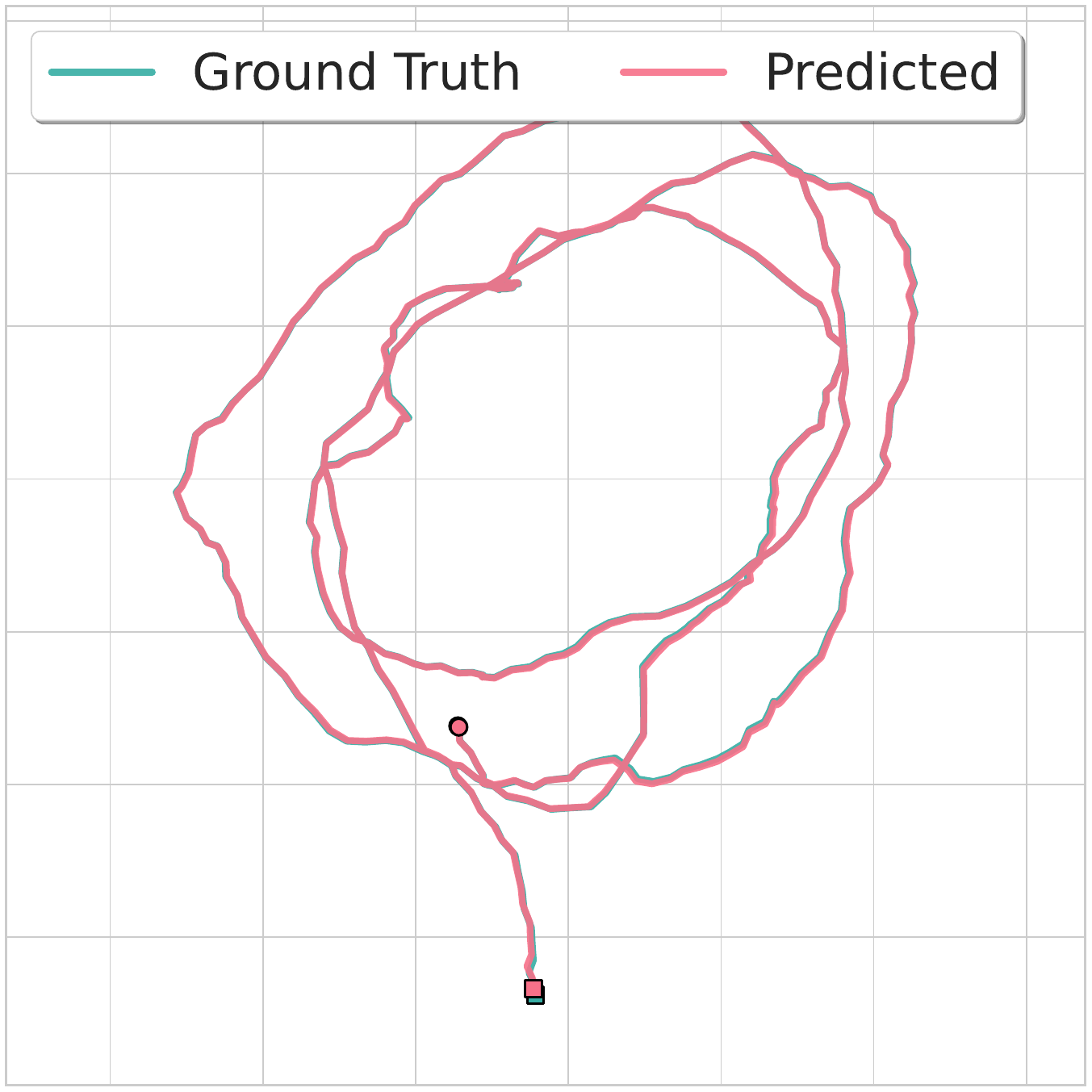} \\

        \raisebox{4.5\height}{Church} & 
        \includegraphics[width=0.15\textwidth]{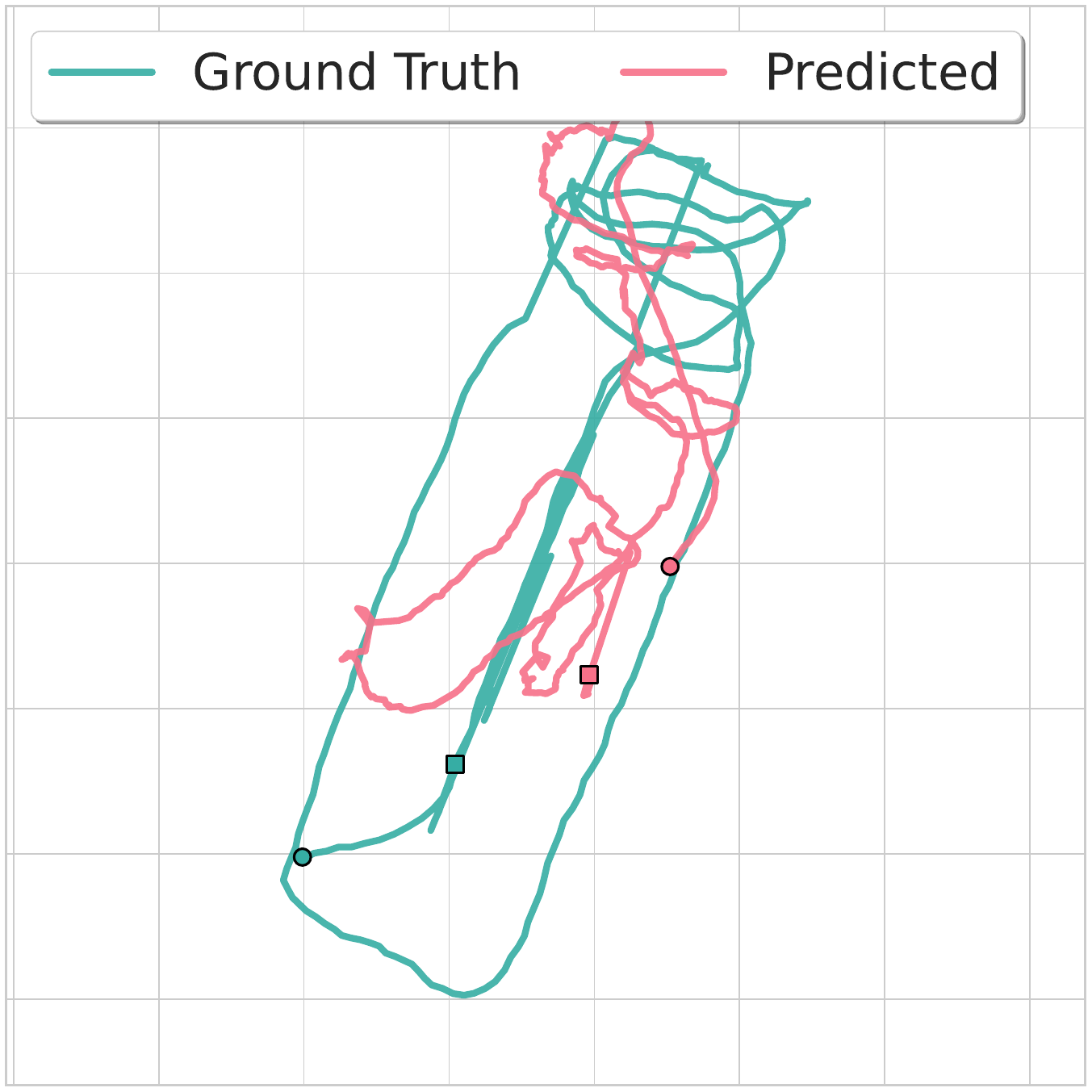} &
        \includegraphics[width=0.15\textwidth]{assets/figures/traj_vis_Church_cut3r.pdf} & 
        \includegraphics[width=0.15\textwidth]{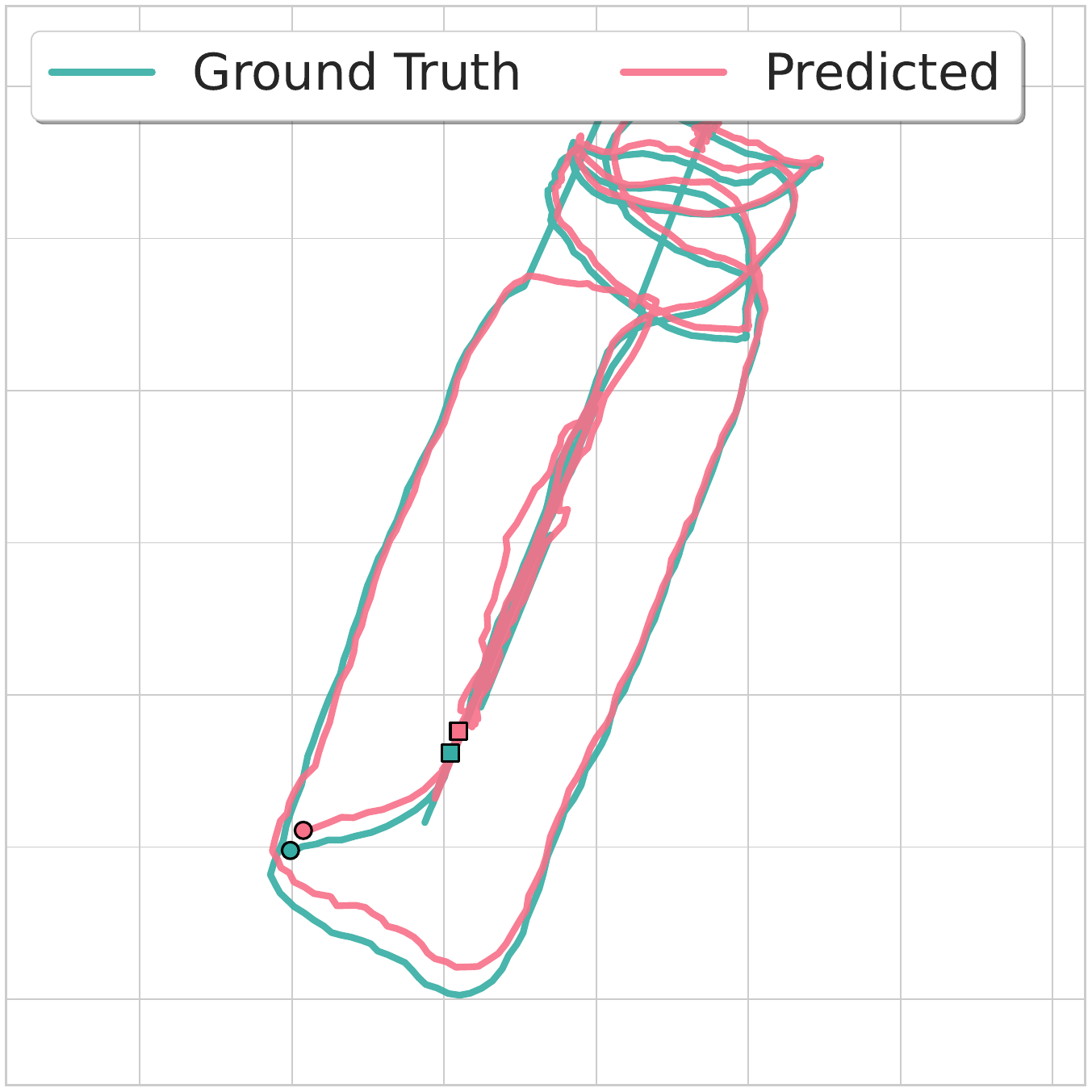} & 
        \includegraphics[width=0.15\textwidth]{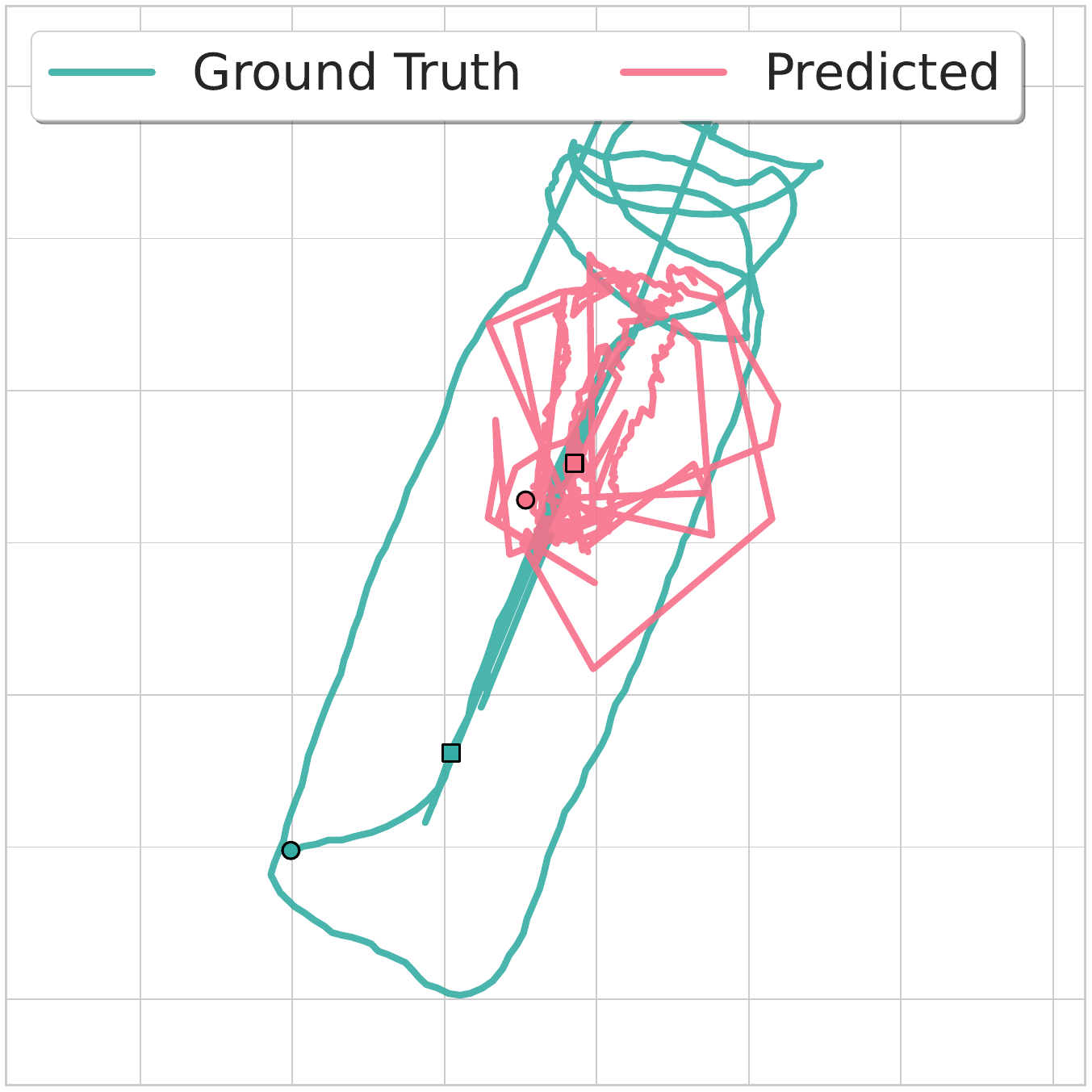} &
        \includegraphics[width=0.15\textwidth]{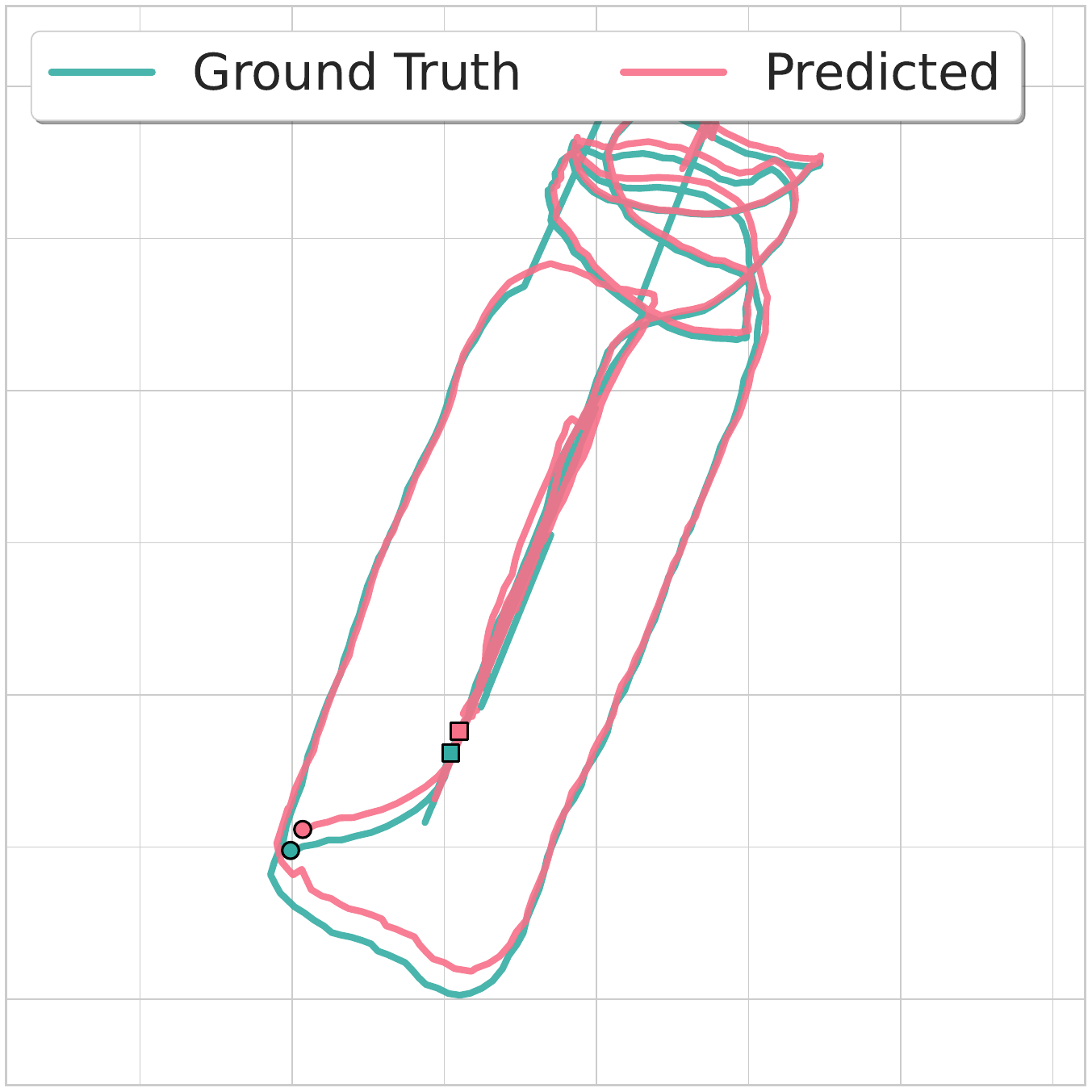} &
        \includegraphics[width=0.15\textwidth]{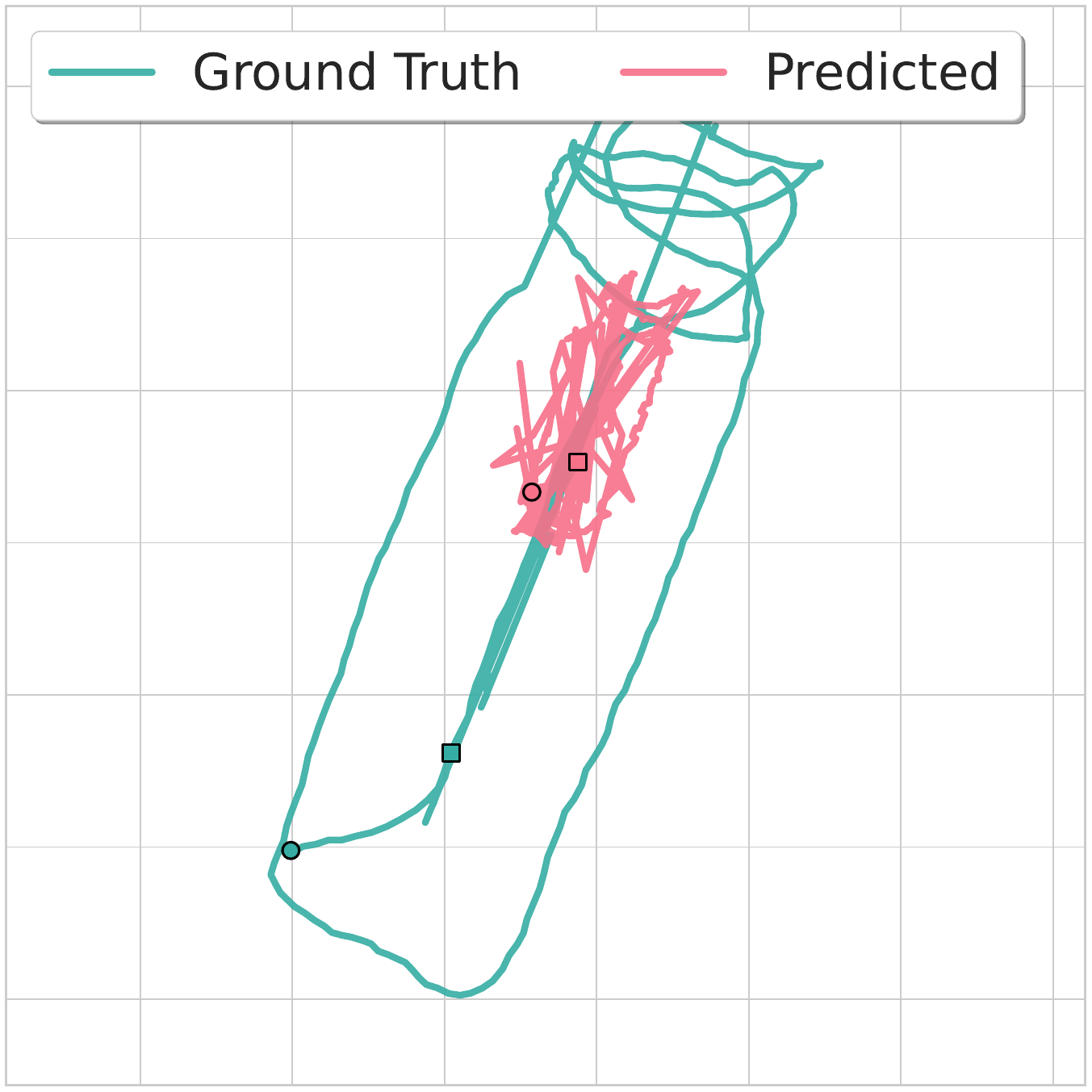} \\

        \raisebox{4.5\height}{Ignatius} & 
        \includegraphics[width=0.15\textwidth]{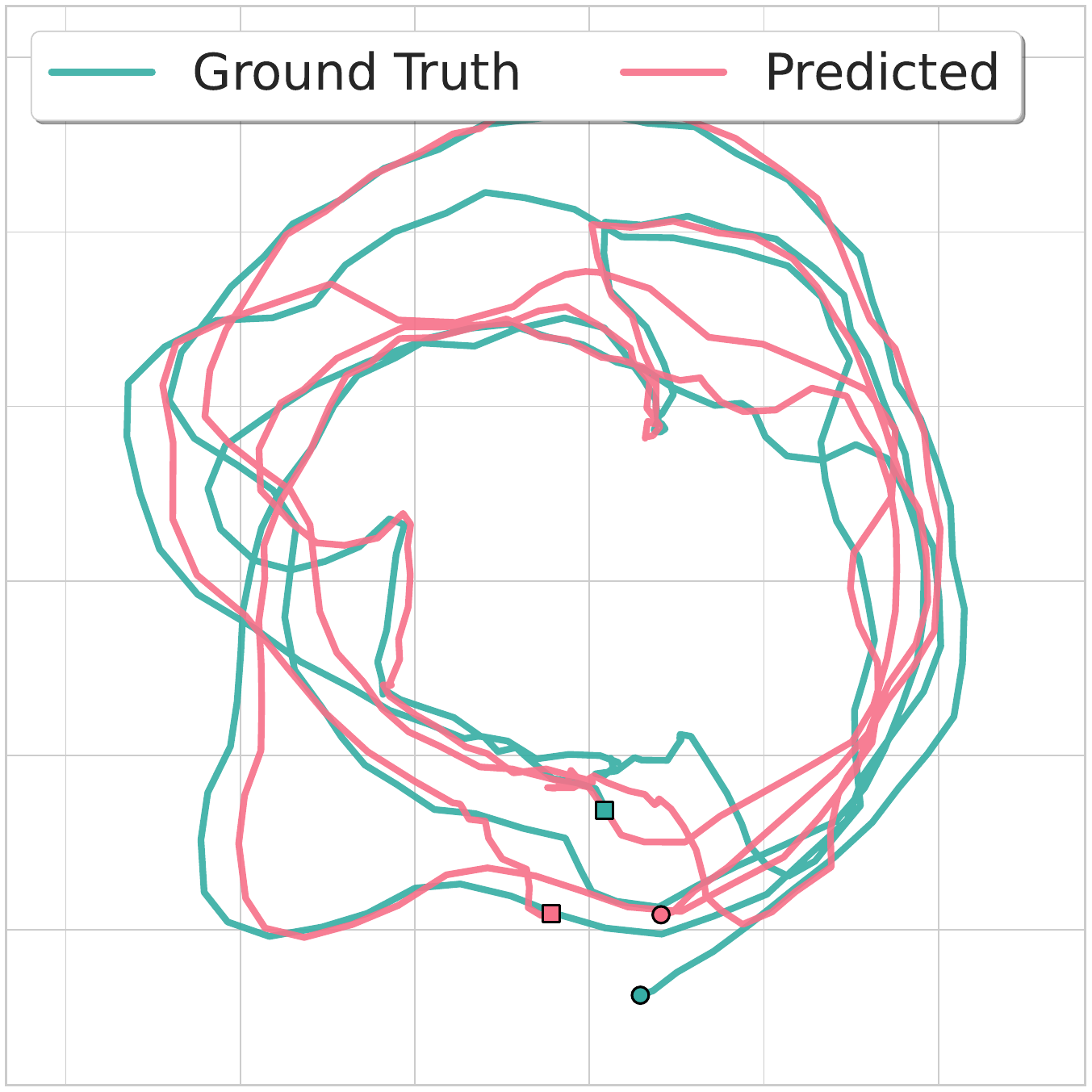} &
        \includegraphics[width=0.15\textwidth]{assets/figures/traj_vis_Ignatius_cut3r.pdf} & 
        \includegraphics[width=0.15\textwidth]{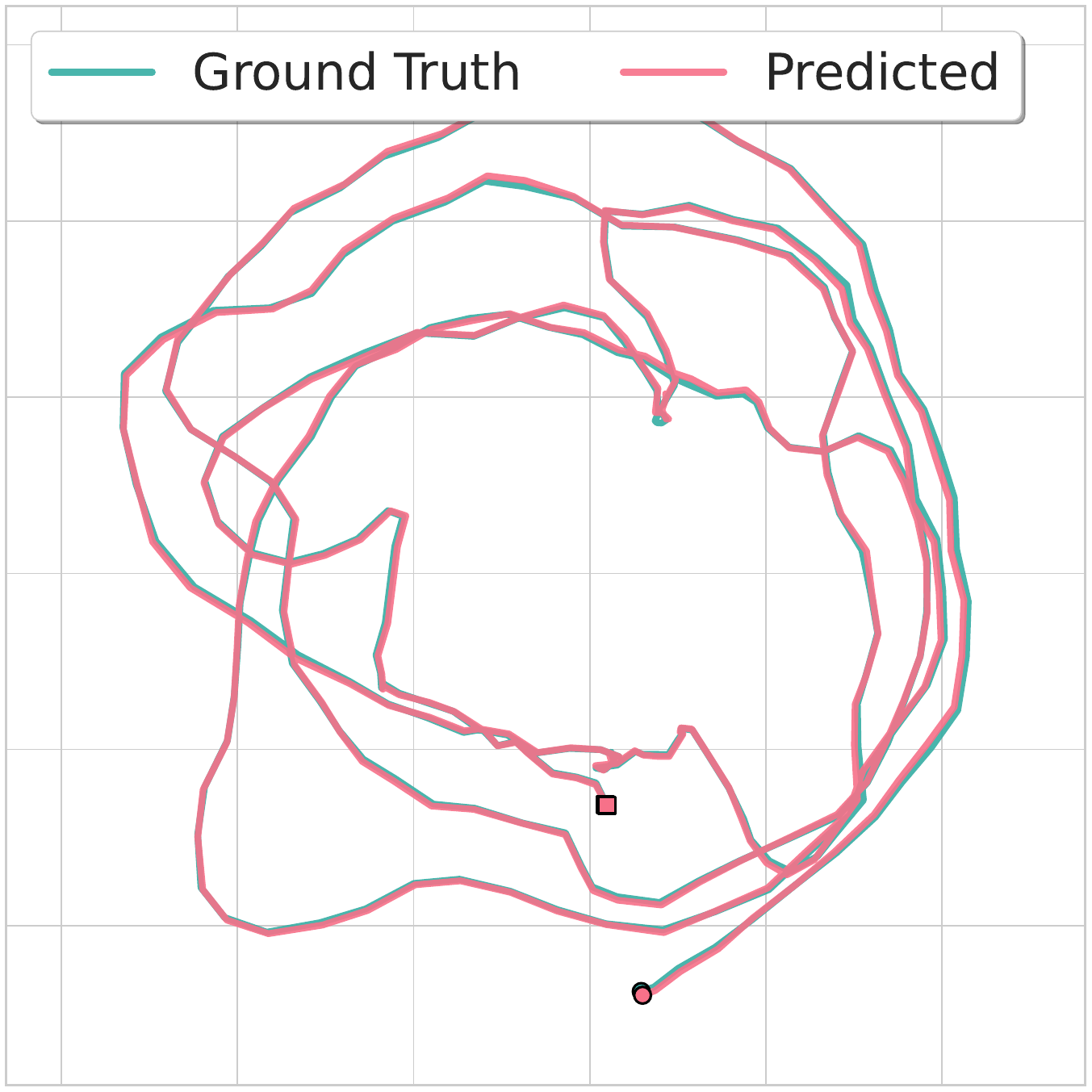} & 
        \includegraphics[width=0.15\textwidth]{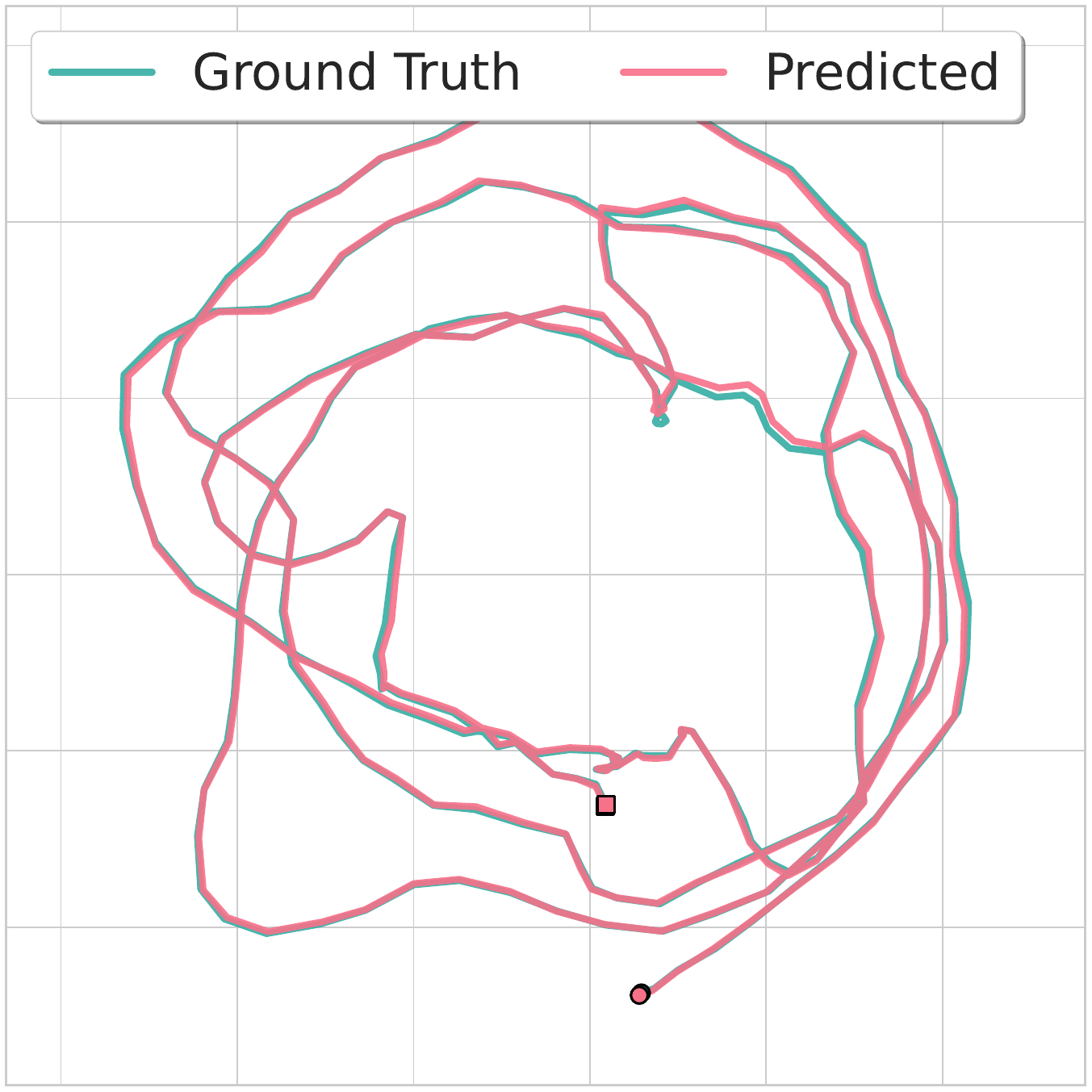} &
        \includegraphics[width=0.15\textwidth]{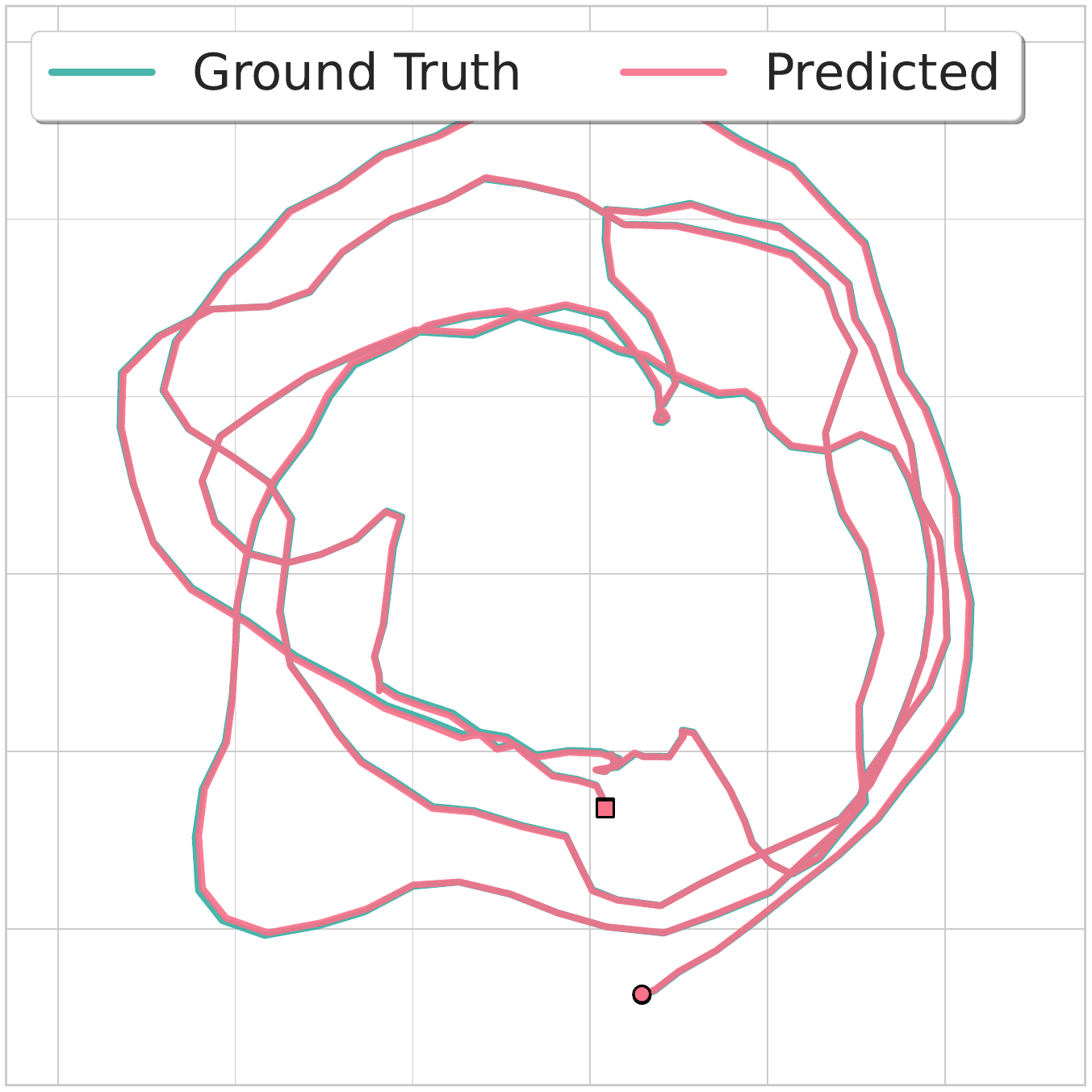} &
        \includegraphics[width=0.15\textwidth]{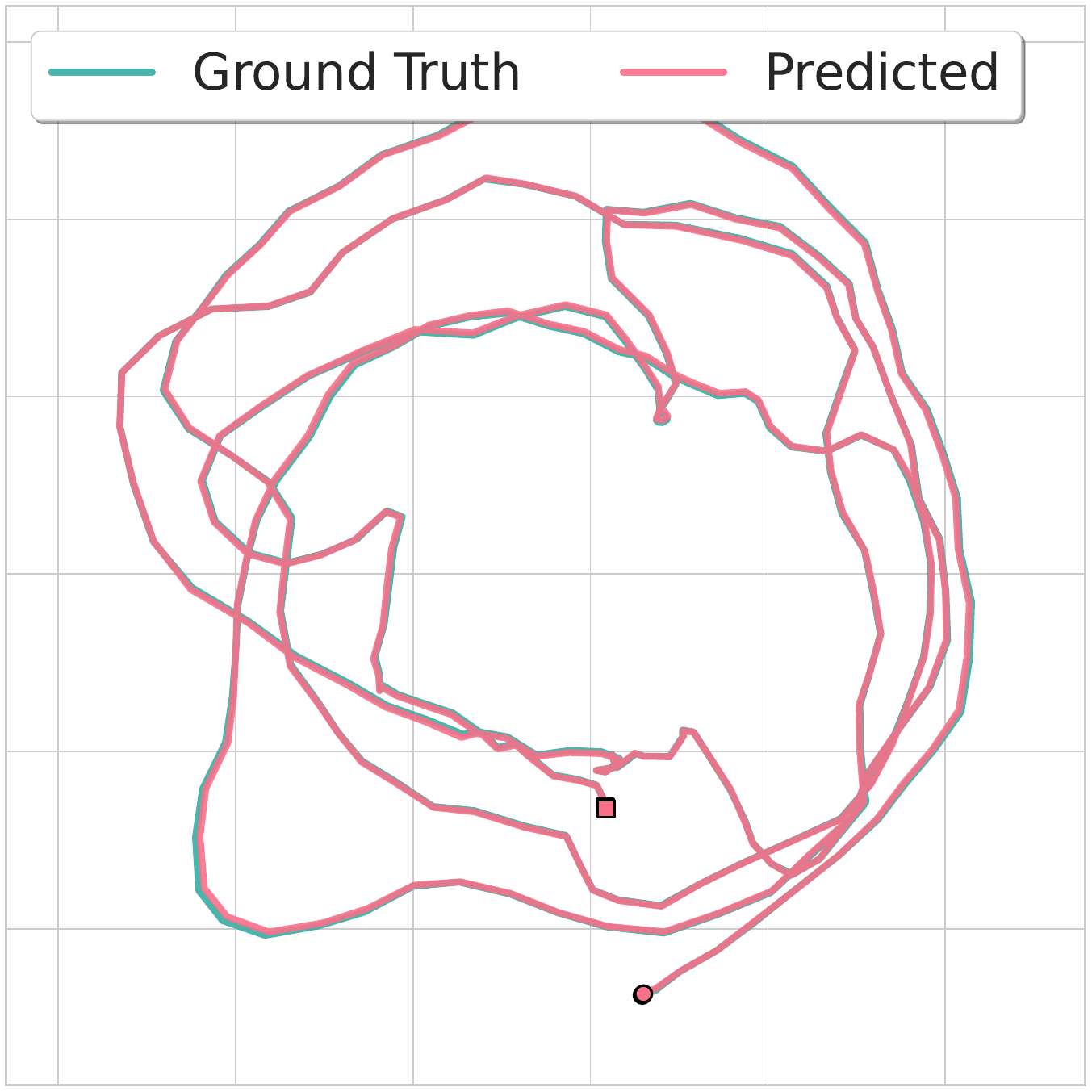} \\

        \raisebox{4.5\height}{Meetingroom} & 
        \includegraphics[width=0.15\textwidth]{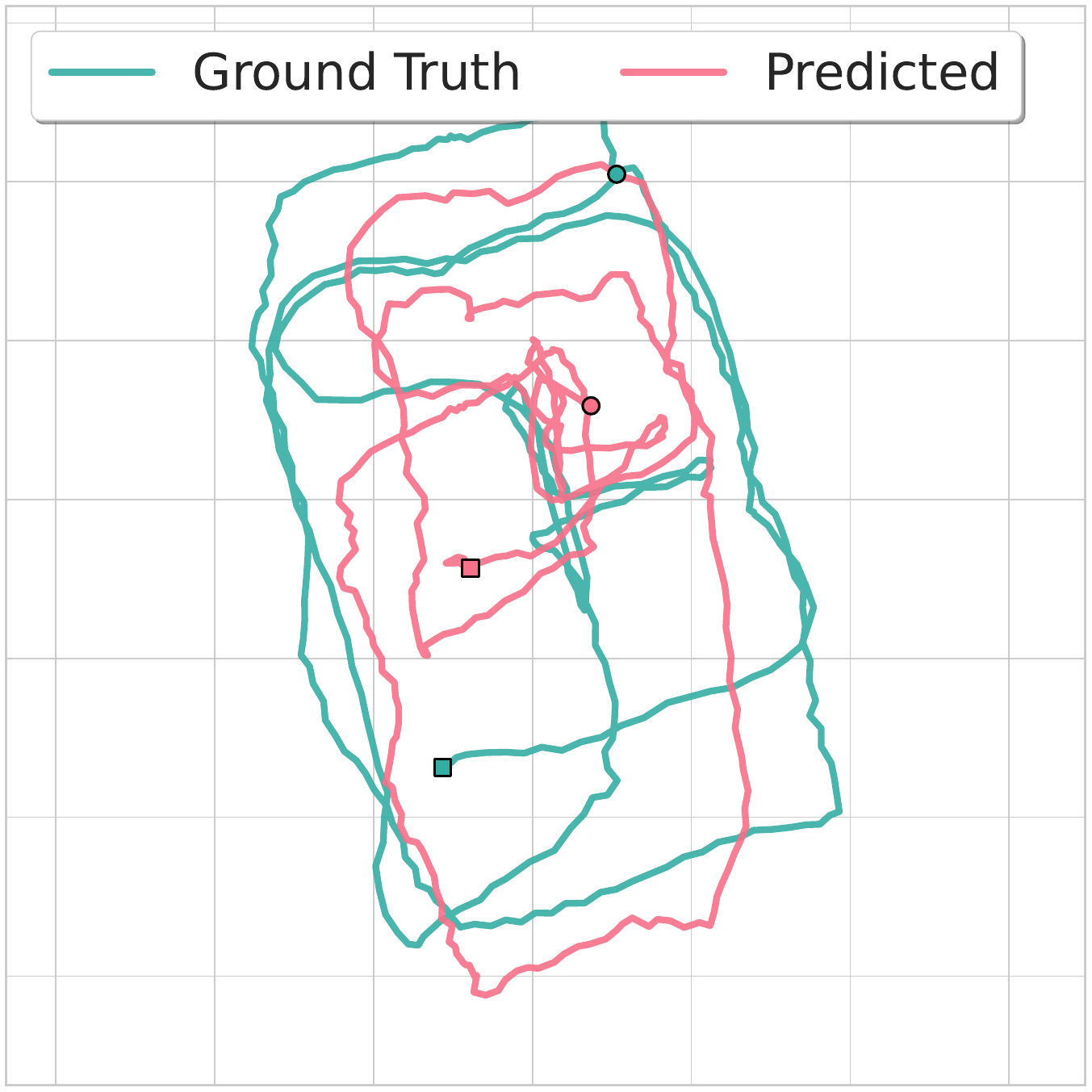} &
        \includegraphics[width=0.15\textwidth]{assets/figures/traj_vis_Meetingroom_cut3r.pdf} & 
        \includegraphics[width=0.15\textwidth]{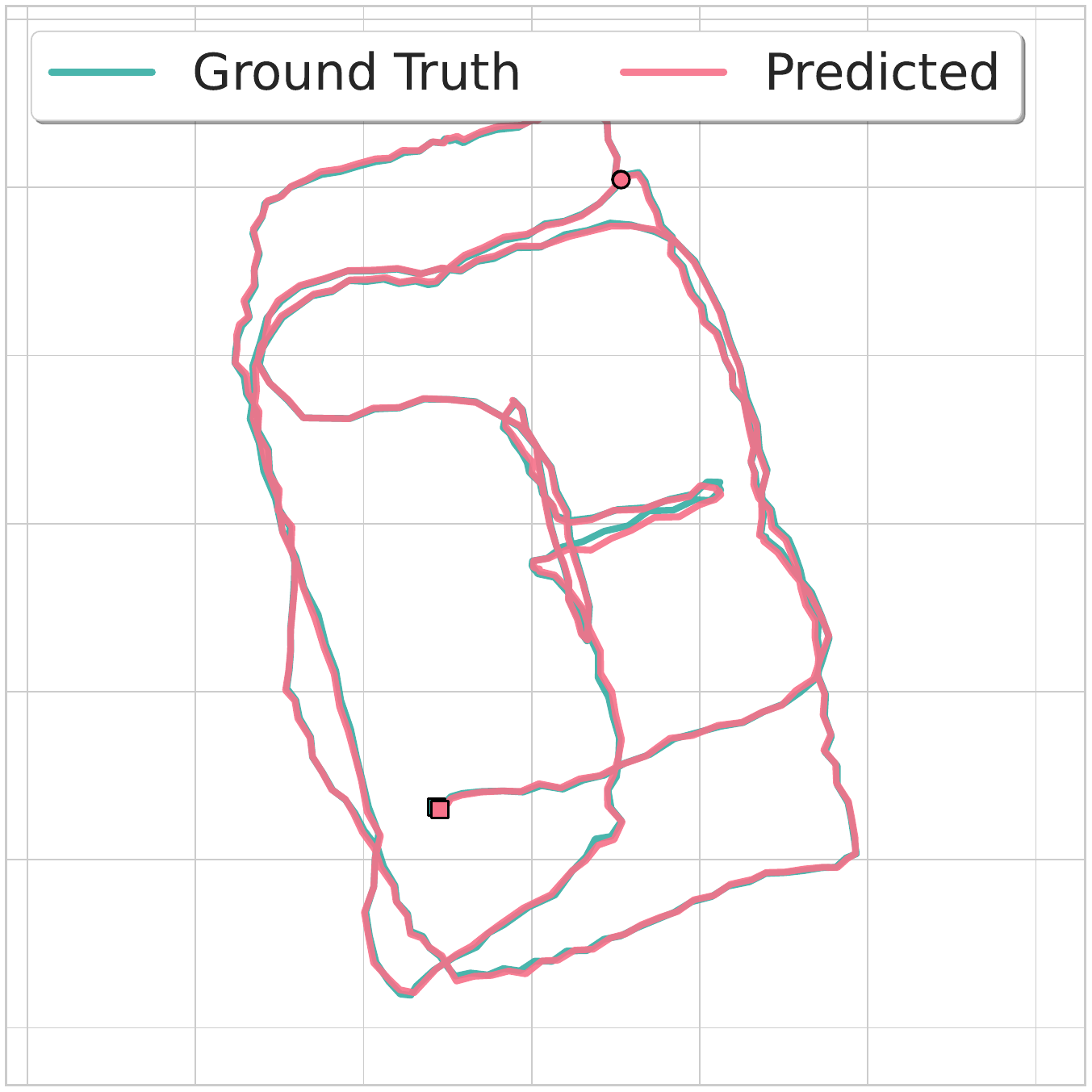} & 
        \includegraphics[width=0.15\textwidth]{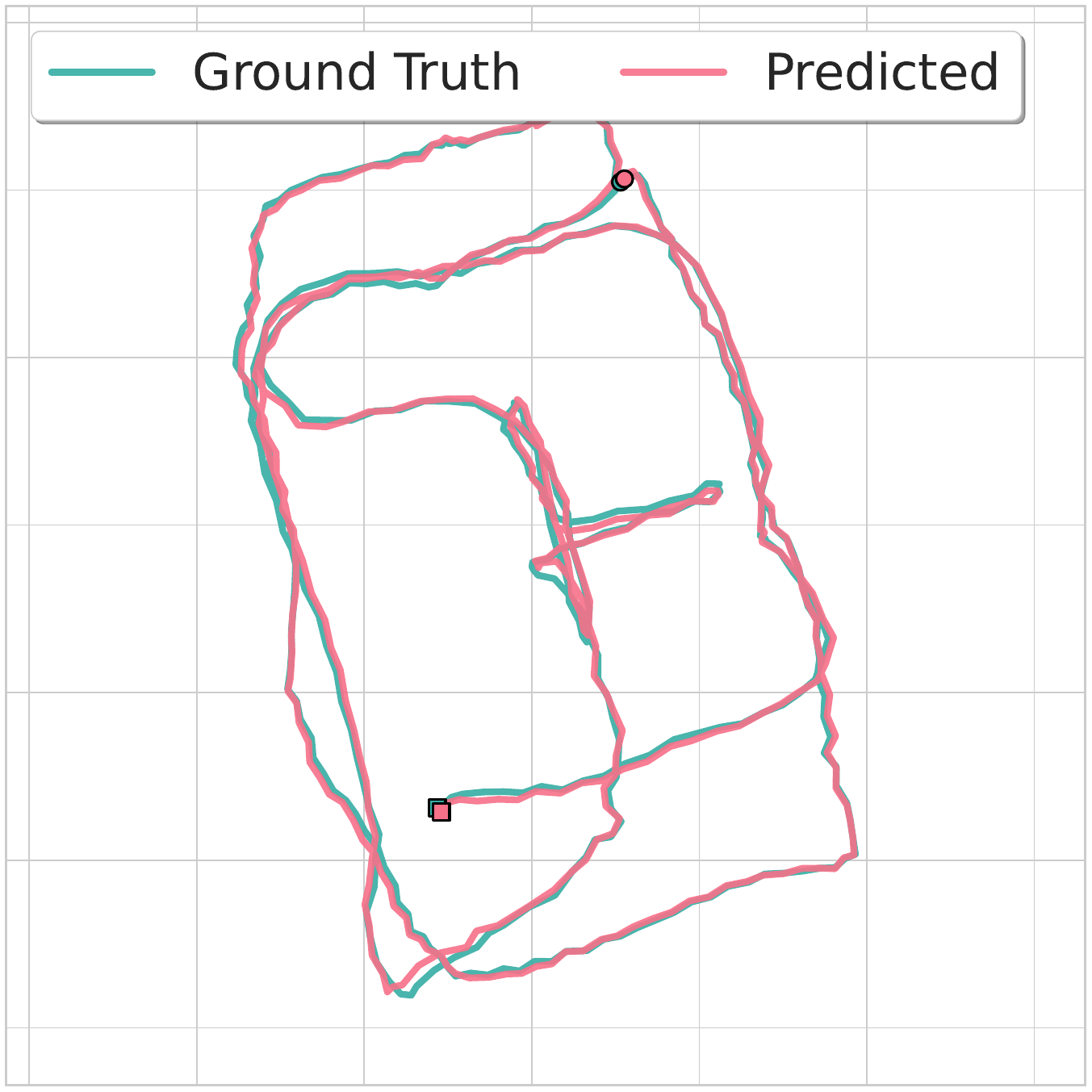} &
        \includegraphics[width=0.15\textwidth]{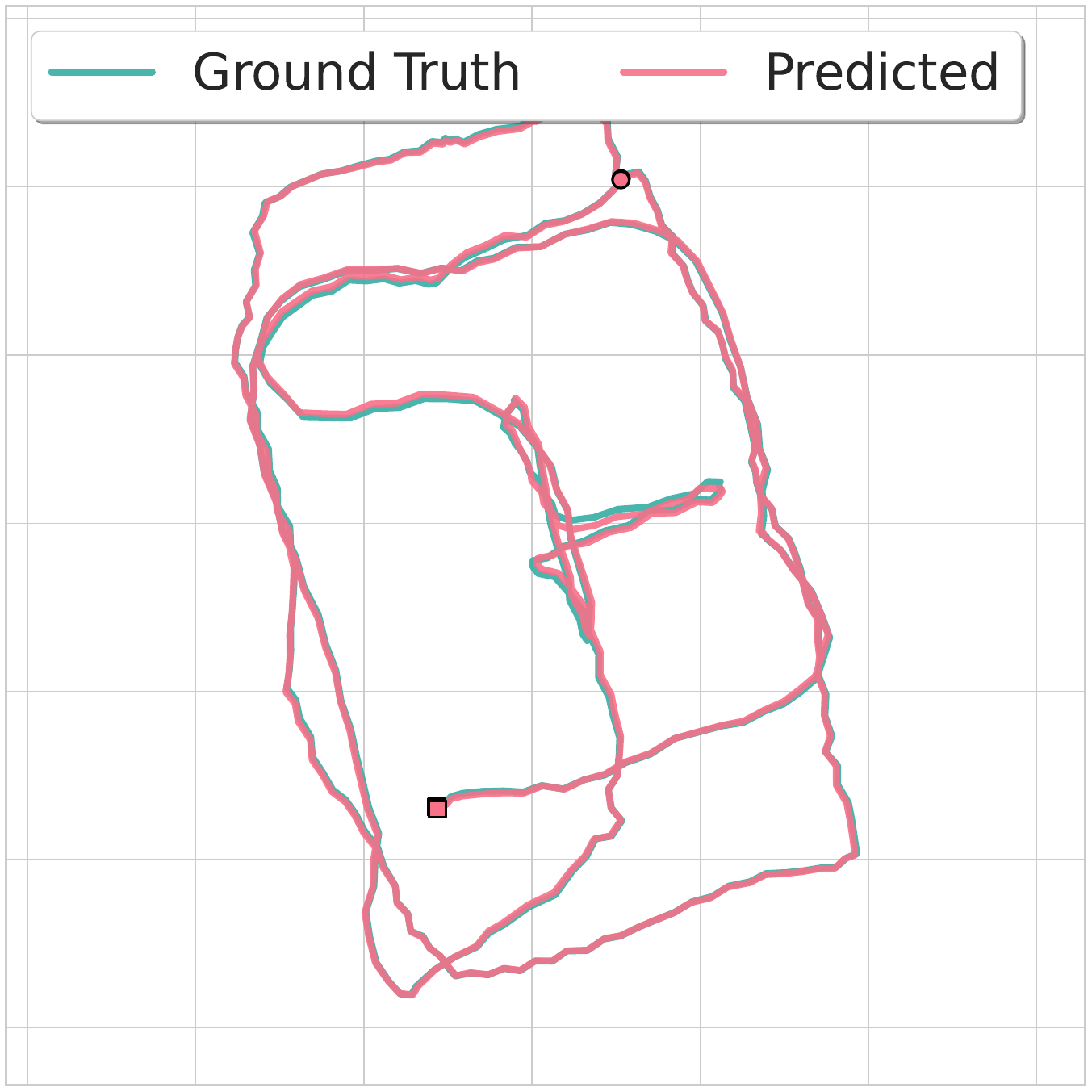} &
        \includegraphics[width=0.15\textwidth]{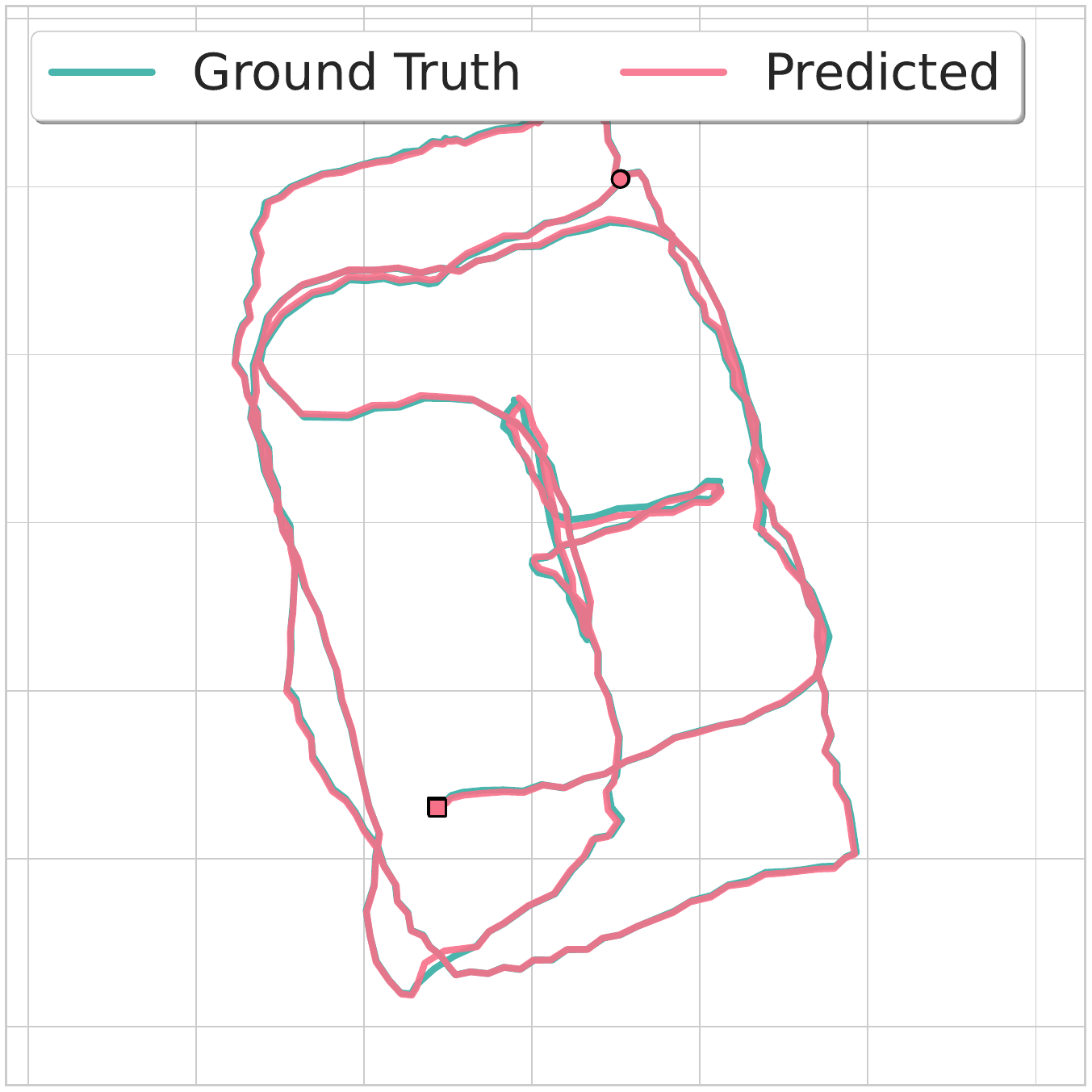} \\

        \raisebox{4.5\height}{Truck} & 
        \includegraphics[width=0.15\textwidth]{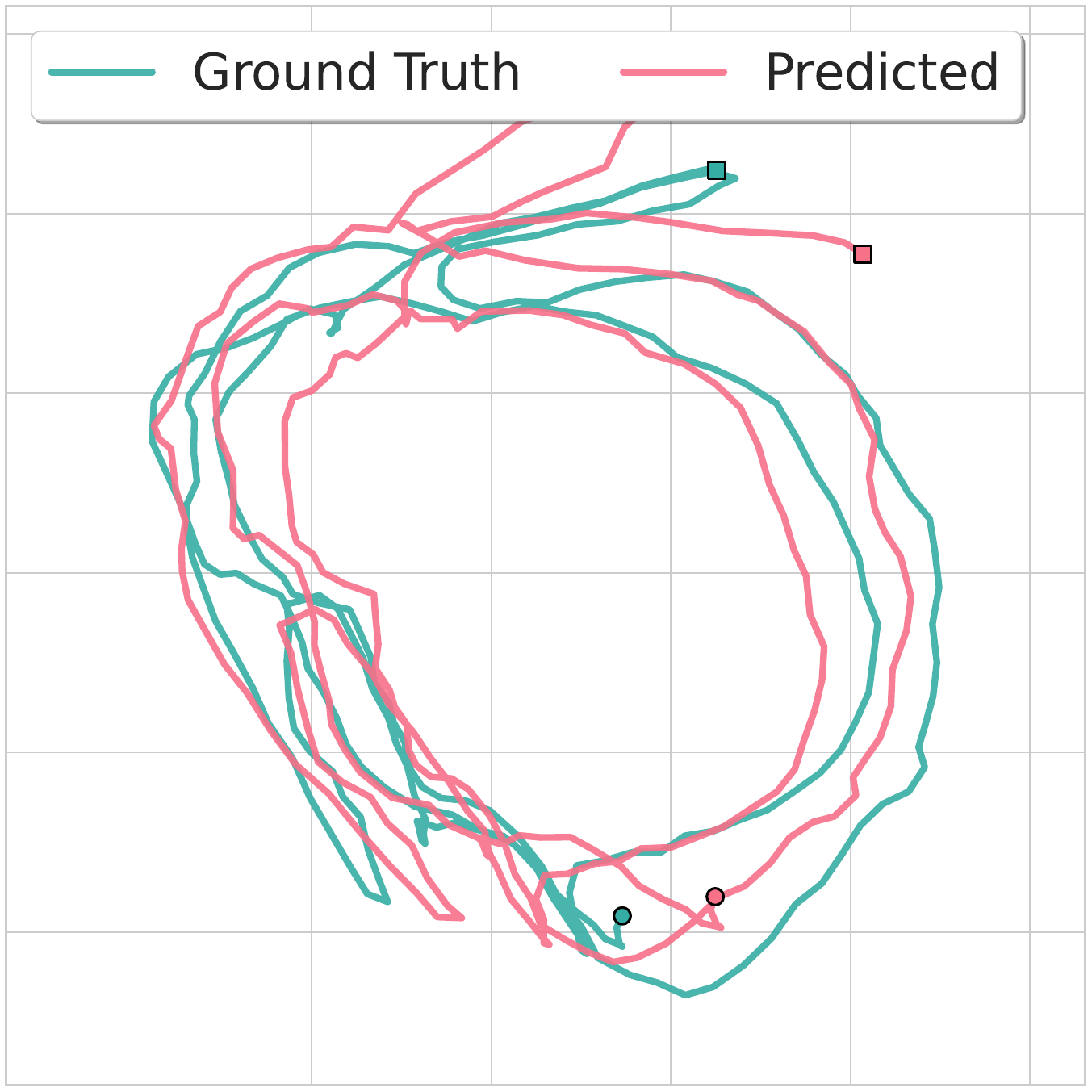} &
        \includegraphics[width=0.15\textwidth]{assets/figures/traj_vis_Truck_cut3r.pdf} & 
        \includegraphics[width=0.15\textwidth]{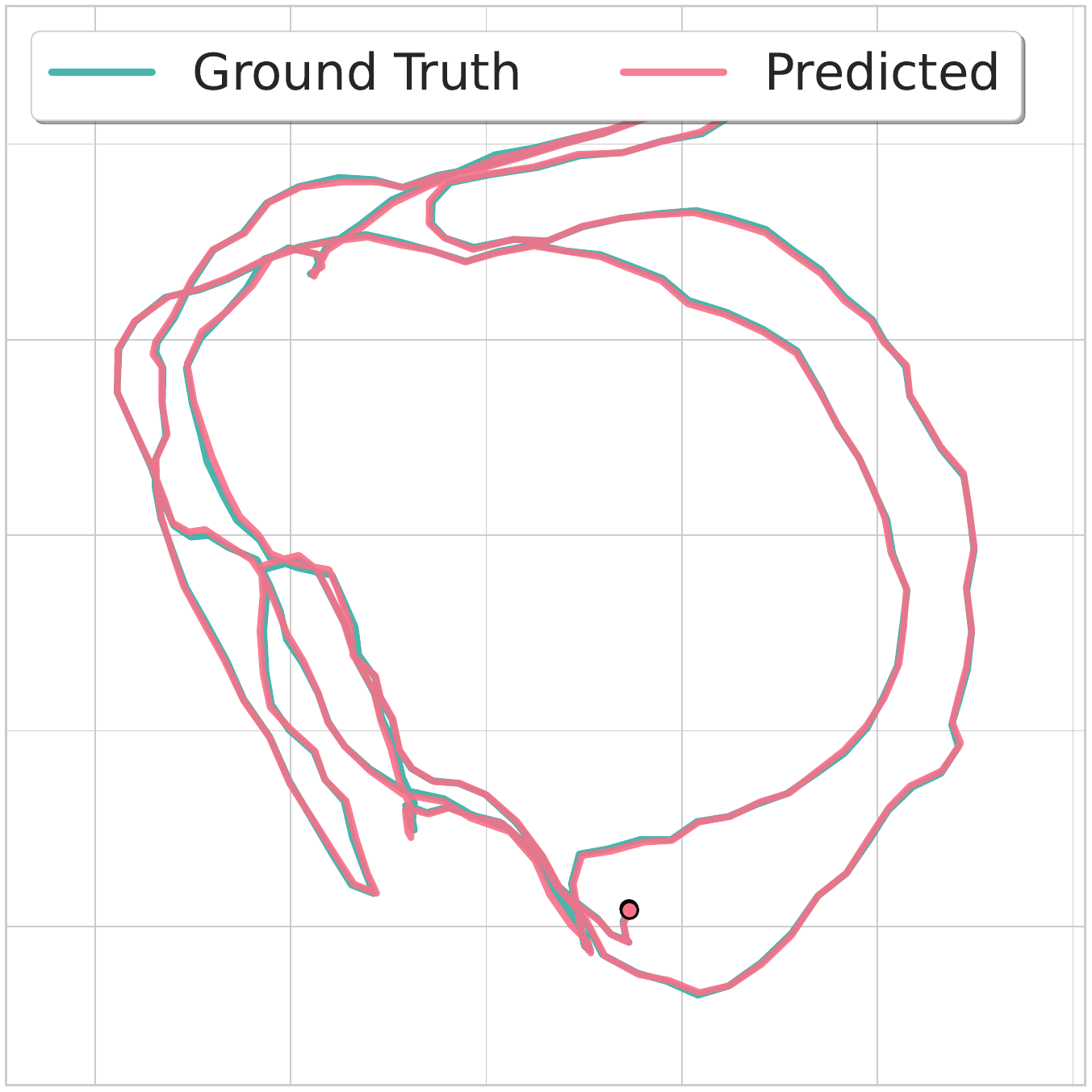} & 
        \includegraphics[width=0.15\textwidth]{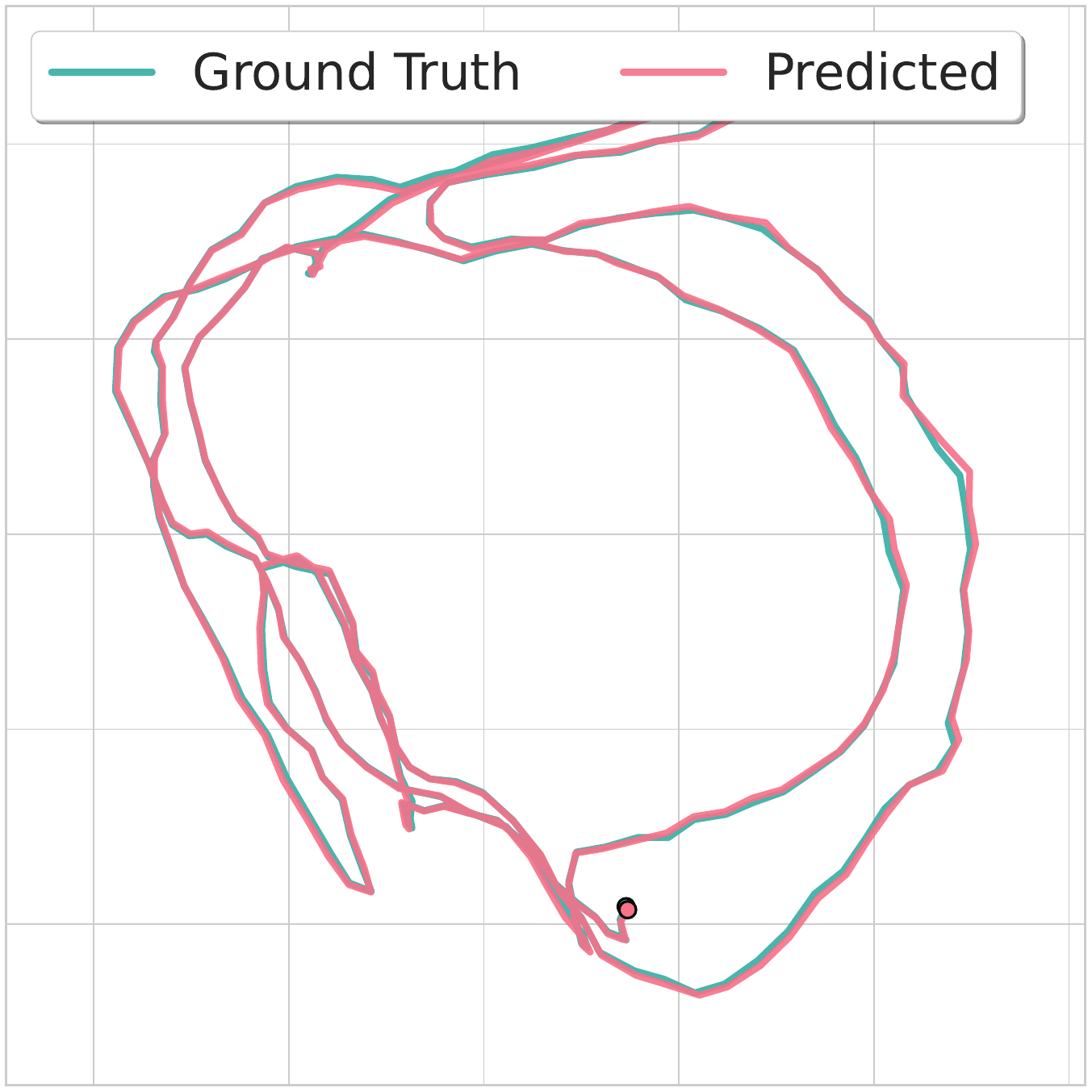} &
        \includegraphics[width=0.15\textwidth]{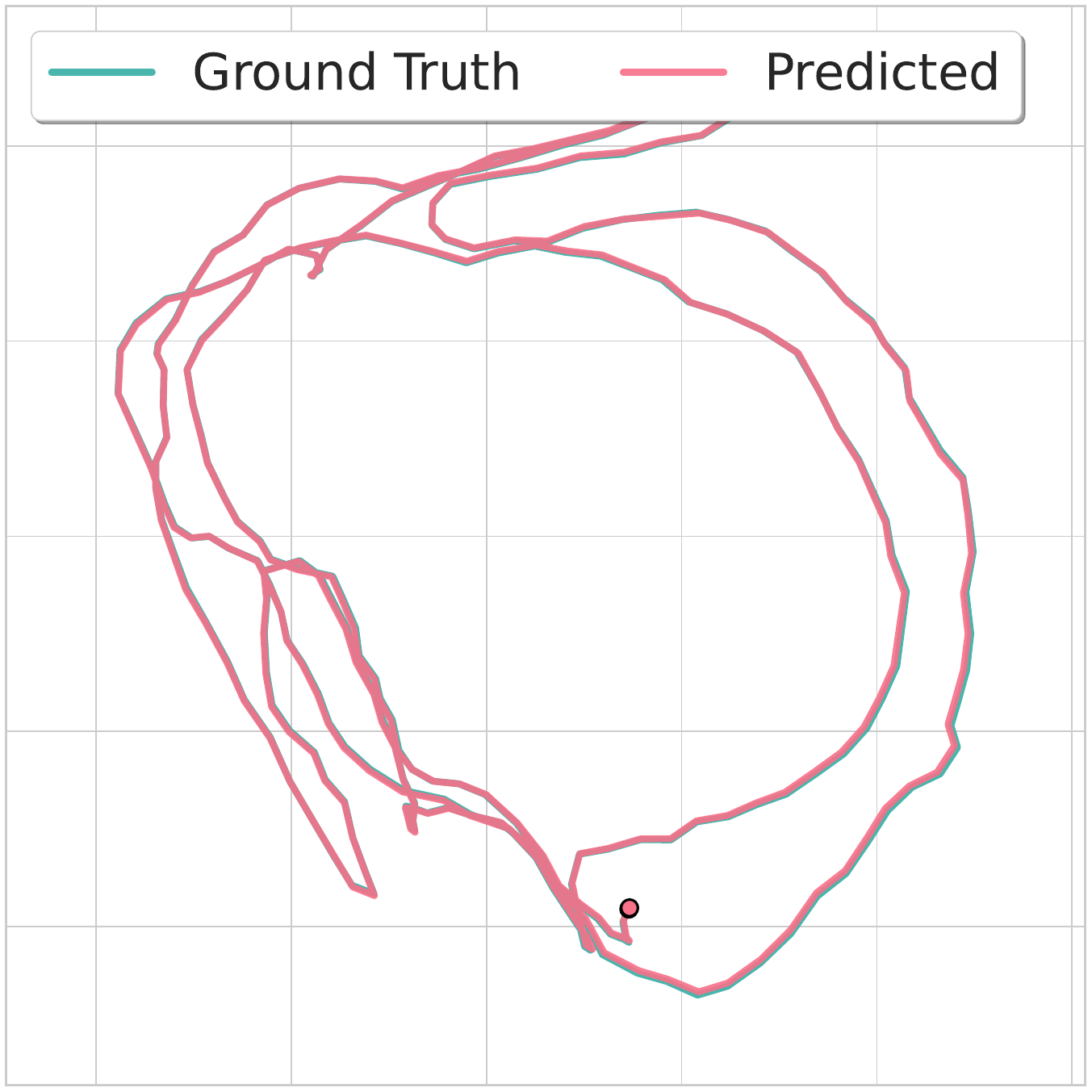} &
        \includegraphics[width=0.15\textwidth]{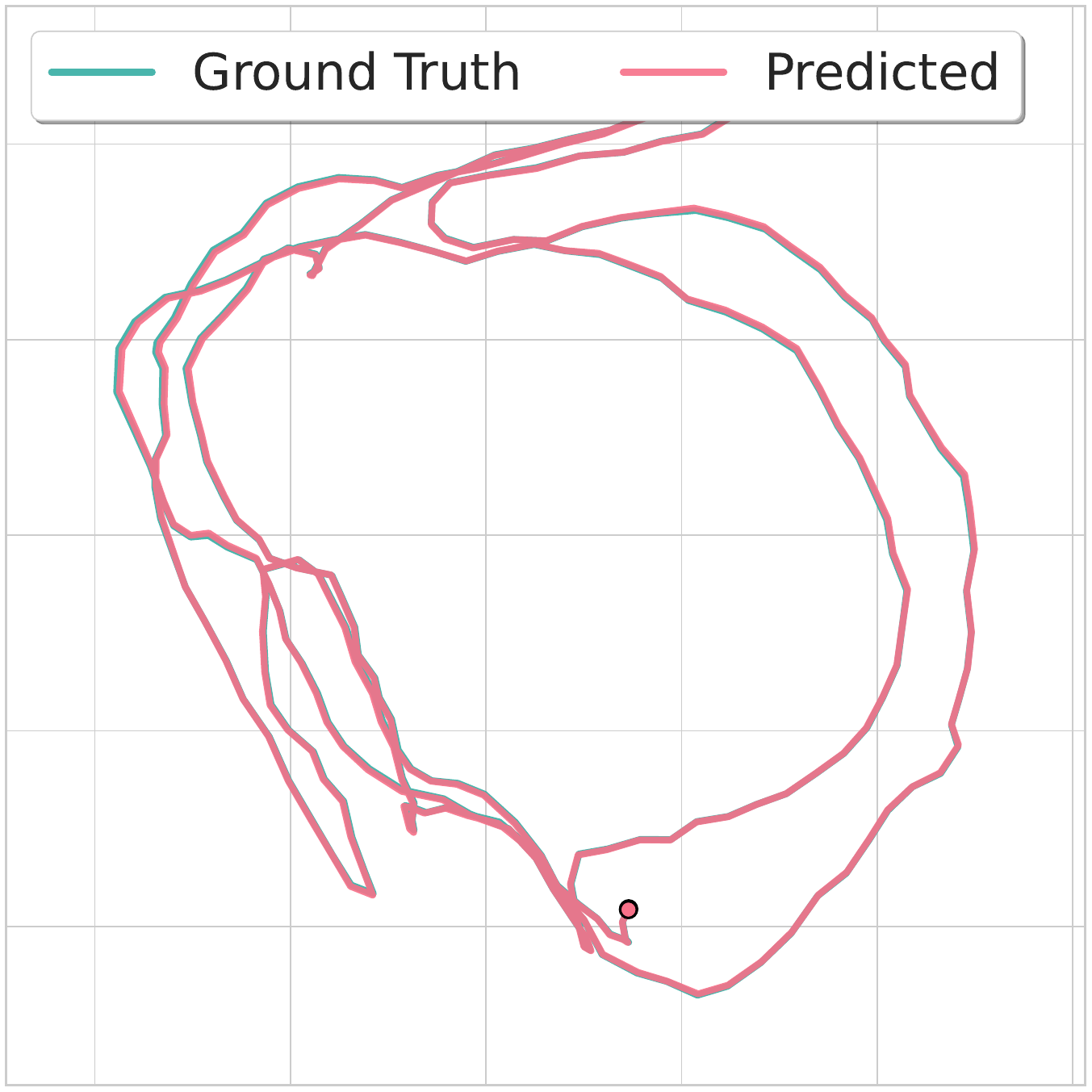} \\

        & \small (a) TTT3R & \small (b) CUT3R & \small  (c) $\pi^3$ & (d) VGGT & (e) ours+$\pi^3$ & ours+VGGT \\[8pt]
    \end{tabular}
    \vspace{-0.5cm}
    \caption{\small Qualitative pose estimation results for all scenes in the Tanks \& Temples dataset. }
    \label{fig:all_tnt_poses}
\end{figure*}

\end{document}